\documentclass{article} % For LaTeX2e
\usepackage{conference,times}

\usepackage{amsmath,amsfonts,bm}

\def\eqref#1{equation~\ref{#1}}
\def\1{\bm{1}}

\def\rvepsilon{{\mathbf{\epsilon}}}

\def\vzero{{\bm{0}}}

\def\vmu{{\bm{\mu}}}

\def\vb{{\bm{b}}}
\def\vc{{\bm{c}}}
\def\vd{{\bm{d}}}

\def\vh{{\bm{h}}}

\def\vp{{\bm{p}}}
\def\vq{{\bm{q}}}

\def\vs{{\bm{s}}}

\def\vu{{\bm{u}}}
\def\vv{{\bm{v}}}
\def\vw{{\bm{w}}}
\def\vx{{\bm{x}}}

\def\vz{{\bm{z}}}

\def\mA{{\bm{A}}}
\def\mB{{\bm{B}}}

\def\mD{{\bm{D}}}

\def\mI{{\bm{I}}}

\def\mL{{\bm{L}}}

\def\mU{{\bm{U}}}
\def\mV{{\bm{V}}}
\def\mW{{\bm{W}}}
\def\mX{{\bm{X}}}

\def\mSigma{{\bm{\Sigma}}}

\DeclareMathAlphabet{\mathsfit}{\encodingdefault}{\sfdefault}{m}{sl}
\SetMathAlphabet{\mathsfit}{bold}{\encodingdefault}{\sfdefault}{bx}{n}

\def\gN{{\mathcal{N}}}

\def\sA{{\mathbb{A}}}

\def\sG{{\mathbb{G}}}

\def\sM{{\mathbb{M}}}

\def\sX{{\mathbb{X}}}

\newcommand{\R}{\mathbb{R}}

\usepackage{hyperref}
\usepackage{url}
\usepackage{microtype}
\usepackage{graphicx}
\usepackage{subcaption}
\usepackage{booktabs} 
\usepackage{amsmath}
\usepackage{amssymb}
\usepackage{mathtools}
\usepackage{amsthm}
\usepackage[utf8]{inputenc} 
\usepackage[T1]{fontenc}    
\usepackage{hyperref}       
\usepackage{url}            
\usepackage{booktabs}       
\usepackage{amsfonts} 
\usepackage{nicefrac}       
\usepackage{microtype}      
\usepackage{xcolor}         
\usepackage[table]{xcolor}
\usepackage{bm}  
\usepackage{graphicx}
\usepackage{booktabs}
\usepackage{multirow}
\usepackage{booktabs}
\usepackage{tabularx}
\usepackage{longtable}
\usepackage{array}
\usepackage{colortbl}
\usepackage{tcolorbox}
\usepackage{enumitem}
\usepackage{mathtools}
\tcbuselibrary{breakable,skins}
\usepackage{listings}
\usepackage{colortbl}
\usepackage{xcolor}
\usepackage{tikz}
\usepackage{pifont}
\usepackage{lucide-icons}
\usepackage{wrapfig}
\usepackage{soul}
\sethlcolor{green!40}

\definecolor{citeblue}{RGB}{10,175,255}
\definecolor{refcolor}{RGB}{202,30,67}
\hypersetup{
    colorlinks=true,      
    citecolor=citeblue,  
    linkcolor=refcolor,   
    urlcolor=citeblue    
}
\definecolor{promptingcolorbox}{RGB}{205,162,171}    
\definecolor{judgebox}{RGB}{129,173,200}
\definecolor{fluencybox}{RGB}{248,242,220}
\definecolor{extractbox}{RGB}{145,196,153}
\definecolor{oodbox}{RGB}{160,70,104}
\newcommand{\dataset}[1]{\textsc{#1}}

\lstdefinestyle{python}{
    language=Python,
    basicstyle=\ttfamily\small,
    breaklines=true,
    frame=single,
    showstringspaces=false
}

\definecolor{rowgray}{gray}{0.93}
\definecolor{idblue}{RGB}{31,119,180}      
\definecolor{oodorange}{RGB}{214,95,0}     
\definecolor{groupshade}{gray}{0.90}

\definecolor{bestcell}{RGB}{130,10,80}   
\definecolor{secondcell}{RGB}{20,55,125}  
\definecolor{affinecol}{RGB}{142,68,173}
\definecolor{linearcol}{RGB}{27,158,119}

\newcommand{\conceptIcon}{\textcolor{axblue}{\Large\lucideicon{lightbulb}}}
\newcommand{\fluencyIcon}{\textcolor{axblue}{\Large\lucideicon{message-circle-check}}}
\newcommand{\instructIcon}{\textcolor{axblue}{\Large\lucideicon{list-checks}}}
\newcommand{\instDivIcon}{\textcolor{divgreen}{\Large\lucideicon{square-stack}}}
\newcommand{\catDivIcon}{\textcolor{divgreen}{\Large\lucideicon{network}}}
\newcommand{\axBenchIcon}{\textcolor{axblue}{\Large\lucideicon{axe}}}
\newcommand{\maxBenchIcon}{\textcolor{maxamber}{\Large\lucideicon{move-3d}}}
\newcommand{\divIcon}{\textcolor{divgreen}{\Large\lucideicon{shapes}}}

\newcommand{\tgt}[1]{\colorbox{blue!15}{\strut #1}}
\newcommand{\dta}[1]{\colorbox{gray!15}{\strut #1}}

\newcommand{\maxbench}{\textsc{MAxBench}}

\newcommand{\onedimicon}{{\Large\lucideicon{move-diagonal}}}
\newcommand{\multidimicon}{{\Large\lucideicon{layers-2}}}
\newcommand{\features}{{\Large\lucideicon{book-open}}}
\usepackage{tikz}

\newcommand{\gmm}{%
  \tikz[baseline=-0.1ex, scale=0.55, transform shape]{%
    \draw[line width=1.0pt, line cap=round, smooth, samples=60,
          domain=-3.5:3.5, variable=\x]
      plot ({0.12*\x},
            {0.45*(exp(-(\x+1.2)*(\x+1.2)/0.8)
                 + 0.8*exp(-(\x-1.3)*(\x-1.3)/1.1))});
  }%
}
\newcommand{\splineicon}{{\Large\lucideicon{chart-spline}}}

\definecolor{axblue}{HTML}{185FA5}
\definecolor{divgreen}{HTML}{3B6D11}
\definecolor{maxamber}{HTML}{D4537E}

\newcommand{\cmark}{\ding{51}}
\newcommand{\xmark}{\ding{55}}

\usepackage[capitalize,noabbrev]{cleveref}

\theoremstyle{plain}

\theoremstyle{definition}

\theoremstyle{remark}

\usepackage[textsize=tiny]{todonotes}

\title{\textsc{MAxBench}: A Multinomial concept recovery benchmark}

\iclrfinalcopy

\author{
Divya Appapogu$^{1}$, \enspace Freya Behrens$^{1}$, \enspace
Yonatan Belinkov$^{2,3}$, \enspace Aaron Mueller$^{1}$ \\[4pt]
$^{1}$Department of Computer Science, Boston University, Boston, MA 02215, USA \\
$^{2}$Department of Computer Science, Technion – Israel Institute of Technology, Israel \\
$^{3}$Department of Computer Science, Kempner Institute, Harvard University, Boston, USA \\
\texttt{\{divsp,fbehrens,amueller\}@bu.edu} \\
\texttt{belinkov@technion.ac.il}
}

\begin{document}

\maketitle

\begin{abstract}

Fine-grained control of language model behaviors (e.g., \emph{steering}) is among the more actionable outcomes of interpretability research. For binary concepts such as refusal, a single direction in activation space often suffices for steering. However, many concepts are not binary: \textsc{Animals} and \textsc{Countries} contain many subcategories, each with multiple instances. For these concepts, the search space over possible representation geometries is far larger than for binary concepts; it is thus not clear what geometries are most appropriate, nor what methods are most effective at recovering them. In this work, we introduce \textsc{MAxBench}, a geometry-agnostic evaluation framework for multinomial concept representations based on sampling from the recovered concept representation. We use \textsc{MAxBench} to compare 10 localization methods (covering 5 geometry types) across 6 concepts and 4 models. Using this framework, we find that (i)~affine subspaces steer more reliably and have greater recall than rank-one or linear subspaces; (ii)~much of this advantage is due to better non-zero offsets rather than the choice of bases; (iii)~manifold steering is competitive with the best methods when applicable; and (iv)~no method consistently outperforms prompting, in alignment with prior findings on binary concepts. These findings underscore the importance of expanding the scope of interpretability research and meta-evaluation to concepts with more varied structure.\footnote{Code is available at: \url{https://github.com/baaigl/MAxBench}}

\end{abstract}
\section{Introduction}

Among the most impactful ideas in interpretability is that some human-understandable concepts can be represented as linear directions in activation space. 
This simple geometric abstraction has enabled both insights into model internals~\citep{word2vec,Bolukbasineurips2016,elhage2021mathematical,park2023LRH,bricken2023SAE1,nanda2023emergent,huben2024SAE2} as well as techniques for modifying model behavior at inference time. 
Steering methods~\citep{turner2025OneDimSteer3,li2023OneDimSteer1,zou2025OneDimSteer2,subramani2022OneDimSteer4} provide a lightweight and interpretable alternative to prompting and fine-tuning (albeit with somewhat lower performance than prompting~\citep{wu2025axbench}).
For binary concepts such as refusal~\citep{NEURIPS2024_f5454485}, a steerable representation can be captured by a single direction in activation space that separates the two states.
However, this intuition becomes less straightforward for more complex concepts, as these may require richer representations.

Indeed, recent evidence suggests that concepts in language models can take many forms: e.g., they may be polytopes \citep{park2025hierarchical} or may lie on multidimensional (and potentially non-linear) manifolds~\citep{engels2025not,shafran2026MFA,tiblias2025SMDS,karkada2026symmetry,bhalla2026SAEmanifold,wurgaft2026manifoldsteeringrevealsshared,gurnee2025geometryofcounting,modell2025originsrepresentationmanifoldslarge,fel2026structuringsparsityblocksparsefeaturizers,sarfati2026shapebeliefs,feucht2026arithmetic}. With regards to evaluation and steering, moving from one-dimensional/binary to multidimensional/multinomial concepts therefore introduces several challenges. First, the space of possible geometries becomes much larger, and may depend on the concept. In the multidimensional setting, a concept may be represented by a set of directions, a Gaussian cloud within the representation space, linear or affine subspaces or non-linear manifolds, among many other possibilities. Second, the dimensionality (or rank) of a concept representation is generally not known in advance and likely varies significantly across concepts. While some concepts have an intuitive geometry (e.g., days of the week tend to form a circle with ordered values~\citep{engels2025not}) we cannot assume \emph{a~priori} that any given concept will have such a structure.

We therefore ask three questions. First, how can we measure and compare the quality of concept representations in a geometry-agnostic manner? Second, how can one steer a concept given arbitrary multinomial concept geometries? Finally, do there exist methods that generally tend to perform well for multinomial concept steering---and if so, why do they perform well?

To investigate these questions, we first study  methods (both existing and novel to this work) for recovering multidimensional concept geometries. We then propose a sampling-based intervention method for steering multinomial concepts in language models.

In this work, we introduce \maxbench, a framework for evaluating and comparing representations of multinomial concepts. As part of \maxbench, we propose evaluation metrics and data for measuring the quality of a multinomial concept localization method. We define quality as encompassing \textbf{exclusivity} (whether the representation captures only the target concept, analogous to precision) and \textbf{completeness} (whether the representation fully covers the concept, analogous to recall). This framework is agnostic to chosen geometry.

Our main contributions are as follows:
\begin{enumerate}[noitemsep,topsep=0pt]
    \item We propose a sampling-based steering framework for evaluating how well a given representation captures a target concept. This framework is geometry-agnostic, and can be applied to linear and non-linear representations.
    \item We introduce \maxbench, a benchmark that leverages this sampling-based framework  to enable comparisons of concept representations across multiple concepts and geometric structures.
    \item We use \maxbench~to compare 10 concept representation discovery methods across 6 concepts, 5 geometries, and 4 models. We find that (i) affine subspaces steer more reliably and have greater recall than rank-one or linear subspaces; (ii) most of this advantage is due to better non-zero offsets rather than the choice of bases; (iii) manifold steering is competitive with the best methods when applicable; and (iv) no method consistently outperforms prompting.
    \item We introduce affine generalizations of existing steering methods for recovering multinomial concept representations.
    % \item We introduce an application of concept-space sampling. 
\end{enumerate}

\begin{figure}[t]
\centering
\includegraphics[width=\linewidth]{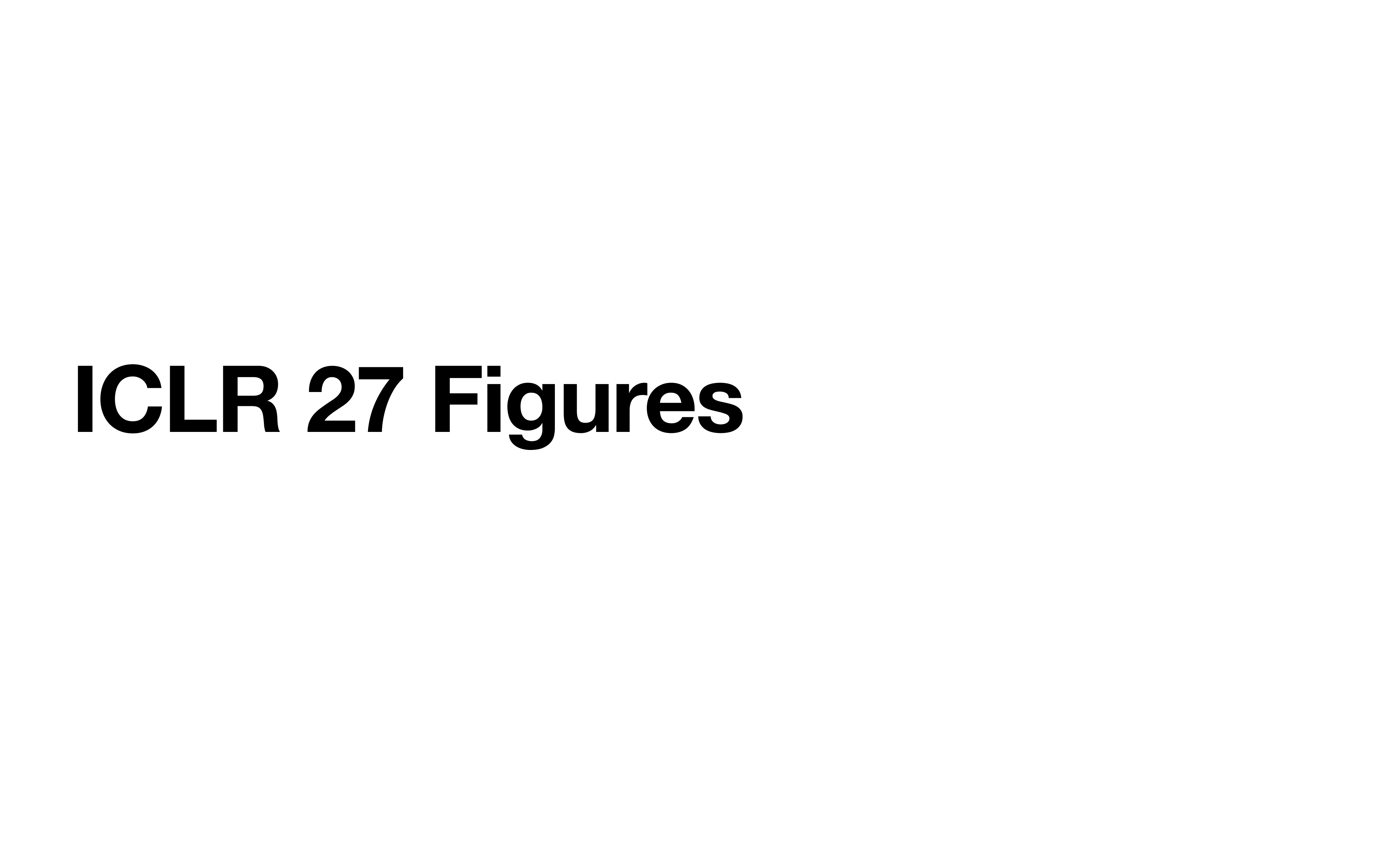}
\caption{
A pipeline to evaluate multinomial concept representations via the \maxbench{} dataset. 
\textbf{(a)}~For a given multinomial concept dataset we can localize a representing geometry in the activation space of a given LLM via different methods §\ref{subsec:concept_recovery_methods}.
\textbf{(b)}~Neutral prompts generate concept-dependent continuations when steered via samples from a specific concept representation §\ref{subsec:sampling}. 
\textbf{(c)} We evaluate both the exclusivity and completeness of different concept representations from a set of continuations that were generated for different samples $\mathbf{s}$ and several neutral prompts. Taken together, several metrics define the final \maxbench{} score §\ref{subsec:scoring}.}
\label{fig:teaser}
\end{figure}

\section{Preliminaries}
\label{sec:prelims}

\paragraph{Concept representations.}
We informally define a \textit{concept representation} as a geometric object in a model’s activation space that captures the boundaries of and variance within a semantic concept, such as \textsc{Animals} or \textsc{Days} of the week. Ideally, a concept representation should be both \emph{complete}, i.e., all the information relevant to the concept can be extracted from this representation, and \emph{exclusive}, i.e. information unrelated to the concept cannot be extracted from this representation. For example, if concept $C$ is represented by an affine subspace $\sA$, then all information required to recognize or generate text about $C$ (but no more) should be recoverable from $\sA$, while directions orthogonal to it should ideally contain no additional information specific to the concept. We formally define the type of geometries we consider below.

\subsection{Geometries}
\label{subsec:geometries_definitions}
We consider linear and affine subspaces (including existing one-dimensional linear geometries), as well as non-linear manifolds~\citep{wurgaft2026manifoldsteeringrevealsshared,karkada2026symmetry,engels2025not,sarfati2026shapebeliefs}. For concision, we focus primarily on affine subspaces in the main text, as this is the most general linear geometry applicable to all concepts considered in our experiments.

Formally, we define an affine subspace $\sA \subset \R^{d}$ of the form \( \sA = \{ \mL\vq + \vb \mid \vq \in \R^{k} \} \) where $\mL \in \R^{d \times k}$ is a matrix with orthonormal columns defining a $k$-dimensional linear subspace, $\vb \in \R^d$ is an offset vector, and vector $\vq \in \R^k$ denotes the coefficients in the subspace basis. A linear subspace is an affine subspace where the offset $\vb$ is the origin. 
 
We additionally study non-linear manifolds. In this work, we use non-linear manifold to refer to a manifold~\citep{lee2012smooth} that cannot be cast as an affine subspace of $\R^d$. Unlike an affine subspace, which is globally linear, a non-linear manifold may exhibit curvature while remaining locally Euclidean. In this work, we study manifold representations that are obtained by spline-fitting as introduced in~\cite{sarfati2026shapebeliefs,wurgaft2026manifoldsteeringrevealsshared}. 

In their current form, these methods can only be applied to concepts with ordered values, such as \textsc{Days} and \textsc{Years}. Extending these methods to unordered categorical concepts like \textsc{Animals} is non-trivial, so we leave evaluation of non-linear manifold methods on unordered concepts to future work.

\subsection{Data}

We assume that the concept geometries described above correspond to low-dimensional structures in the space of language model hidden activations from the residual stream at a given layer. Specifically, for a set of $m$ input tokens ${x}_{i=1}^m$ that occur just before a concept-related token, we observe hidden activation vectors \(\{\vh^{(i)}\}_{i=1}^m\), where \(\vh^{(i)} \in \R^d\); $d$ is the dimensionality of the language model representations. We assume that each concept is represented by a low-dimensional structure of dimensionality $r$, where $r < d$, for example, an affine subspace $\sA_{c} \subset \R^{d}$ or non-linear manifold $\sM_{c} \subset \R^{d}$ embedded in this ambient space. We will use ``concept recovery'' to refer to methods used to locate these geometric objects in language model activations. 

Recovering a concept geometry requires observing a concept across diverse contexts.

In realistic settings, we may have activations for only a subset of instances belonging to a concept. For example, when learning the geometry corresponding to the concept \textsc{Animals}, it is tedious and likely unrealistic to collect a dataset containing the complete set of all possible instances and corresponding activations. A good concept recovery method should nevertheless recover a geometry that generalizes well (i.e., contains instances not seen during concept recovery) in spite of incomplete data.

Based on these properties, we construct a dataset consisting of six concepts, chosen in part because they have been studied to understand the geometry of semantic representations in language models~\citep{lee2026decomposingQK,park2025hierarchical} (\dataset{Animals}, \dataset{Countries}, \dataset{Vehicles}, \dataset{Plants}, \dataset{Days}, and \dataset{Years}) and a random baseline. Details of our dataset are given in Section~\ref{subsec:concept_dataset}.

\section{\maxbench: A Multinomial Concept Recovery Benchmark}

We introduce \maxbench, a benchmark for evaluating  completeness and exclusivity of concept geometries. The evaluation pipeline consists of three stages. First, we apply different concept localization methods to recover different concept-related geometric objects (Section~\ref{subsec:concept_recovery_methods}). Second, we introduce a sampling-based evaluation where we sample points from the recovered object and use them to intervene on language model activations (Section~\ref{subsec:sampling}). Third, we score the resulting outputs using metrics designed to measure the completeness and exclusivity of the recovered concept geometry (Section~\ref{subsec:scoring}). An overview of our full evaluation pipeline is in Figure~\ref{fig:teaser}.

\subsection{Concept Localization Methods}
\label{subsec:concept_recovery_methods}
We consider methods for concept localization that differ in three key aspects: the geometry they assume, how the parameters (if any) are set, and supervision type. Learning approaches vary between degrees of supervised or unsupervised representation. If the method is supervised, the supervision type describes if the method uses only positive examples (i.e., valid instances of the concept) or positive--negative pairs (i.e., pairs of inputs where one contains a concept instance and the other does not). All methods we consider are summarized in Table~\ref{tab:table1_methods}.

\begin{wraptable}[28]{r}{0.53\textwidth}
\caption{
Comparison of concept localization methods evaluated in \maxbench{}. \textbf{Labels} indicates whether a method is supervised or unsupervised. \textbf{Data} describes the type of data used: concept refers to data specifically collected about the target concept(described in Appendix Table~\ref{tab:dataset-examples}), whereas PILE~\citep{pile} is general-purpose data containing broad information beyond the concept of interest. \textbf{Pairs} indicates whether both positive and negative examples are required. For example, a binary method such as DiffMean requires examples containing the concept (e.g., animal) as well as examples not containing it.
$^\ast$SAE and MFA are trained without contrastive supervision; pos/neg examples are used only for post-hoc feature/component selection.
}
\centering
\scriptsize
\setlength{\tabcolsep}{2.5pt}
\renewcommand{\arraystretch}{0.95}
\begin{tabular}{llclc}
\toprule
\textbf{Method} &
\textbf{Geometry} &
\textbf{ Labels} &
\textbf{ Data} &
\textbf{ Pairs} \\
\midrule
DiffMean-r1       & \onedimicon\ Direction         & \cmark & Concept  & \cmark \\
ReFT-r1           & \onedimicon\ Direction         & \cmark   & Concept  & \cmark \\\midrule
DiffMean          & \multidimicon\ Affine subspace      & \cmark   & Concept  & \xmark \\
Probe             & \multidimicon\ Affine subspace      & \cmark   & Concept  & \xmark \\
Schatten Probe      & \multidimicon\ Affine subspace      & \cmark   & Concept  & \xmark \\
PCA               &  \multidimicon\ Affine subspace      & \xmark & Concept  & \xmark \\
Factor Analysis   &  \multidimicon\ Affine subspace      & \xmark & Concept  & \xmark \\\midrule
SAE               & \features\ Collection of Directions   & \xmark & Pile            & \xmark$^\ast$ \\
MFA               & \gmm\ Mixture of Gaussians     & \xmark & Pile            & \xmark$^\ast$ \\\midrule
Spline Fitting & \splineicon\ Non-linear manifold   & \cmark   & Concept  & \xmark \\
\bottomrule
\end{tabular}

\label{tab:table1_methods}
\end{wraptable}

Across all methods, the input is a collection of activation vectors \(\{\vx^{(i)}\}_{i=1}^m\), with \(\vx^{(i)}\in\R^d\). Methods that require concept supervision additionally have labels \(y^{(i)} \in \{0,\ldots,K-1\}\) for each activation. For one-dimensional methods, we use binary labels \(y^{(i)} \in \{0,1\}\), indicating the absence or presence of the target concept.
\subsubsection{Linear}

\paragraph{\onedimicon~~One-dimensional methods.}
In the one-dimensional setting, concept localization methods aim to identify a single direction in activation space that captures variation associated with a target concept. A common approach is \textit{Difference in Means} (DiffMean-r1)~\citep{marks2024geometrytruthemergentlinear}, which defines the concept direction as the difference between the mean activations of positive and negative examples defined as 
\(\vd = \frac{1}{|\{i:y^{(i)}=1\}|} \sum_{i:y^{(i)}=1}\vx^{(i)} - \frac{1}{|\{i:y^{(i)}=0\}|} \sum_{i:y^{(i)}=0}\vx^{(i)}\).

Another method proposed specifically to improve both steering performance and concept detection is \textit{rank-1 Representation Finetuning} (ReFT-r1;\citealp{wu2025axbench,wu2024reft}). ReFT-r1 learns a single direction \(\vd\) from supervised positive and negative examples, where positive responses express the target concept and negative responses do not. The learned direction is used both to detect the presence of the concept in model representations and to steer the model toward it, with \(\vd\) optimized through a language-modeling objective with additional sparsity regularization. Further details are provided in~\cite{wu2025axbench}. For one-dimensional methods we only have direction $\vd$ and offset \(\vb=\mathbf{0}\).

\paragraph{\multidimicon~~Affine methods.}
Affine concept recovery methods aim to localize a \(k\)-dimensional affine subspace \(\sA=\{\mL\vq+\vb \mid \vq\in\R^k\}\) rather than a single direction. 
The methods listed in Table~\ref{tab:table1_methods}~differ primarily in how \(\mL\) and \(\vb\) are estimated.

\textit{PCA}~\citep{pearson1901lines} estimates \(\mL\) as the subspace spanned by the top principal directions of the centered activations. \textit{Factor Analysis}~\citep{spearman1961general} estimates \(\mL\) as the subspace spanned by the learned factor loadings. \textit{Linear Probe}~\citep{alain2016understanding} first learns a classifier weight matrix \(\mW\) from the activations. We then apply SVD to  \(\mW\) and define \(\mL\) as the matrix containing its top singular vectors.

We propose a \textit{Schatten Probe} ~\citep{braun2025schatten,Giampouras2020schatten} (or also referred to as Sparse Probe) as an extension to linear probe to encourage low-rank weights through approximate Schatten-$p$ norm regularization. Specifically, we parameterize the classifier weights as \(\mW=\mA\mB^\top\) and jointly regularize corresponding columns of \(\mA\) and \(\mB\). After training, we reconstruct \(\mW\), apply SVD, and define \(\mL\) from the leading singular directions in activation space.

We also extend
\textit{Difference in Means} (DiffMean; \citealp{marks2024geometrytruthemergentlinear}), where directions are constructed from pairwise differences between class-conditional mean activations. For each pair of classes \(c_i,c_j\), we compute
\[
\vd_{i,j}=\vmu_{c_i}-\vmu_{c_j},
\qquad
\vmu_{c_j}=\frac{1}{|\sX_{c_j}|}\sum_{\vx\in\sX_{c_j}}\vx.
\]
where $\sX_{c_j}$ is the set of all activations that correspond to a particular class $c_j$.
We stack these difference vectors to form \(\mD\), apply SVD, and define \(\mL\) from its leading singular directions in activation space.

For all affine methods, we set the offset \(\vb\) to the mean of the activations of positive examples; we leave other strategies for estimating the offset to future work. All affine methods are trained using only positive examples of the target concept.

\paragraph{\features~~Collection of Directions}
Unlike the methods described above, which recover a single concept direction or a single affine subspace, \textit{sparse autoencoders} (SAEs; \citealp{bricken2023SAE1,huben2024SAE2}) learn a dictionary of feature directions $f_{\mathrm{enc}}(\vx) = \operatorname{ReLU}\!\left(\mL_{\mathrm{enc}}\vx+\vb_{\mathrm{enc}}\right)$. Each dictionary element corresponds to a one-dimensional direction in activation space. We use existing pre-trained SAEs for our experiments. To identify which features correspond to the target concept, we score each feature by the Matthews Correlation Coefficient (MCC) between its activations and a binary label indicating whether the point is related or unrelated to the concept, and retain the top-($k$) scoring features. Each selected feature defines a one-dimensional direction. We use the corresponding SAE decoder vectors as the columns of the recovered basis, \(\mL = [\vw_{f_1}\mid \cdots \mid \vw_{f_k}]\)
and sample from the subspace spanned by these directions. We set the offset to \(\vb=\mathbf{0}\).

\paragraph{\protect\gmm~~Gaussian mixture}
A \textit{Mixture of Factors Analyzer} (MFA; \citealp{shafran2026MFA}) learns a collection of low-dimensional Gaussian components, each associated with an affine subspace \(\sA_c\). Each component \(c\) has a centroid \(\vb_c \in \R^d\) and factor-loading matrix \(\mL_c \in \R^{d\times k}\) and MFA models the activation space as \(\{\sA_c\}_{c=1}^{C}\). For a target concept, we select the component \(c^\star\) that is most associated with the target concept and use \(\sA_{c^\star}\) as the recovered concept representation. The method we use to select \(\sA_{c^\star}\) is described in Appendix~\ref{app:MFA_select_gaussian}.

\subsubsection{Non-Linear}
\paragraph{\splineicon~~Manifold Methods.} Following~\citet{wurgaft2026manifoldsteeringrevealsshared}, we first compute a centroid for each concept value and project these centroids into a low-dimensional subspace using PCA. A cubic spline is then fit through the projected centroids; this forms the concept manifold. We use the implementation of \cite{wurgaft2026manifoldsteeringrevealsshared} to fit the spline and its associated centroids.

Further details about the methods are provided in Appendices~\ref{appsec: method_all_details}, ~\ref{appsec: sae_feature_selection} and ~\ref{app:MFA_select_gaussian}.

\subsection{Sampling-based Evaluation}
\label{subsec:sampling}

After we recover a concept representation using one of the localization methods described above, we next evaluate whether the points contained in that representation correspond to the intended concept. Intuitively, if a geometry faithfully captures a semantic concept, then points sampled from the geometry should decode only
to valid instances of that concept (exclusivity) while covering the full range of values and categories contained by the concept (completeness). For example, samples drawn from the representation corresponding to the concept \textsc{Animals} are used to intervene on a language model, the resulting generations should exclusively contain sentences related to animals. Furthermore, the samples should cover diverse instances of the concept rather than collapsing to a narrow subset (e.g., only mammals).  We describe how we sample from different geometries below.

\paragraph{Sampling from an affine subspace.}

Given a recovered affine subspace $\sA_{\text{recovered}} = \{\mL\vc + \vb \mid \vc \in \R^k\}$, we sample $n$ points $\{\vp^{(i)}\}_{i=1}^n$ from the subspace by drawing coefficients $\vz^{(i)} \sim \mathcal{N}(\mathbf{0}, \sigma^2 \mI_k)$ and computing $\vp^{(i)} = \mL\vz^{(i)} + \vb$. Since affine subspaces are unbounded, we select $\sigma$ empirically to ensure sampled points remain within the typical range of the model's activation space. More details about how we select hyperparameters are in Appendix~\ref{app:hyperparam_tuning}. 

A linear subspace is a special case of affine subspaces with $\vb=\vzero$. A one-dimensional representation corresponds to the special case $r=1$, making sampling with one-dimensional methods (e.g., steering vectors) a special case of sampling from an affine subspace.

\paragraph{Sampling from a non-linear manifold.}

Given a recovered manifold \(\sM\), we sample \(n\) points \(\{\vp^{(i)}\}_{i=1}^n\) by defining a global parameter \(s \in [0,K)\), where \(K\) is the number of concept values. Each integer \(k\) corresponds to the centroid of the \(k\)th concept value. For a sampled value \(s^{(i)}\sim\mathcal{U}(0,K)\), we write \(s^{(i)} = k+\gamma\), where \(k\) identifies the corresponding spline segment and \(\gamma\in[0,1)\) specifies the position along the spline segment connecting centroids \(k\) and \(k+1\). We then evaluate this spline at \(s^{(i)}\) to obtain the sampled point \(\vp^{(i)}\). Repeating this procedure yields samples from the spline-based manifold. If the spline forms a closed loop, as in the case of days of the week, we include the spline segment connecting the final concept-value centroid back to the first.

\paragraph{Steering with samples.} To decode each sampled point, we intervene on the language model's internal representation using that point and observe the resulting model output. Interventions are applied at the same layer from which the concept representations are extracted. For a sampled point \(\vs\) and the current hidden activation \(\vh\), we intervene by replacing it with
\[
\vh' = (1-\alpha)\vh + \alpha\vs. \footnote{For one-dimensional methods such as DiffMean-r1 and ReFT-r1, additive steering of the form \(\vh'=\vh+\alpha\vs\) is commonly used and has often been found to be effective. In our experiments, we observed that additive steering substantially reduced output diversity, with generations frequently collapsing to a small set of entities from the target concept.}
\]

The sampling procedure determines the direction or location within the recovered concept geometry, while \(\alpha\) controls the strength of the intervention. We select \(\alpha\) using a grid search, with additional hyperparameter-tuning details provided in Appendix~\ref{app:hyperparam_tuning}. During generation, we exponentially decay \(\alpha\) across decoding steps, encouraging the model toward the target concept early in generation while preserving fluency and instruction following.

\subsection{Scores}
\label{subsec:scoring}

Given the outputs produced by the sampling procedure above, we quantify how well each recovered geometry captures its target concept using the following metrics.

\textbf{\conceptIcon~Concept Score.} Following \textsc{AxBench}~\citep{wu2025axbench}, we measure whether the generated output contains text relevant to the target concept. For example, if the target concept is \textsc{Animals}, we check whether the output mentions any animal. For each sampled point, we assign a concept score in \(\{0,1,2\}\). A score of \(0\) indicates that the target concept is absent; \(1\) indicates that the concept is present but expressed unnaturally, or outside its intended meaning; and \(2\) indicates that the concept is clearly and correctly expressed. This quantifies exclusivity, and is analogous to precision.

\textbf{\divIcon~Diversity Score.} We introduce a diversity score to quantify how well a recovered geometric representation covers the intended concept. The score is computed in two stages. First, we extract the entities mentioned in each generated sentence (e.g., \emph{cat}, \emph{sparrow}) together with their associated category labels (e.g., \emph{mammal}, \emph{bird} for the \textsc{Animals} concept). Second, we aggregate the extracted entities and categories across all generated samples and compute the normalized Shannon entropy at two levels: (i) \instDivIcon~\emph{instance-level}, which measures the diversity of distinct concept instances in any category (e.g., dog, lion, tiger), and (ii) \catDivIcon~\emph{category-level}, which measures how broadly the generations span higher-level semantic categories within the concept (e.g., mammal, bird, reptile). 
For details on the entropy computation and the nuances of its normalization, we refer to Appendix~\ref{appsec:diversity_normalization}.\\
The final Diversity Score is the sum of the normalized~(Section~\ref{subsec:llm_judge_experiments}) entity- and category-level entropies, yielding a value in $[0,2]$. A high diversity score indicates that the recovered representation covers the concept broadly, rather than collapsing to a small set of entities or semantic subcategories. This is analogous to recall.

\maxBenchIcon~\textbf{\maxbench~Score.}
Following \textsc{AxBench}~\citep{wu2025axbench}, we also measure the fluency and instruction-following scores to validate whether steering does not cause degenerate outputs. The \textsc{AxBench} Score is then defined as the harmonic mean of the concept, fluency, and instruction-following scores. This captures whether an intervention successfully elicits the target concept while preserving fluent generation and adherence to the original instruction. We then define the \maxbench~score as the harmonic mean of the concept, fluency, instruction-following, \emph{and diversity} scores. The \maxbench~score rewards representations that both steer reliably toward the target concept and capture a broad range of values of that concept.

\begin{table}[t]
\caption{Scores on the test set for Gemma-3-1B. We report \emph{Concept}, \emph{Fluency}, \emph{Instruction}, and \emph{Diversity} scores, together with the composite \textsc{AxBench} and \maxbench~scores. Full metric definitions are given in Section~\ref{subsec:scoring}. For each concept and method, we select the rank that achieves the highest mean \maxbench~score. Each reported value is the mean~$\pm$~standard deviation over all the data splits.}
\centering
\resizebox{\textwidth}{!}{%
\begin{tabular}{lcccccc}
\toprule
\textbf{Method} & \textbf{Concept} & \textbf{Fluency} & \textbf{Instruction} & \textbf{Diversity} & \textbf{\textsc{AxBench}} & \textbf{\maxbench} \\
\addlinespace[1pt]
\textit{\footnotesize } & \conceptIcon & \fluencyIcon & \instructIcon & {\small\textcolor{divgreen}{\divIcon\,=\,\instDivIcon$+$\catDivIcon}} & {\small\textcolor{axblue}{\axBenchIcon\,=\,HM(\conceptIcon,\fluencyIcon,\instructIcon)}} & {\small\textcolor{maxamber}{\maxBenchIcon\,=\,HM(\conceptIcon,\fluencyIcon,\instructIcon,\divIcon)}} \\
\midrule
Prompting & $0.67 \pm 0.34$ & $\mathbf{2.00 \pm 0.01}$ & $1.65 \pm 0.05$ & $0.90 \pm 0.37$ & $0.61 \pm 0.32$ & $0.98 \pm 0.32$ \\
\midrule
Isotropic & $1.12 \pm 0.41$ & $1.80 \pm 0.14$ & $1.47 \pm 0.09$ & \underline{$1.53 \pm 0.27$} & $0.84 \pm 0.35$ & $1.39 \pm 0.29$ \\
Embedding  & $0.85 \pm 0.38$ & $1.65 \pm 0.33$ & \underline{$1.48 \pm 0.14$} & $1.27 \pm 0.41$ & $0.60 \pm 0.32$ & $1.15 \pm 0.39$ \\
\midrule
DiffMean-r1 & $0.63 \pm 0.20$ & $1.33 \pm 0.13$ & $1.42 \pm 0.06$ & $1.42 \pm 0.27$ & $0.41 \pm 0.17$ & $1.04 \pm 0.16$ \\
ReFT-r1 & $0.34 \pm 0.23$ & $1.42 \pm 0.07$ & $1.45 \pm 0.06$ & $1.14 \pm 0.42$ & $0.24 \pm 0.18$ & $0.68 \pm 0.33$ \\
\midrule
MFA & $1.37 \pm 0.48$ & $1.73 \pm 0.19$ & $1.40 \pm 0.18$ & $1.44 \pm 0.31$ & $0.91 \pm 0.32$ & $1.40 \pm 0.17$ \\
SAE (top-k) & $0.63 \pm 0.27$ & $1.44 \pm 0.28$ & $1.45 \pm 0.08$ & $1.18 \pm 0.27$ & $0.43 \pm 0.19$ & $0.99 \pm 0.24$ \\
\midrule
Factor Analysis & \underline{$1.64 \pm 0.26$} & $1.86 \pm 0.13$ & $1.41 \pm 0.11$ & $1.52 \pm 0.15$ & $1.19 \pm 0.24$ & \underline{$1.58 \pm 0.11$} \\
Schatten Probe & $1.55 \pm 0.26$ & $1.83 \pm 0.14$ & $1.42 \pm 0.10$ & $\mathbf{1.55 \pm 0.18}$ & $1.13 \pm 0.23$ & $1.56 \pm 0.12$ \\
Linear Probe & $1.50 \pm 0.31$ & $1.85 \pm 0.13$ & $1.44 \pm 0.10$ & $1.49 \pm 0.20$ & $1.12 \pm 0.26$ & $1.54 \pm 0.13$ \\
PCA & $1.59 \pm 0.29$ & \underline{$1.94 \pm 0.06$} & $1.44 \pm 0.11$ & $1.51 \pm 0.17$ & \underline{$1.21 \pm 0.24$} & \underline{$1.58 \pm 0.11$} \\
Diff Mean (Pairs) & $\mathbf{1.64 \pm 0.27}$ & $1.87 \pm 0.13$ & $1.43 \pm 0.10$ & $1.53 \pm 0.15$ & $\mathbf{1.22 \pm 0.25}$ & $\mathbf{1.59 \pm 0.11}$ \\

\bottomrule
\end{tabular}%
}
\label{tab:best_rank_gemma1b_v2}
\end{table}

\section{Experiments}

\subsection{Models}
\label{subsec:models}
We conduct experiments across three model families, including two model sizes from the Gemma family: Llama-3.1-8B~\citep{grattafiori2024llama3herdmodels}, Gemma-3-1B and Gemma-3-27B~\citep{gemmateam2025gemma3technicalreport}, and Qwen-3.5-4B~\citep{qwenteam2026qwen35omnitechnicalreport}. All models are instruction-tuned.

For each concept dataset, we extract hidden activations from later-to-middle layers, since semantic information is well formed at that stage of the model layers and it has been shown to work well in prior work~\citep{turner2025OneDimSteer3,rimsky-etal-2024-steering} for interventions. Specifically, we use layer $19$ of Llama-3.1-8B, layer $17$ of Gemma-3-1B, layer $53$ of Gemma-3-27B, and layer $27$ of Qwen-3.5-4B. We leave evaluating different layers using our sampling evaluation framework to future work.\footnote{Note that the choice of layer is also partially constrained by the availability of SAEs; we wish to compare methods applied at the same layer.}

\subsection{Dataset}
\label{subsec:concept_dataset}
We construct a dataset of six concepts: \dataset{Animals}, \dataset{Countries}, \dataset{Vehicles}, and \dataset{Plants} are hierarchical and unordered. \dataset{Days} is categorical and ordered and \dataset{Years} is continuous and ordered. Hierarchical concepts  are divided into categories. For example, \dataset{Animals} includes categories such as mammals and reptiles etc, with each category containing multiple entities. Mammal category, for instance, includes entities like \emph{lion}, \emph{beaver}, and \emph{horse}. For each entity, we use GPT-5~\cite{singh2025openaigpt5} to generate 100 sentences such that the sentence ends with the target entity. The sentences without the ground truth words at the end compose the inputs of our datasets. An example for the concept \dataset{countries} from the dataset is as follows

\begin{quote}
\textbf{Prompt:} \dta{The country that contains the Great Pyramid of Giza is} \tgt{Egypt}.
\end{quote}

We evaluate whether the models from Section~\ref{subsec:models} can perform the target-token prediction task, i.e., whether the intended concept word (e.g., \textit{Egypt} in the above example) is the most likely next token given the constructed prompt. Across all concepts, the correct entity appears among the model's top-5 predictions with at least 70\% accuracy. We also observe that larger models generally achieve higher prediction accuracy compared to smaller models. We provide detailed results for this prediction task in Appendix Table~\ref{tab:task-pref}. Examples of the dataset construction and additional details are provided in Appendix Table~\ref{tab:dataset-examples}\&~\ref{tab:concept-categories}.

\paragraph{Dataset Splits.} In general, we do not assume that a concept dataset contains all possible entities associated with that concept. For example, \dataset{Animals} may contain only a subset of all possible animals. We therefore evaluate each localization method across multiple dataset splits. This tests whether the method can recover a stable concept representation from different incomplete subsets of the data, without depending only on a few specific entities. For hierarchical concepts, we use two types of splits. In the first, we hold out an entire semantic category and recover the representation using the remaining categories. In the second, we retain all categories but randomly hold out individual entities across them. We generate 10 splits in total, with five splits of each type. For ordered concepts such as \dataset{Days} and \dataset{Years}, we instead randomly hold out a subset of concept values, such as particular days of the week or specific years.

\paragraph{Alpaca Eval.} We use prompts from AlpacaEval~\citep{dubois2024lengthcontrolled} as inputs to the language model to perform interventions. For each prompt, we intervene on the model's hidden activations using a sampled point from the recovered concept geometry and observe the resulting model output. We randomly select five validation prompts for hyperparameter tuning and evaluate on a disjoint set of five test prompts. For each prompt we sample 30 points from the recovered geometry and perform one intervention with each sampled point.

\paragraph{Ranks.} 

For methods that recover multi-dimensional subspaces, we evaluate ranks \(k\in\{1,2,4,8,16\}\). The precise meaning of \(k\) depends on the localization method. For PCA, \(k\) is the number of leading singular directions retained in \(\mL\). For factor analysis, it is the number of latent factors. For MFA, \(k\) similarly denotes the number of latent factors per mixture component, and we train a separate model for each rank. For DiffMean (pairs) and Linear Probe, \(k\) is the number of leading singular directions retained in \(\mL\). For SAE, \(k\) is the number of selected SAE features whose decoder directions form the basis \(\mL\). For Schatten Probe, we define the rank using the effective rank~\citep{effective_rank} of the learned classifier weight matrix. DiffMean-r1 and ReFT-r1 recover a single direction and therefore always have $k=1$.

\begin{figure}[htbp]
    \centering
    \includegraphics[width=\textwidth]{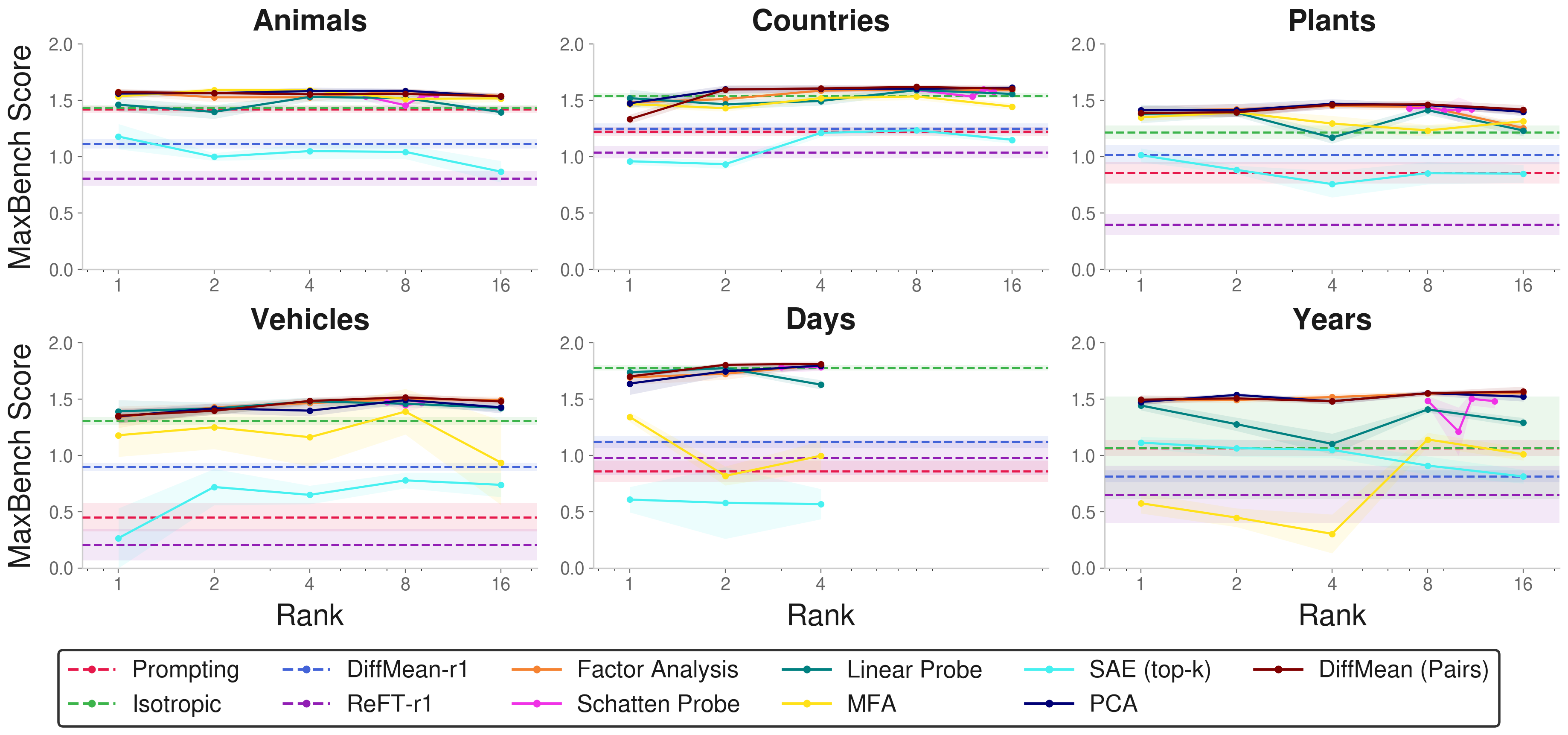}
    \caption{\maxbench~score by rank for Gemma-3-1B for all concepts. The x-axis is
rank (days is capped at rank $4$, as \textsc{Days} has 7 values, and we assume the rank cannot exceed the number of concept values); the y-axis is \maxbench~score; confidence intervals are computed via dataset split with 10 samples. Each point is the mean \maxbench~score at that (concept, method, rank); the shaded band is std.\ dev. Prompting, Isotropic, DiffMean-r1, and ReFT-r1 have no rank hyperparameter, so they are drawn as dashed lines. For Schatten Probes, we measure \emph{effective} rank instead of manually setting its rank; we round to the nearest integer.
}
    \label{fig:best-variant-per-method}
\end{figure}

\subsection{LLM Judge}
\label{subsec:llm_judge_experiments}
Given this set of steered outputs, we follow ~\cite{wu2025axbench} by using large language model (LLM) judges to evaluate each steered output. We evaluate each method using the three metrics from \textsc{AxBench}: \emph{concept}, \emph{fluency}, and \emph{instruction following}, together with the \emph{instance diversity} and \emph{category diversity} components of our proposed \emph{diversity} score.  For the \textsc{AxBench} metrics, the judge assigns each generated response a score from 0 to 2 for all three metrics. We use open-source model, Qwen-3.6-27B (instruction-tuned) as our LLM judge. All prompts we use for LLM judge are in Appendix Section~\ref{app:llm_judge_prompts}.

To evaluate diversity, we use an LLM to extract all concept instances mentioned across the steered outputs and, where applicable, assign each instance to a semantic category. For concepts that contain categories (\textsc{Animals}, \textsc{Countries}, \textsc{Plants}, \textsc{Vehicles}), we provide the judge with the category set used in our dataset (Appendix Table~\ref{tab:concept-categories}). Instances that belong to the concept but that did not occur in the training and do not belong to a predefined category are assigned to an "other" category (e.g. dragon). We then compute the normalized Shannon entropy of the resulting instance and category distributions, yielding the instance diversity and category diversity scores, respectively (see Appendix~. For concepts \textsc{Days} and \textsc{Years} that do not contain categories we treat each instance as its own category.

\section{Results}
\begin{wraptable}[15]{r}{0.6\textwidth}
\vspace{-1em}
  \centering
  \setlength{\tabcolsep}{2pt}
  \renewcommand{\arraystretch}{0.95}
  \scriptsize
  \caption{\maxbench~score on the \dataset{days} concept, across all four models. Per column, the highest mean is bold and the second highest is underlined.}
  \label{tab:days_manifold}
  \resizebox{0.53\textwidth}{!}{%
  \begin{tabular}{lcccc}
    \toprule
     & Gemma-3-1B & Gemma-3-27B & Llama-3.1-8B & Qwen-3.5-4B \\
    \midrule
    Prompting & $0.86 \pm 0.10$ & $\underline{1.87 \pm 0.01}$ & $\mathbf{1.86 \pm 0.02}$ & $\mathbf{1.80 \pm 0.02}$ \\
    \midrule
    Diff Mean-r1 & $1.12 \pm 0.05$ & $1.33 \pm 0.07$ & $1.39 \pm 0.09$ & $0.55 \pm 0.10$ \\
    ReFT-r1 & $0.98 \pm 0.15$ & $1.33 \pm 0.04$ & $1.15 \pm 0.12$ & $0.73 \pm 0.11$ \\
    \midrule
    MFA & $1.34 \pm 0.04$ & $1.51 \pm 0.03$ & $0.32 \pm 0.13$ & $1.55 \pm 0.04$ \\
    SAE (top-k) & $0.61 \pm 0.11$ & $1.31 \pm 0.07$ & $1.40 \pm 0.05$ & $0.54 \pm 0.09$ \\
    \midrule
    Factor Analysis & $1.80 \pm 0.02$ & $1.84 \pm 0.01$ & $1.80 \pm 0.01$ & $1.72 \pm 0.03$ \\
    Schatten Probe & $1.78 \pm 0.02$ & $1.77 \pm 0.02$ & $1.75 \pm 0.01$ & $1.70 \pm 0.01$ \\
    Linear Probe & $1.77 \pm 0.03$ & $1.79 \pm 0.01$ & $1.80 \pm 0.02$ & $1.71 \pm 0.03$ \\
    PCA & $1.80 \pm 0.03$ & $1.79 \pm 0.01$ & $1.78 \pm 0.02$ & $1.70 \pm 0.02$ \\
    Diff Mean (Pairs) & $\underline{1.81 \pm 0.02}$ & $1.76 \pm 0.03$ & $\underline{1.81 \pm 0.01}$ & $1.71 \pm 0.02$ \\
    \midrule
    Spline (missing days) & $1.73 \pm 0.01$ & $1.84 \pm 0.02$ & $1.69 \pm 0.06$ & $1.64 \pm 0.03$ \\
    Spline  & $\mathbf{1.83 \pm 0.07}$ & $\mathbf{1.93 \pm 0.01}$ & $1.71 \pm 0.05$ & $\underline{1.73 \pm 0.03}$ \\
    \bottomrule
  \end{tabular}%
  }
\end{wraptable}

\subsection{Linear Methods}
\paragraph{Directions vs. Subspaces.} 

We first ask whether multinomial concepts are better represented by a single direction or by a higher-dimensional subspace. Table~\ref{tab:best_rank_gemma1b_v2} summarizes the performance of the different concept-localization methods. Multidimensional methods consistently achieve higher \maxbench~scores than one-dimensional methods across concepts. This suggests that concepts with categorical non-binary structure are not well captured by a single direction, and instead benefit from subspace-based representations. 

We next see how each method performs across ranks. Figure~\ref{fig:best-variant-per-method} shows the results for Gemma-3-1B across concepts. For the affine methods, \maxbench{} scores are generally stable over the range of ranks we evaluate, with slight improvement as rank increases for some concepts. This suggests that much of the benefit of these methods can be obtained with relatively low-dimensional subspaces, provided that the representation is centered at an appropriate offset. Results for the other models are provided in Appendix Figures~\ref{fig:rank_x_maxbench_gemma27b},~\ref{fig:rank_x_maxbench_llama8b}, and~\ref{fig:rank_x_maxbench_qwen4b}.

\begin{figure}[thbp]
    \centering
    \includegraphics[width=\textwidth]{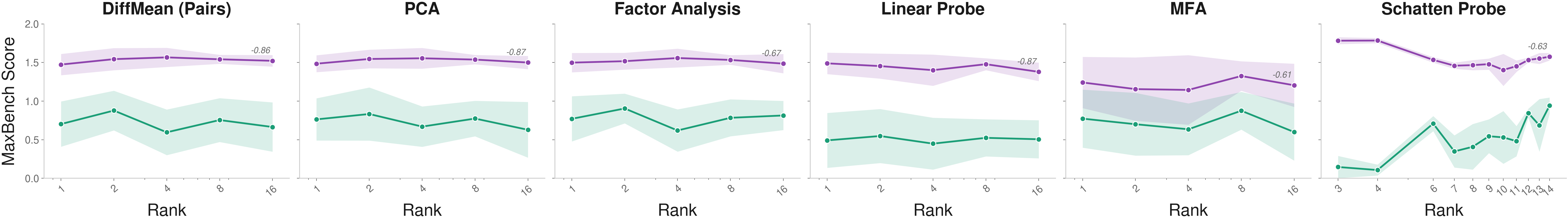}
    \caption{Comparison of multinomial concept localization methods for Gemma-3-1B with and without an offset (\textcolor{affinecol}{affine} and \textcolor{linearcol}{linear}, respectively). The x-axis shows the rank hyperparameter for each method, and the y-axis reports the~\maxbench~score. For the Schatten probe, we plot the effective rank of the learned weight matrix as the Schatten $p$-norm parameter is varied from $0.1$ to $0.9$. For each method, we average the scores across all concepts and compute std.\ dev.\ via bootstrap. Sampling around the concept mean offset is consistently significantly better than assuming the subspace is centered at the origin. We observe similar trends for other models(Appendix Figures~\ref{fig:linear-vs-affine_gemma27b},~\ref{fig:linear-vs-affine_llama},~\ref{fig:linear-vs-affine_qwen}).}
    \label{fig:linear-vs-affine}
\end{figure}

\paragraph{Linear vs. Affine.} We then ask how much of each method's performance is attributable to the offset $\vb$ of the affine subspace, rather than the choice of bases $\mL$.  In Figure~\ref{fig:linear-vs-affine}, for Gemma-3-1B, we observe that affine variants (with offset $\vb$ equal to the mean of concept-related points) achieve higher \maxbench~scores than their linear counterparts (with $\vb = \mathbf{0}$). This holds true for all the other models as well as shown in Appendix Figures~\ref{fig:linear-vs-affine_gemma27b},~\ref{fig:linear-vs-affine_llama},~\ref{fig:linear-vs-affine_qwen}. To assess $\mL$'s contribution,
% Finally, we ask whether samples must lie within the recovered low-dimensional affine subspace at all, or whether isotopically sampling around the concept's offset is sufficient. To test this,
we include an isotropic baseline that samples points isotropically around the concept centroid without constraining them to the span of the recovered basis vectors. Interestingly, for Gemma-3-1B, the isotropic baseline outperforms various linear and one-dimensional methods, and underperforms affine methods by a relatively small margin. Thus, both the offset and bases contribute significantly to effective concept localization; that said, assuming our sampling procedure, the choice of where to center one's samples may be more important than the directions in which one moves around that centroid. 

For Gemma-3-1B, this suggests that correctly locating the center of the concept geometry may be at least as important as recovering its specific basis directions, and may provide indirect support for a perspective of concepts as prototypes \citep{fel2026rabbithull} over concepts as directions \citep{park2023LRH}. 
The isotropic baseline performs substantially worse for the other models, suggesting that the centroid alone maybe insufficient and that the recovered basis directions also play an important role.

\paragraph{Prompting.} We include a prompting baseline, in which we explicitly instruct the model to answer each AlpacaEval prompt while incorporating the target concept. Examples of these prompts are provided in Appendix~\ref{app:prompting-baseline}. This baseline measures how well the target concept can be elicited through prompting alone, without intervening on the model's hidden activations. While prompting is not as competitive for smaller models such as Gemma-3-1B (Table~\ref{tab:best_rank_gemma1b_v2}), it outperforms all methods on larger models. Results are reported in Appendix~\ref{tab:best_rank_gemma1b_app}, ~\ref{tab:best_rank_llama31_app}, ~\ref{tab:best_rank_qwen35_app}.

\setlength{\intextsep}{1pt}
\begin{wrapfigure}{5}{0.45\textwidth}
    \centering
    \vspace{-1em}
    \includegraphics[width=0.40\textwidth]{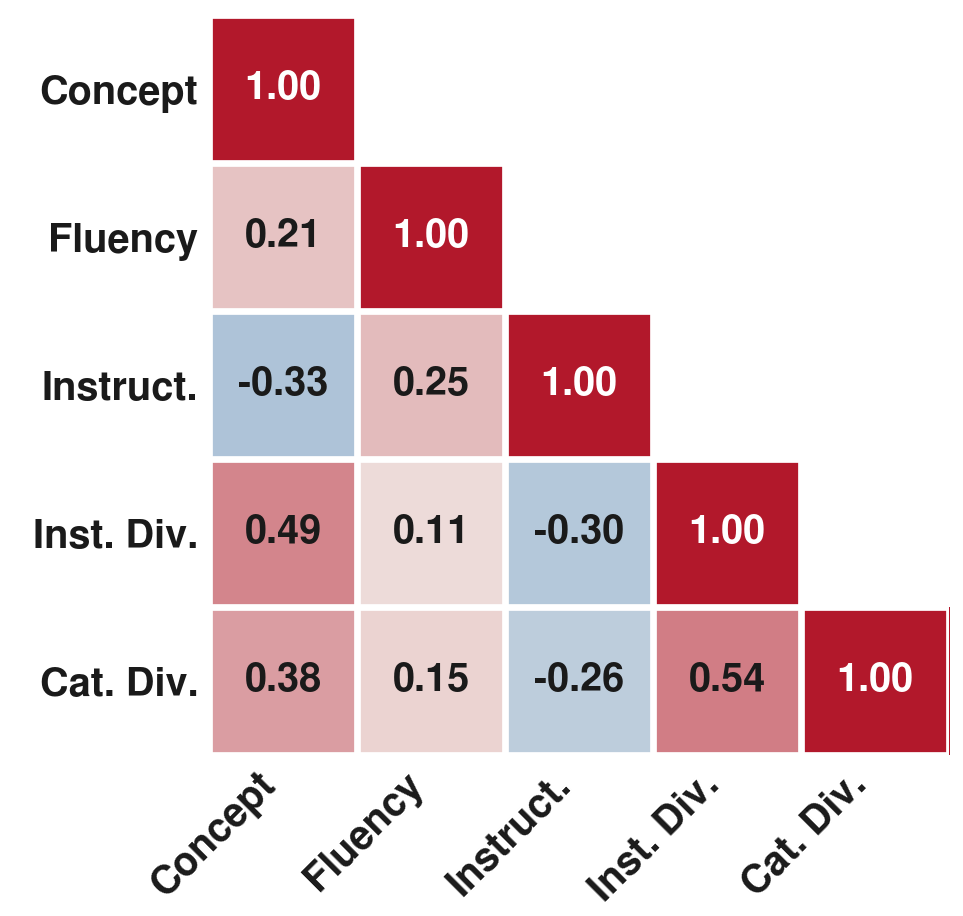}
    \caption{Spearman correlation heatmaps for all the metrics used in our evaluation. This is computed over all steered outputs across models, methods, concepts, ranks, and splits. A more detailed analysis is provided in Appendix Figure~\ref{appfig:metrics_corr}.}
    \label{fig:metrics_corr}
    \vspace{-3em}
\end{wrapfigure}
\paragraph{Embedding baseline.} We also include a embedding baseline to show the performance of different methods when no context is provided. Representations are extracted directly from the embeddings of the concept's instance tokens. This allows us to evaluate how much contextual information contributes to recovering a concept representation. Across models, we observe better performance when using representations from intermediate layers. This suggests that steering is operating over a space that contains more than the concept values' token identities.

\subsection{Non-Linear Methods}
We include spline fitting as an additional method for the \dataset{Days} concept. Because this method requires an ordered set of concept values and (in its standard form) access to the complete sequence, we restrict its main evaluation to \dataset{Days}. We also evaluate spline fitting under incomplete observations by withholding selected days before fitting the spline, making this setting more comparable to the other concept localization methods, which operate with only a subset of concept instances. Results for \dataset{Days} are summarized in Table~\ref{tab:days_manifold}. Within this concept, spline fitting using the complete set of concept values outperforms spline fitting when some values are held out. More broadly, spline fitting achieves the highest overall performance among the methods evaluated on \dataset{Days} if the complete set of concept examples are provided.

\subsection{Generalization beyond concept data}
How well do the recovered concept representations generalize beyond the concept instances used to derive them? We provide the LLM judge with all concept instances and taxonomy labels observed in the training set (from the respective split) and ask it to classify each generated concept instance as either \emph{seen}, if it appears in the training data, or \emph{unseen}, otherwise. We count the total number of unseen instances and report results in Table~\ref{tab:best_rank_ood_concept_gemma1b}. We find that many methods often generate unseen concept instances, indicating that the recovered representations can generalize beyond the examples provided during training rather than merely reproducing observed instances. Notably, the unsupervised methods, particularly Factor Analysis and MFA, produce the largest number of unseen instances. We provide additional results for this analysis in Appendix Table~\ref{tab:ood_max_by_method}.

\subsection{Metric Meta-evaluation}
Does each metric (including the proposed diversity score) measure truly independent factors of performance, or do some factors strongly covary? To investigate, we compute Spearman correlations across all five metrics to assess how independent they are. We find (Figure~\ref{fig:metrics_corr}) that the concept score is somewhat correlated with instance and category diversity. This is expected, as the diversity score can only be maximized when all generated outputs contain concept-related values. The concept score and instruction score are negatively correlated, suggesting that when the target concept is injected more strongly, it can become harder for the model to stay on topic with respect to the original instruction. Fluency scores have relatively weak correlations with other metrics.

\section{Related Work}

\begin{wraptable}[17]{r}{0.53\textwidth}
  \caption{Per-concept unseen instance count for Gemma-3-1B, at each (method, concept)'s own best rank (the rank with the highest mean \maxbench~score across bootstraps). Mean $\pm$ std is over bootstrap draws at that rank. All results are in Appendix Table~\ref{ood-example}.}
  \centering
    \scriptsize
    \setlength{\tabcolsep}{5pt}
    \renewcommand{\arraystretch}{0.95}
  \begin{tabular}{lcccc}
    \toprule
    Method & Animals & Countries & Plants & Vehicles \\
    \midrule
    Diff Mean (R1) & 41 $\pm$ 3 & 11 $\pm$ 3 & 49 $\pm$ 10 & 29 $\pm$ 8 \\
    ReFT R1 & 6 $\pm$ 8 & 4 $\pm$ 2 & 11 $\pm$ 6 & 4 $\pm$ 3 \\
    \midrule
    MFA & 61 $\pm$ 12 & \textbf{65} $\pm$ 8 & 46 $\pm$ 7 & 35 $\pm$ 15 \\
    SAE (top-k) & 36 $\pm$ 8 & 24 $\pm$ 9 & 59 $\pm$ 8 & 22 $\pm$ 8 \\
    \midrule
    Factor Analysis & \textbf{72} $\pm$ 7 & 26 $\pm$ 6 & 62 $\pm$ 22 & \textbf{76} $\pm$ 5 \\
    Schatten Probe & 60 $\pm$ 19 & 26 $\pm$ 7 & \textbf{67} $\pm$ 16 & 55 $\pm$ 12 \\
    Linear Probe & 66 $\pm$ 7 & 28 $\pm$ 8 & 63 $\pm$ 9 & 61 $\pm$ 8 \\
    PCA & 64 $\pm$ 7 & 26 $\pm$ 6 & 56 $\pm$ 5 & 37 $\pm$ 8 \\
    Diff Mean (Pairs) & 72 $\pm$ 6 & 36 $\pm$ 7 & 66 $\pm$ 9 & 73 $\pm$ 7 \\
    \bottomrule
  \end{tabular}
  \label{tab:best_rank_ood_concept_gemma1b}
\end{wraptable}

\textbf{Concept geometry.} 
Early concept geometry studies operated under the Linear Representation Hypothesis \citep{park2023LRH}, which holds that concepts are linear (and typically one-dimensional) subspaces of LM activations. More recent work has proposed that concepts may be multidimensional~\citep{gurnee2025geometryofcounting,engels2025not,park2025hierarchical,park2026informationgeometry,vielhaben2023MCD}. This structure can be non-linear, and includes geometries such as convex hulls \citep{fel2026rabbithull}, polytopes \citep{park2025hierarchical}, vector magnitudes (rather than directions; \citealp{csordas2024recurrent}), dense features~\citep{sun2025dense,lubana2026priors} or manifolds \citep{PhysRevX.8.031003,chung2021,modell2025originsrepresentationmanifoldslarge,kantamneni2025languagemodelsusetrigonometry,engels2025not,gurnee2025geometryofcounting,yocum2025neuralmanifoldgeometry,karkada2026symmetry,feucht2026arithmetic,wurgaft2026manifoldsteeringrevealsshared,bhalla2026SAEmanifold,sarfati2026shapebeliefs,fel2026structuringsparsityblocksparsefeaturizers}. We focus extensively on affine subspaces, as they lie at an intermediate point of complexity: they enable one to go beyond single dimensions and beyond subspaces centered at the origin, but retain the simplicity and tractability of linear methods.

\textbf{Evaluating concept representations.}~Given a concept representation, how can one verify its quality? Prior work has proposed correlational and causal metrics for evaluating their quality. Some evaluate geometric goodness-of-fit \citep{tiblias2025SMDS}, whereas others evaluate counterfactual interventions on downstream behavior \citep{chen-etal-2025-steer,faithsteer,engels2025not,mueller2025mib,shafran2026MFA,wu2025axbench,lee2026decomposingQK}. Others still use toy settings with known ground truths \citep{gupta2024interpbench}. These works evaluate whether a concept was leveraged after steering, but not whether the full concept space was covered; our diversity score fills this gap. These works also focus largely on binary concepts; our work specifically focuses on multinomial concepts with varied internal structure.

\textbf{Multidimensional/Multinomial concept steering.} 
Most steering approaches assume that concepts correspond to single directions in activation space, and thus intervene by adding a scaled steering vector to the hidden state~\citep{li2023OneDimSteer1,zou2025OneDimSteer2,turner2025OneDimSteer3,subramani2022OneDimSteer4,marks2024geometrytruthemergentlinear,you2026spherical}.
\citet{engels2025not} instead steer by replacing the angular coordinate corresponding to one value (e.g., Monday) with another in a circular latent space, and then reconstruct the activation; this alters the model’s predictions on periodic variables like days of the week. \citet{shafran2026MFA} propose steering based on a Mixture of Factor Analyzers (MFA) decomposition of activation space, which models activations as belonging to local Gaussian regions with associated low-dimensional subspaces. This representation enables two forms of intervention: steering activations toward a component centroid to move the model toward a concept region, and modifying the coordinates within the component’s local subspace to control finer semantic variations.
% There are no existing methods 
Our work builds on this growing recognition that concepts in language models often exhibit intrinsically multi-dimensional structure. We propose a new evaluation framework to evaluate steering on multinomial concepts, which has only preliminarily been explored in prior work (although see concurrent work by \citealp{wurgaft2026manifoldsteeringrevealsshared}, whose method we include in our evaluation).

\section{Conclusion}
In this work, we propose a sampling-based steering framework for evaluating multinomial concepts. Based on this framework, we propose \maxbench~to benchmark multinomial and multidimensional concept localization methods. Using \maxbench, we observe that (i) affine subspaces more reliably recover such concepts than rank-one or linear subspaces; (ii) much of this advantage is due to better non-zero offsets rather than the choice of bases; (iii) manifold steering is competitive with the best methods when applicable; and (iv) no method consistently outperforms prompting when models are sufficiently capable. These findings highlight how interpretability research may benefit from studying a broader range of concepts, and from developing methods that better capture their underlying geometry.

\section{Limitations and Future Work}

Most of our experiments rely on LLM judges, which do not always agree with human judgments. Second, our experiments are limited to a single intervention layer. Our sampling procedure also relies on tuning hyperparameters for reliable steering and future work could explore better ways to sample from geometries without relying on hyperparameters. More broadly, although our evaluation framework is designed to be geometry-agnostic, current methods we benchmark are largely limited to affine representations. Future work should extend \maxbench~with a wider range of non-linear geometric representations. 

\section*{Acknowledgments}
We thank Yukyung Lee, Gabriel Franco, Kevin Quinn, Lucas Tassis, Themistoklis Nikas, Zhengyang Shan, Micah Benson, John Seon Keun Yi, Angelos Poulis, Mark Crovella, Evimaria Terzi and BAAIGL group at Boston University for constructive feedback. This material is supported by the National Science Foundation under Grant No.\ 2530728 and the U.S.-Israel Binational Science Foundation (BSF) under Grant No.\ 2025670.
This research was partly supported by the Israel Science Foundation (grant No.\ 2942/25) and the European Union (ERC, Control-LM, 101165402). Views and opinions expressed are however those of the author(s) only and do not necessarily reflect those of the National Science Foundation, U.S.-Israel Binational Science Foundation, European Union, or the European Research Council Executive Agency. Neither the European Union nor the granting authorities can be held responsible for them. 
FB was supported by the Swiss National Science Foundation (SNSF) under Grant 239054.

\bibliography{references}
\bibliographystyle{conference}

\newpage
\appendix
\appendix

\section{Datasets}

We generate 6 concept datasets and 1 control dataset. Some example prompts from our dataset are presented in Table~\ref{tab:dataset-examples} and different categories and entities per concept are in Table~\ref{tab:concept-categories}. Categories for \dataset{Animals} are derived from
WordNet~\citep{miller1992wordnet}, plants are classified according to their growth habits\footnote{Source USDA website:
\url{https://plants.sc.egov.usda.gov/}}, while categories
for the remaining two concepts are grouped manually using
GPT-5~\citep{singh2025openaigpt5}.
\vspace{0.5cm}
\begin{table}[h]
\centering
\caption{Example prompts for each concept}
\label{tab:dataset-examples}
\small
\setlength{\tabcolsep}{6pt}
\renewcommand{\arraystretch}{1.3}
\begin{tabularx}{\textwidth}{l X}
\toprule
Concept & \dta{Prompt} \tgt{Correct Continuation} \\
\midrule
\dataset{Animals}   & \dta{The big cat of genus \textit{Panthera} referred to as the king of the beasts due to its mane} \dta{and dominant pride structure is called a} \tgt{lion}. \\
\dataset{Countries} & \dta{The country that contains the Great Pyramid of Giza is} \tgt{Egypt}. \\
\dataset{Days}      & \dta{The day immediately following the weekend in many calendars is} \tgt{Monday}. \\
\dataset{Plants}    & \dta{The crop most associated with Popeye gaining strength is} \tgt{spinach}. \\
\dataset{Vehicles}  & \dta{The aircraft that remains airborne through the rotation of large overhead blades is called a} \tgt{helicopter}. \\
\dataset{Years}     & \dta{The year when the COVID-19 pandemic spread across the world was} \tgt{2020}. \\
\dataset{Random}    & \dta{If several ice cubes are left on a kitchen counter until they completely melt, they become} \tgt{water}. \\
\bottomrule
\end{tabularx}
\end{table}

\paragraph{Concept Prediction Task Performance}
\label{subsec:task-performance}

\begin{table}[h]
\centering
\caption{For each concept, we define a set of categories.}
\label{tab:concept-categories}
\small
\setlength{\tabcolsep}{6pt}
\begin{tabular}{lll}
\toprule
Concept & Category & Entities \\
\midrule
\multirow{6}{*}{\dataset{Animals}}
 & Mammals    & beaver, lion, giraffe, horse, koala \\
 & Reptiles   & lizard, turtle, dinosaur, crocodile, gecko \\
 & Birds      & parrot, owl, duck, eagle, penguin \\
 & Amphibians & frog, salamander, toad, axolotl, newt \\
 & Fish       & salmon, shark, pufferfish, clownfish, anglerfish \\
 & Insects    & firefly, dragonfly, butterfly, bee, grasshopper \\
\cmidrule(lr){1-3}
\multirow{6}{*}{\dataset{Countries}}
 & Africa        & Egypt, Morocco, Kenya, Ethiopia, Madagascar \\
 & Asia          & China, India, Japan, Thailand, Vietnam \\
 & Europe        & Germany, France, Italy, Spain, Portugal \\
 & North America & Canada, Mexico, Cuba, Jamaica, United States \\
 & South America & Brazil, Argentina, Chile, Colombia, Peru \\
 & Oceania       & Australia, Fiji, Samoa, Tonga, New Zealand \\
\cmidrule(lr){1-3}
\multirow{4}{*}{\dataset{Vehicles}}
 & Road  & car, bus, truck, bicycle, ambulance \\
 & Rail  & train, subway, funicular, monorail, tram \\
 & Air   & airplane, helicopter, drone, glider, rocket \\
 & Water & canoe, submarine, yacht, hovercraft, ferry, cruise \\
\cmidrule(lr){1-3}
\multirow{5}{*}{\dataset{Plants}}
 & Trees    & oak, pine, maple, pineapple, banana \\
 & Shrubs   & rose, lavender, lilac, boxwood, jasmine \\
 & Herbs    & basil, dill, thyme, mint, spinach \\
 & Climbers & grape, passionflower, wisteria, cucumber, watermelon \\
 & Creepers & ivy, pumpkin, strawberry, sweet potato \\
\midrule\midrule
 Concept & Entities & \\
\cmidrule(lr){1-3}
\dataset{Days} & \multicolumn{2}{l}{Monday, Tuesday, Wednesday, Thursday, Friday, Saturday, Sunday} \\
\cmidrule(lr){1-3}
\multirow{2}{*}{\dataset{Years}} & \multicolumn{2}{l}{1066, 1215, 1347, 1453, 1492, 1517, 1588, 1607, 1620, 1666, 1687, 1776,} \\
 & \multicolumn{2}{l}{1789, 1804, 1815, 1861, 1865, 1903, 1914, 1918, 1929, 1939, 1945, 1963,} \\
\cmidrule(lr){1-3}
\multirow{2}{*}{\dataset{Random}} & \multicolumn{2}{l}{water, dog, coffee, bird, king, rain, sky, door, bus, train, window,} \\
 & \multicolumn{2}{l}{tree, house, road, book, chair, computer, city, apple, bread} \\
\bottomrule
\end{tabular}
\end{table}

We report performance on the task of predicting the correct concept word in Table~\ref{tab:task-pref}. 
Here, we evaluate whether the models can reliably predict the target concept entities using our prompts. For example, for the concept \dataset{Countries}, a prompt may describe a country such that the expected answer is \emph{United States}. 

First, we evaluate all four models on the six concept datasets as well as the random control dataset. We report Top-$1$, Top-$5$, and continuation accuracy, averaged over all prompts and entities within each concept. Top-$1$ and Top-$5$ accuracy measure whether the beginning of the target entity appears among the model's highest-probability next-token predictions, while continuation accuracy measures whether the full target entity appears within the model's greedy continuation of up to five tokens. For example, if the target entity is \emph{United States}, Top-$1$ is correct if \emph{United} is the highest-probability next token, Top-$5$ is correct if \emph{United} appears among the five highest-probability next tokens, and continuation accuracy is correct if \emph{United States} appears in the generated continuation. Results are reported in Table~\ref{tab:task-pref}. Overall, all models achieve above $70$\% Top-$5$ accuracy across the concept datasets. A high Top-$5$ accuracy suggests that the hidden activation immediately preceding the concept word generally contains information about that word.

\vspace{0.5cm}
\begin{table}[h]
\centering
\caption{\textbf{Task performance across concepts and models.} Accuracy measures whether the model predicts the target concept entity for each prompt. \emph{Top-1} reports whether the target entity begins with the model's highest-probability next token, while \emph{Top-5} reports whether it begins with any of the five highest-probability next tokens. This prefix matching accounts for entities that may span multiple tokens. For example, if the target entity is \emph{United States}, a prediction of the token \emph{United} counts as a correct Top-1 match. \emph{Cont} reports whether the complete target entity appears anywhere in the model's greedy continuation of up to five tokens. For example, if the target entity is \emph{Egypt} and the prompt is ``The country that contains the Great Pyramid of Giza is'', Cont is correct if \emph{Egypt} appears as the continuation. Results are averaged over the entire dataset for each concept.}
\label{tab:task-pref}
\small
\setlength{\tabcolsep}{4pt}
\begin{tabular}{llccccccc}
\toprule
Model & Metric & Animals & Vehicles & Countries & Days & Plants & Years & Random \\
\midrule
\multirow{3}{*}{Gemma-3-1B}
 & Top-1 & 0.465 & 0.458 & 0.777 & 0.546 & 0.403 & 0.733 & 0.684 \\
 & Top-5 & 0.707 & 0.701 & 0.892 & 0.922 & 0.668 & 0.942 & 0.892 \\
 & Cont  & 0.444 & 0.435 & 0.772 & 0.524 & 0.331 & 0.375 & 0.704 \\
\cmidrule(lr){1-9}
\multirow{3}{*}{Gemma-3-27B}
 & Top-1 & 0.821 & 0.875 & 0.920 & 0.959 & 0.819 & 0.948 & 0.868 \\
 & Top-5 & 0.942 & 0.958 & 0.963 & 0.999 & 0.962 & 0.999 & 0.968 \\
 & Cont  & 0.900 & 0.872 & 0.920 & 0.959 & 0.834 & 0.900 & 0.879 \\
\cmidrule(lr){1-9}
\multirow{3}{*}{Llama-3.1-8B}
 & Top-1 & 0.686 & 0.782 & 0.890 & 0.931 & 0.703 & 0.885 & 0.727 \\
 & Top-5 & 0.896 & 0.931 & 0.966 & 1.000 & 0.918 & 0.966 & 0.933 \\
 & Cont  & 0.705 & 0.786 & 0.888 & 0.931 & 0.649 & 0.859 & 0.746 \\
\cmidrule(lr){1-9}
\multirow{3}{*}{Qwen-3.5-4B}
 & Top-1 & 0.735 & 0.744 & 0.881 & 0.953 & 0.638 & 0.991 & 0.825 \\
 & Top-5 & 0.905 & 0.941 & 0.938 & 0.999 & 0.868 & 1.000 & 0.958 \\
 & Cont  & 0.764 & 0.719 & 0.874 & 0.942 & 0.585 & 0.929 & 0.829 \\
\bottomrule
\end{tabular}
\end{table}

\section{Concept Localization Methods}
\label{appsec: method_all_details}
\paragraph{DiffMean-r1}\citep{marks2024geometrytruthemergentlinear}. 
DiffMean-r1 (commonly known as \textit{difference in means} in literature, but we call it DiffMean-r1 to differentiate it from our \emph{multidimensional} difference in means method) learns a concept representation from binary concept labels by computing the difference between the mean activations of the positive and negative examples. Let $\mX^{+}$ and $\mX^{-}$ denote the activation matrices corresponding to the positive and negative concept examples, respectively. The localized concept direction is given by $\sG_{\mathrm{DiffMean}} = \boldsymbol{\mu}^{+} - \boldsymbol{\mu}^{-}$, where $\boldsymbol{\mu}^{+} = \frac{1}{|\mX^{+}|}\sum_{\vx \in \mX^{+}} \vx$ and $\boldsymbol{\mu}^{-} = \frac{1}{|\mX^{-}|}\sum_{\vx \in \mX^{-}} \vx$. The localized representation is therefore a one-dimensional affine subspace of the form $\sG_{\mathrm{DiffMean}} = {\mL q + \vb \mid q \in \R}$, where  $\vb=\vzero$, and $\mL$ is normalized to unit norm.

\paragraph{ReFT-r1}\citep{wu2025axbench,wu2024reft}:
ReFT-r1 learns concept representation from supervised concept labels by jointly optimizing concept detection and representation-level steering. During training, it learns a single 1D subspace $\sG_{\mathrm{ReFT}} \in \R^d$ that is used both to detect the presence of the concept and to intervene on the hidden representations. Complete training objective is present in~\citep{wu2025axbench,wu2024reft}. Similar to DiffMean-r1 localized representation is therefore a one-dimensional direction of the form $\sG_{\mathrm{ReFT}}={\mL q+\vb \mid q\in\R}$, where $\vb=\vzero$ and $\mL=\sG_{\mathrm{ReFT}}$ is normalized to unit norm. Steering for ReFT-r1 is also similar to DiffMean-r1. 

Note that the concept dataset used for REFT-r1 differs slightly from those used by the other methods. REFT-r1 requires paired training examples of two types: responses that answer an instruction while incorporating the target concept, and responses to the instruction that do not incorporate the concept. The representation is then learned from this dataset. We generate this dataset using GPT-5~\cite{singh2025openaigpt5}. For example, for the concept \dataset{animals}, given the instruction \textit{`Write one sentence about a busy city''}, a positive response could be \textit{`The busy city streets were filled with people, cars, and pigeons circling above the buildings,''} while a negative response could be \textit{``The busy city streets were filled with people and cars moving between tall buildings.''}

\paragraph{Linear Probe}\citep{alain2016understanding}:
Linear Probe learns a linear classifier to predict concept labels from hidden activations. Given activation vectors $\vx^{(i)}$ and corresponding concept labels $y^{(i)} \in {1,\ldots,|C|}$, it learns classifier parameters $\mW \in \R^{d\times |C|}$ and $\vb \in \R^{|C|}$ by minimizing the cross-entropy loss. We then perform the singular value decomposition $\mW=\mU\mSigma\mV^\top$ and localize the concept basis as the top $k$ left singular vectors, $\mL=[\vu_1|\cdots|\vu_k]\in\R^{d\times k}$. The learned classifier bias $\vb$ cannot directly serve as the affine offset, since it lies in the $|C|$-dimensional classifier output space rather than the $d$-dimensional residual-stream space. We therefore set the affine offset to the global activation mean, $\vmu_C=\frac{1}{m}\sum_{i=1}^{m}\vx^{(i)}$. The localized representation is therefore a $k$-dimensional affine subspace of the form $\sG_{\mathrm{Probe}}={\mL\vq+\vb \mid \vq\in\R^k}$, where $\vb=\vmu_C$.

We treat the localized representation as an affine subspace and sample intervention points using the Gaussian sampling procedure described in Section~\ref{subsec:sampling}.

\paragraph{PCA}\citep{pearson1901lines}:
Given activation vectors $\{\vx^{(i)}\}_{i=1}^{m}$, we first compute the mean activation $\vmu_C=\frac{1}{m}\sum_{i=1}^{m}\vx^{(i)}$ and center the activations. We then perform singular value decomposition of the centered activation matrix (stack of $\{\vx^{(i)}\}_{i=1}^{m}$), $\tilde{\mX}=\mU\mSigma\mV^\top$, and retain the first $k$ right singular vectors to form the basis $\mL=[\vv_1|\cdots|\vv_k]\in\R^{d\times k}$. The resulting concept representation is a $k$-dimensional affine subspace centered at the mean activation, $\sG_{\mathrm{PCA}}={\mL\vq+\vmu_C \mid \vq\in\R^k}$. 
%Similar to probe we therefore sample coefficients $\vq^{(i)}\sim\mathcal{N}(\mathbf{0},\sigma^2\mI_k)$ and construct intervention points as $\vs^{(i)}=\mL\vq^{(i)}+\vmu_C$, following the affine subspace sampling procedure described above.
\paragraph{Factor Analysis}\citep{spearman1961general}:
Factor Analysis models the observed activations as arising from a low-dimensional latent representation corrupted by gaussian noise. Specifically, activations are modeled as $\vx^{(i)}=\vmu+\mL\vz^{(i)}+\rvepsilon^{(i)}$, where $\vmu\in\R^d$ is the mean activation, $\mL\in\R^{d\times k}$ is the factor loading matrix, $\vz^{(i)}\sim\gN(\vzero,\mI_k)$ are latent factors, and $\rvepsilon^{(i)}\sim\gN(\vzero,\bm{\Psi})$ is Gaussian noise with diagonal covariance $\bm{\Psi}=\mathrm{diag}(\psi_1,\ldots,\psi_d)$. The parameters ${\mL,\bm{\Psi},\vmu}$ are estimated using the Expectation-Maximization (EM) algorithm. The localized representation is a $k$-dimensional affine subspace of the form $\sG_{\mathrm{FA}}={\mL\vq+\vmu \mid \vq\in\R^k}$. We sample intervention points from the localized affine subspace using the gaussian sampling procedure as we do for other methods.

\paragraph{DiffMean}
DiffMean-Pairs extends DiffMean-r1~\citep{marks2024geometrytruthemergentlinear} to multi-class concepts by computing pairwise difference vectors between class centroids. Let $C={c_1,\ldots,c_{|C|}}$ denote the set of concept classes and let $\vmu_{c_j}$ be the mean activation of class $c_j$. For every pair of classes $(c_i,c_j)$ with $i<j$, we compute the difference vector $\vd_{i,j}=\vmu_{c_i}-\vmu_{c_j}$. Stacking all pairwise difference vectors forms the matrix $\mD\in\R^{d\times p}$, where $p=\binom{|C|}{2}$. We then perform singular value decomposition, $\mD=\mU\mSigma\mV^\top$, and retain the first $k$ left singular vectors to form the basis $\mL=[\vu_1|\cdots|\vu_k]\in\R^{d\times k}$. The affine offset is set to the global activation mean, $\vmu_C=\frac{1}{m}\sum_{i=1}^{m}\vx^{(i)}$.

The localized representation is therefore a $k$-dimensional affine subspace of the form $\sG_{\mathrm{DMP}}={\mL\vq+\vmu_C \mid \vq\in\R^k}$. We treat the localized representation as an affine subspace and sample intervention points using the Gaussian sampling procedure described above. 

\paragraph{Schatten Probe.}
Schatten Probe extends the linear probe by encouraging the classifier weight matrix to be low-rank through approximate Schatten-$p$ norm regularization. The classifier weight matrix is factorized as $\mW=\mA\mB^\top$, where $\mA\in\R^{d\times k}$ and $\mB\in\R^{|C|\times k}$. The parameters are learned by minimizing the cross-entropy loss together with a Schatten $p$-norm regularizer inspired by~\citep{Giampouras2020schatten,braun2025schatten} that encourages the probe's weights to be low-rank. After training, the weight matrix is reconstructed as $\mW=\mA\mB^\top$ and decomposed via singular value decomposition, $\mW=\mU\mSigma\mV^\top$. We retain the first $k_{\mathrm{eff}}$ left singular vectors, where $k_{\mathrm{eff}}$ is determined by the effective rank~\citep{effective_rank} of $\mW$, and set the affine offset to the global activation mean $\vmu_C=\frac{1}{m}\sum_{i=1}^{m}\vx^{(i)}$.

The localized representation is therefore a $k_{\mathrm{eff}}$-dimensional affine subspace of the form $\sG_{\mathrm{Schatten}}={\mL\vq+\vmu_C \mid \vq\in\R^{k_{\mathrm{eff}}}}$, where $\mL=[\vu_1|\cdots|\vu_{k_{\mathrm{eff}}}]$.

\paragraph{MFA}~\citep{shafran2026MFA}.
Mixture of Factor Analyzers (MFA) is an unsupervised method that models the activation space as a mixture of low-dimensional gaussian subspaces. Following~\citep{shafran2026MFA}, we train MFA on activations from the PILE~\citep{pile} dataset. We train one MFA model with 9,000 Gaussian components and defer the other details of model fitting to the original work. Unlike conept dataset, we take activation at every token of PILE while training MFAs.

To identify the component corresponding to a target concept, we use the concept dataset after training. For each activation $\vx$, MFA computes the posterior responsibility $r_k(\vx)=p(k\mid\vx)$ for every Gaussian component $k$. More details about how to compute responsibilities are in Section~\ref{app:MFA_select_gaussian}. Given positive concept activations $\mathcal{P}$ and negative activations $\mathcal{N}$, we compute a concept score for each Gaussian as $s_k=\frac{1}{|\mathcal{P}|}\sum_{\vx\in\mathcal{P}}r_k(\vx)-\frac{1}{|\mathcal{N}|}\sum_{\vx\in\mathcal{N}}r_k(\vx)$. Intuitively, a high score indicates that the component is consistently activated by the target concept while assigning low responsibility to unrelated activations. We rank all Gaussian components according to $s_k$ and select the highest-scoring component.

The localized representation is therefore one single gaussian component and we treat this as $k$-dimensional affine subspace $\sG_{\mathrm{MFA}}={\mL\vq+\vmu \mid \vq\in\R^k}$, where $\mL$ and $\vmu$ denote the factor loading matrix and centroid of the selected gaussian component.

\paragraph{SAE}\citep{bricken2023SAE1}.
SAE represents concepts using sparse one-dimensional features learned by a pretrained sparse autoencoder. We encode activations using a pretrained SAE and identify the subset of features that best predicts a target concept. For each feature, we compute its Matthews Correlation Coefficient (MCC) with the binary concept labels and rank features according to this score. We then select the top $k$ features together with their decoder vectors. We try two different strategies for sampling with SAE top-$k$ features and select the best performing strategy. More details are in Section~\ref{appsec: sae_feature_selection}.

\section{Sampling Evaluation Results across Models}
Here we discuss additional results for different models families. We have results for breakdown of different metrics in Tab~\ref{tab:best_rank_gemma27b_app},~\ref{tab:best_rank_llama31_app},~\ref{tab:best_rank_qwen35_app}. We show instance diversity and category diversity separately to indicate how much each component contributes to overall diversity.\footnote{Instance and category diversity do not always sum directly to the reported Diversity Score. This is because concepts without a category structure, such as \dataset{Days} and \dataset{Years}, the Diversity Score is defined as twice the instance diversity.}

\begin{figure}[htbp]
    \centering
    \includegraphics[width=\textwidth]{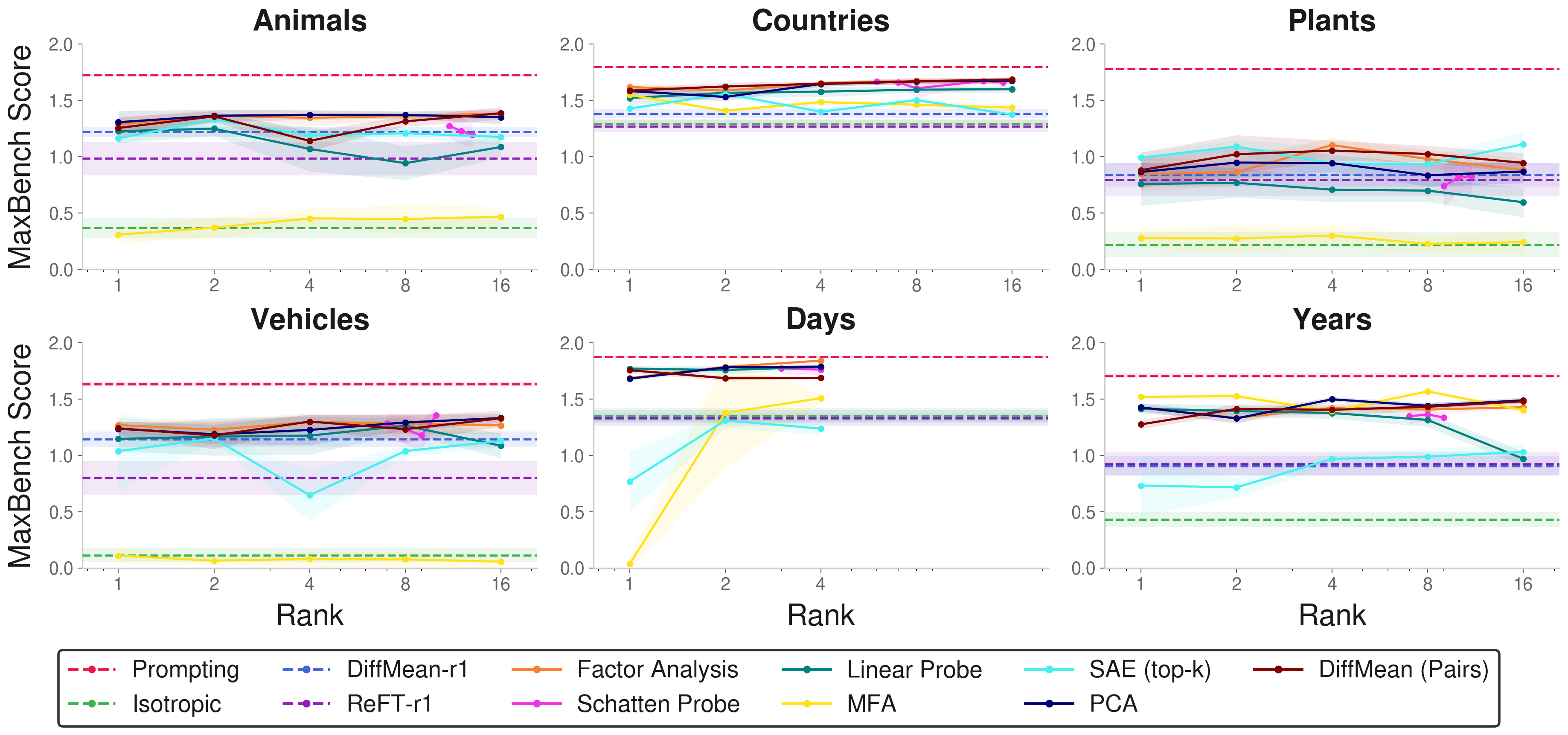}
    \caption{\maxbench~score by rank for Gemma-3-27B for all concepts. The x-axis is
rank (days is capped at rank $4$, as \textsc{Days} has 7 values, and we assume the rank cannot exceed the number of concept values); the y-axis is \maxbench~score; confidence intervals are computed via dataset split with 10 samples. Each point is the mean \maxbench~score at that (concept, method, rank); the shaded band is std.\ dev. Prompting, Isotropic, DiffMean-r1, and ReFT-r1 have no rank hyperparameter, so they are drawn as dashed lines. For Schatten Probes, we measure \emph{effective} rank instead of manually setting its rank; we round to the nearest integer.
}
    \label{fig:rank_x_maxbench_gemma27b}
\end{figure}

\begin{figure}[htbp]
    \centering
    \includegraphics[width=\textwidth]{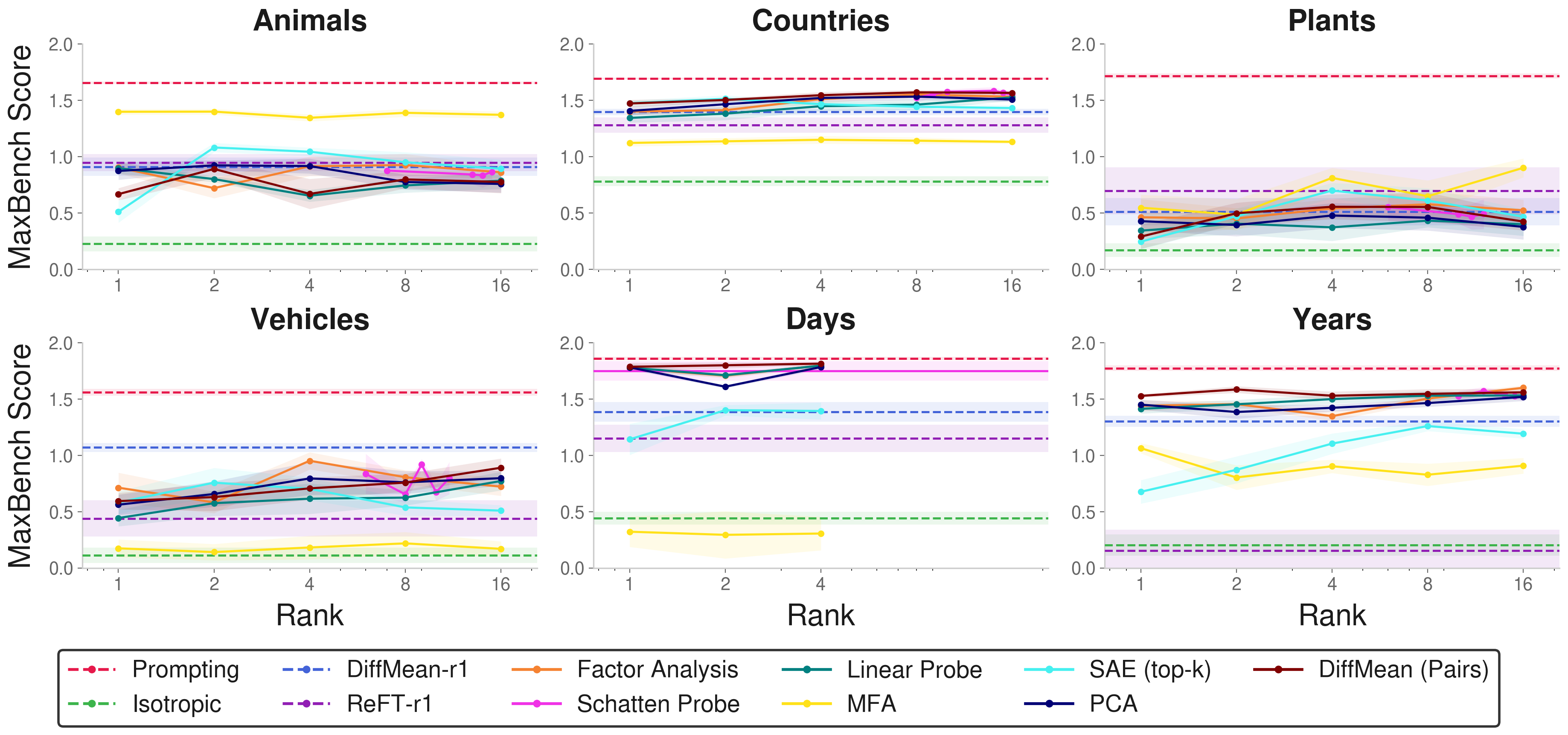}
    \caption{\maxbench~score by rank for Llama-3.1-8B for all concepts. The x-axis is
rank (days is capped at rank $4$, as \textsc{Days} has 7 values, and we assume the rank cannot exceed the number of concept values); the y-axis is \maxbench~score; confidence intervals are computed via dataset split with 10 samples. Each point is the mean \maxbench~score at that (concept, method, rank); the shaded band is std.\ dev. Prompting, Isotropic, DiffMean-r1, and ReFT-r1 have no rank hyperparameter, so they are drawn as dashed lines. For Schatten Probes, we measure \emph{effective} rank instead of manually setting its rank; we round to the nearest integer.
}
    \label{fig:rank_x_maxbench_llama8b}
\end{figure}

\begin{figure}[htbp]
    \centering
    \includegraphics[width=\textwidth]{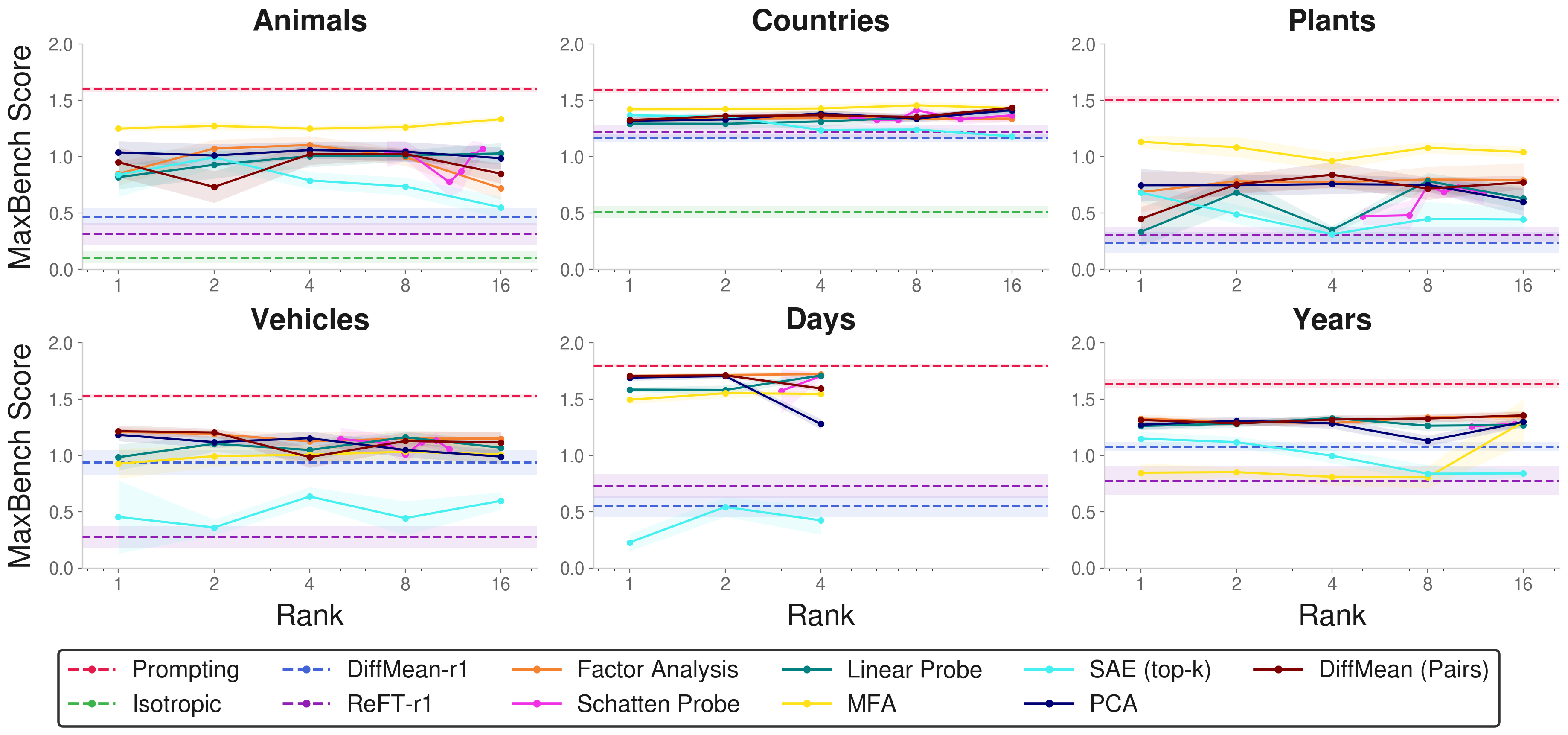}
    \caption{\maxbench~score by rank for Qwen-3.5-4B for all concepts. The x-axis is
rank (days is capped at rank $4$, as \textsc{Days} has 7 values, and we assume the rank cannot exceed the number of concept values); the y-axis is \maxbench~score; confidence intervals are computed via dataset split with 10 samples. Each point is the mean \maxbench~score at that (concept, method, rank); the shaded band is std.\ dev. Prompting, Isotropic, DiffMean-r1, and ReFT-r1 have no rank hyperparameter, so they are drawn as dashed lines. For Schatten Probes, we measure \emph{effective} rank instead of manually setting its rank; we round to the nearest integer.
}
    \label{fig:rank_x_maxbench_qwen4b}
\end{figure}

Results across ranks are in Figure~\ref{fig:rank_x_maxbench_gemma27b} for Gemma-3-27B, Figure~\ref{fig:rank_x_maxbench_llama8b} for Llama-3.1-8B and Figure~\ref{fig:rank_x_maxbench_qwen4b} for Qwen-3.5-4B. In Table~\ref{tab:best_rank_cross_model_maxbench_v2}, we compare \textsc{Axbench}, Diversity, and \maxbench{} scores.

\begin{table}[t]
\caption{Scores on the test set for Gemma-3-1B. We report \emph{Concept}, \emph{Fluency}, \emph{Instruction}, and \emph{Diversity} scores, together with the composite \textsc{AxBench} and \maxbench~scores. Full metric definitions are given in Section~\ref{subsec:scoring}. For each concept and method, we select the rank that achieves the highest mean \maxbench~score. Each reported value is the mean~$\pm$~standard deviation over all the data splits.}
\centering
\resizebox{\textwidth}{!}{%
\begin{tabular}{lcccccccc}
\toprule
\textbf{Method} & \textbf{Concept} & \textbf{Fluency} & \textbf{Instruction} &\textbf{ Inst.\ Div.} & \textbf{Cat.\ Div.} & \textbf{Diversity} & \textbf{\textsc{AxBench}} & \textbf{\maxbench} \\
\addlinespace[1pt]
\textit{\footnotesize } & \conceptIcon & \fluencyIcon & \instructIcon & {\small\textcolor{divgreen}{\instDivIcon}} & {\small\textcolor{divgreen}{\catDivIcon}} & {\small\textcolor{divgreen}{\divIcon\,=\,\instDivIcon$+$\catDivIcon}} & {\small\textcolor{axblue}{\axBenchIcon\,=\,HM(\conceptIcon,\fluencyIcon,\instructIcon)}} & {\small\textcolor{maxamber}{\maxBenchIcon\,=\,HM(\conceptIcon,\fluencyIcon,\instructIcon,\divIcon)}} \\
\midrule
Prompting & $0.67 \pm 0.34$ & $\mathbf{2.00 \pm 0.01}$ & \textbf{$1.65 \pm 0.05$} & $0.39 \pm 0.19$ & $0.49 \pm 0.20$ & $0.90 \pm 0.37$ & $0.61 \pm 0.32$ & $0.98 \pm 0.32$ \\
\midrule
Isotropic & $1.12 \pm 0.41$ & $1.80 \pm 0.14$ & $1.47 \pm 0.09$ & \underline{$0.73 \pm 0.15$} & $0.80 \pm 0.06$ & \underline{$1.53 \pm 0.27$} & $0.84 \pm 0.35$ & $1.39 \pm 0.29$ \\
Embedding & $0.85 \pm 0.38$ & $1.65 \pm 0.33$ & \underline{$1.48 \pm 0.14$} & $0.56 \pm 0.20$ & $0.80 \pm 0.05$ & $1.27 \pm 0.41$ & $0.60 \pm 0.32$ & $1.15 \pm 0.39$ \\
\midrule
DiffMean-r1 & $0.63 \pm 0.20$ & $1.33 \pm 0.13$ & $1.42 \pm 0.06$ & $0.67 \pm 0.15$ & $0.73 \pm 0.14$ & $1.42 \pm 0.27$ & $0.41 \pm 0.17$ & $1.04 \pm 0.16$ \\
ReFT-r1 & $0.34 \pm 0.23$ & $1.42 \pm 0.07$ & $1.45 \pm 0.06$ & $0.51 \pm 0.23$ & $0.61 \pm 0.15$ & $1.14 \pm 0.42$ & $0.24 \pm 0.18$ & $0.68 \pm 0.33$ \\
\midrule
MFA & $1.37 \pm 0.48$ & $1.73 \pm 0.19$ & $1.40 \pm 0.18$ & $0.71 \pm 0.16$ & $0.72 \pm 0.15$ & $1.44 \pm 0.31$ & $0.91 \pm 0.32$ & $1.40 \pm 0.17$ \\
SAE (top-k) & $0.63 \pm 0.27$ & $1.44 \pm 0.28$ & $1.45 \pm 0.08$ & $0.54 \pm 0.12$ & $0.72 \pm 0.09$ & $1.18 \pm 0.27$ & $0.43 \pm 0.19$ & $0.99 \pm 0.24$ \\
\midrule
Factor Analysis & \underline{$1.64 \pm 0.26$} & $1.86 \pm 0.13$ & $1.41 \pm 0.11$ & $0.72 \pm 0.10$ & $0.82 \pm 0.06$ & $1.52 \pm 0.15$ & $1.19 \pm 0.24$ & \underline{$1.58 \pm 0.11$} \\
Schatten Probe & $1.55 \pm 0.26$ & $1.83 \pm 0.14$ & $1.42 \pm 0.10$ & $\mathbf{0.73 \pm 0.12}$ & $0.82 \pm 0.06$ & $\mathbf{1.55 \pm 0.18}$ & $1.13 \pm 0.23$ & $1.56 \pm 0.12$ \\
Linear Probe & $1.50 \pm 0.31$ & $1.85 \pm 0.13$ & $1.44 \pm 0.10$ & $0.70 \pm 0.12$ & $0.81 \pm 0.08$ & $1.49 \pm 0.20$ & $1.12 \pm 0.26$ & $1.54 \pm 0.13$ \\
PCA & $1.59 \pm 0.29$ & \underline{$1.94 \pm 0.06$} & $1.44 \pm 0.11$ & $0.71 \pm 0.10$ & \underline{$0.82 \pm 0.08$} & $1.51 \pm 0.17$ & \underline{$1.21 \pm 0.24$} & \underline{$1.58 \pm 0.11$} \\
Diff Mean (Pairs) & $\mathbf{1.64 \pm 0.27}$ & $1.87 \pm 0.13$ & $1.43 \pm 0.10$ & $0.71 \pm 0.10$ & $\mathbf{0.84 \pm 0.07}$ & $1.53 \pm 0.15$ & $\mathbf{1.22 \pm 0.25}$ & $\mathbf{1.59 \pm 0.11}$ \\

\bottomrule
\end{tabular}%
}
\label{tab:best_rank_gemma1b_app}
\end{table}

\begin{table}[t]
\caption{Scores on the test set for Gemma-3-27B. We report \emph{Concept}, \emph{Fluency}, \emph{Instruction}, and \emph{Diversity} scores, together with the composite \textsc{AxBench} and \maxbench~scores. Full metric definitions are given in Section~\ref{subsec:scoring}. For each concept and method, we select the rank that achieves the highest mean \maxbench~score. Each reported value is the mean~$\pm$~standard deviation over all the data splits.}
\centering
\resizebox{\textwidth}{!}{%
\begin{tabular}{lcccccccc}
\toprule
\textbf{Method} & \textbf{Concept} & \textbf{Fluency} & \textbf{Instruction} & \textbf{Inst.\ Div.} & \textbf{Cat.\ Div.} & \textbf{Diversity} & \textbf{\textsc{AxBench}} & \textbf{\maxbench} \\
\addlinespace[1pt]
\textit{\footnotesize } & \conceptIcon & \fluencyIcon & \instructIcon & {\small\textcolor{divgreen}{\instDivIcon}} & {\small\textcolor{divgreen}{\catDivIcon}} & {\small\textcolor{divgreen}{\divIcon\,=\,\instDivIcon$+$\catDivIcon}} & {\small\textcolor{axblue}{\axBenchIcon\,=\,HM(\conceptIcon,\fluencyIcon,\instructIcon)}} & {\small\textcolor{maxamber}{\maxBenchIcon\,=\,HM(\conceptIcon,\fluencyIcon,\instructIcon,\divIcon)}} \\
\midrule
Prompting & $\mathbf{1.99 \pm 0.02}$ & $\mathbf{2.00 \pm 0.01}$ & $1.66 \pm 0.15$ & $\mathbf{0.78 \pm 0.14}$ & $0.67 \pm 0.15$ & $\mathbf{1.54 \pm 0.32}$ & $\mathbf{1.77 \pm 0.13}$ & $\mathbf{1.75 \pm 0.08}$ \\
\midrule
Isotropic & $0.36 \pm 0.39$ & $1.62 \pm 0.21$ & $\mathbf{1.77 \pm 0.05}$ & $0.50 \pm 0.22$ & $0.62 \pm 0.17$ & $1.14 \pm 0.41$ & $0.29 \pm 0.31$ & $0.63 \pm 0.51$ \\
Embedding  & $0.95 \pm 0.45$ & $1.27 \pm 0.39$ & $1.64 \pm 0.06$ & $0.56 \pm 0.16$ & $\mathbf{0.82 \pm 0.06}$ & $1.28 \pm 0.34$ & $0.72 \pm 0.40$ & $1.15 \pm 0.45$ \\
\midrule
DiffMean-r1 & $0.75 \pm 0.32$ & $1.63 \pm 0.26$ & $1.59 \pm 0.08$ & $0.62 \pm 0.12$ & $0.71 \pm 0.10$ & $1.34 \pm 0.21$ & $0.56 \pm 0.29$ & $1.14 \pm 0.22$ \\
ReFT-r1 & $0.59 \pm 0.28$ & $1.60 \pm 0.06$ & $1.63 \pm 0.07$ & $0.62 \pm 0.18$ & $0.71 \pm 0.11$ & $1.34 \pm 0.31$ & $0.37 \pm 0.22$ & $1.02 \pm 0.24$ \\
\midrule
MFA & $0.74 \pm 0.69$ & $1.77 \pm 0.14$ & $1.67 \pm 0.09$ & $0.54 \pm 0.25$ & $0.56 \pm 0.23$ & $1.17 \pm 0.47$ & $0.63 \pm 0.58$ & $0.92 \pm 0.64$ \\
SAE (top-k) & $0.96 \pm 0.34$ & $1.46 \pm 0.27$ & $1.56 \pm 0.12$ & \underline{$0.68 \pm 0.15$} & $0.74 \pm 0.05$ & $1.42 \pm 0.21$ & $0.67 \pm 0.28$ & $1.25 \pm 0.19$ \\
\midrule
Factor Analysis & $1.32 \pm 0.52$ & \underline{$1.79 \pm 0.25$} & $1.68 \pm 0.09$ & $0.61 \pm 0.11$ & $0.80 \pm 0.07$ & $1.37 \pm 0.20$ & \underline{$1.11 \pm 0.50$} & \underline{$1.45 \pm 0.25$} \\
Schatten Probe & $1.21 \pm 0.60$ & $1.76 \pm 0.19$ & \underline{$1.70 \pm 0.08$} & $0.63 \pm 0.11$ & $0.79 \pm 0.06$ & $1.39 \pm 0.18$ & $1.01 \pm 0.52$ & $1.38 \pm 0.31$ \\
Linear Probe & $1.18 \pm 0.56$ & $1.70 \pm 0.17$ & $1.70 \pm 0.07$ & $0.61 \pm 0.16$ & $0.75 \pm 0.09$ & $1.35 \pm 0.28$ & $0.97 \pm 0.48$ & $1.35 \pm 0.33$ \\
PCA & $1.33 \pm 0.56$ & $1.75 \pm 0.21$ & $1.68 \pm 0.10$ & $0.64 \pm 0.14$ & $0.79 \pm 0.08$ & $1.41 \pm 0.25$ & $1.09 \pm 0.51$ & $1.44 \pm 0.28$ \\
Diff Mean (Pairs) & \underline{$1.34 \pm 0.52$} & $1.74 \pm 0.24$ & $1.69 \pm 0.07$ & $0.63 \pm 0.14$ & \underline{$0.81 \pm 0.07$} & $1.41 \pm 0.24$ & $1.11 \pm 0.48$ & $1.45 \pm 0.24$ \\

\bottomrule
\end{tabular}%
}

\label{tab:best_rank_gemma27b_app}
\end{table}

\begin{table}[t]
\caption{Scores on the test set for Llama-3.1-8B. We report \emph{Concept}, \emph{Fluency}, \emph{Instruction}, and \emph{Diversity} scores, together with the composite \textsc{AxBench} and \maxbench~scores. Full metric definitions are given in Section~\ref{subsec:scoring}. For each concept and method, we select the rank that achieves the highest mean \maxbench~score. Each reported value is the mean~$\pm$~standard deviation over all the data splits.}
\centering
\resizebox{\textwidth}{!}{%
\begin{tabular}{lcccccccc}
\toprule
\textbf{Method} & \textbf{Concept} & \textbf{Fluency} & \textbf{Instruction} & \textbf{Inst.\ Div.} & \textbf{Cat.\ Div.} & \textbf{Diversity} & \textbf{\textsc{AxBench}} & \textbf{\maxbench} \\
\addlinespace[1pt]
\textit{\footnotesize } & \conceptIcon & \fluencyIcon & \instructIcon & {\small\textcolor{divgreen}{\instDivIcon}} & {\small\textcolor{divgreen}{\catDivIcon}} & {\small\textcolor{divgreen}{\divIcon\,=\,\instDivIcon$+$\catDivIcon}} & {\small\textcolor{axblue}{\axBenchIcon\,=\,HM(\conceptIcon,\fluencyIcon,\instructIcon)}} & {\small\textcolor{maxamber}{\maxBenchIcon\,=\,HM(\conceptIcon,\fluencyIcon,\instructIcon,\divIcon)}} \\
\midrule
Prompting & $\mathbf{1.98 \pm 0.03}$ & $\mathbf{1.96 \pm 0.04}$ & $1.56 \pm 0.10$ & $\mathbf{0.78 \pm 0.13}$ & $0.62 \pm 0.17$ & $\mathbf{1.52 \pm 0.32}$ & $\mathbf{1.70 \pm 0.08}$ & $\mathbf{1.71 \pm 0.10}$ \\
\midrule
Isotropic & $0.17 \pm 0.22$ & $0.64 \pm 0.08$ & $1.38 \pm 0.08$ & $0.42 \pm 0.18$ & $0.63 \pm 0.12$ & $0.99 \pm 0.35$ & $0.09 \pm 0.11$ & $0.32 \pm 0.24$ \\
Embedding & $0.22 \pm 0.27$ & $1.75 \pm 0.04$ & $\mathbf{1.75 \pm 0.03}$ & $0.31 \pm 0.17$ & $0.52 \pm 0.18$ & $0.74 \pm 0.39$ & $0.20 \pm 0.25$ & $0.43 \pm 0.35$ \\
\midrule
DiffMean-r1 & $0.67 \pm 0.33$ & $1.75 \pm 0.07$ & $1.67 \pm 0.03$ & $0.64 \pm 0.17$ & $0.71 \pm 0.08$ & $1.39 \pm 0.26$ & $0.56 \pm 0.29$ & $1.10 \pm 0.33$ \\
ReFT-r1 & $0.41 \pm 0.30$ & $1.78 \pm 0.06$ & \underline{$1.73 \pm 0.04$} & $0.51 \pm 0.27$ & $0.71 \pm 0.11$ & $1.18 \pm 0.54$ & $0.35 \pm 0.27$ & $0.78 \pm 0.42$ \\
\midrule
MFA & $0.56 \pm 0.45$ & $1.68 \pm 0.18$ & $1.65 \pm 0.12$ & $0.51 \pm 0.24$ & $0.59 \pm 0.09$ & $1.07 \pm 0.36$ & $0.44 \pm 0.32$ & $0.84 \pm 0.44$ \\
SAE (top-k) & $0.73 \pm 0.39$ & $1.62 \pm 0.09$ & $1.65 \pm 0.06$ & \underline{$0.67 \pm 0.18$} & $0.65 \pm 0.14$ & $1.39 \pm 0.35$ & $0.56 \pm 0.31$ & $1.12 \pm 0.32$ \\
\midrule
Factor Analysis & $0.94 \pm 0.60$ & $1.72 \pm 0.18$ & $1.71 \pm 0.07$ & $0.67 \pm 0.17$ & $0.75 \pm 0.06$ & \underline{$1.47 \pm 0.25$} & $0.82 \pm 0.55$ & \underline{$1.23 \pm 0.45$} \\
Schatten Probe & \underline{$0.95 \pm 0.65$} & $1.73 \pm 0.14$ & $1.70 \pm 0.06$ & $0.66 \pm 0.17$ & $\mathbf{0.76 \pm 0.07}$ & $1.46 \pm 0.25$ & $0.81 \pm 0.55$ & $1.21 \pm 0.45$ \\
Linear Probe & $0.89 \pm 0.62$ & $1.77 \pm 0.16$ & $1.72 \pm 0.06$ & $0.62 \pm 0.19$ & $0.69 \pm 0.12$ & $1.37 \pm 0.31$ & $0.78 \pm 0.56$ & $1.17 \pm 0.50$ \\

PCA & $0.88 \pm 0.60$ & $1.83 \pm 0.13$ & $1.70 \pm 0.06$ & $0.64 \pm 0.18$ & $0.71 \pm 0.11$ & $1.41 \pm 0.29$ & $0.77 \pm 0.54$ & $1.18 \pm 0.48$ \\
Diff Mean (Pairs) & $0.94 \pm 0.64$ & \underline{$1.79 \pm 0.14$} & $1.71 \pm 0.07$ & \underline{$0.67 \pm 0.18$} & \underline{$0.75 \pm 0.09$} & $1.46 \pm 0.26$ & \underline{$0.83 \pm 0.57$} & $1.22 \pm 0.47$ \\

\bottomrule
\end{tabular}%
}

\label{tab:best_rank_llama31_app}
\end{table}

\begin{table}[t]
\caption{Scores on the test set for Qwen-3.5-4B. We report \emph{Concept}, \emph{Fluency}, \emph{Instruction}, and \emph{Diversity} scores, together with the composite \textsc{AxBench} and \maxbench~scores. Full metric definitions are given in Section~\ref{subsec:scoring}. For each concept and method, we select the rank that achieves the highest mean \maxbench~score. Each reported value is the mean~$\pm$~standard deviation over all the data splits.}
\centering
\resizebox{\textwidth}{!}{%
\begin{tabular}{lcccccccc}
\toprule
\textbf{Method} & \textbf{Concept} & \textbf{Fluency} & \textbf{Instruction} & \textbf{Inst.\ Div.} & \textbf{Cat.\ Div.} & \textbf{Diversity} & \textbf{\textsc{AxBench}} & \textbf{\maxbench} \\
\addlinespace[1pt]
\textit{\footnotesize } & \conceptIcon & \fluencyIcon & \instructIcon & {\small\textcolor{divgreen}{\instDivIcon}} & {\small\textcolor{divgreen}{\catDivIcon}} & {\small\textcolor{divgreen}{\divIcon\,=\,\instDivIcon$+$\catDivIcon}} & {\small\textcolor{axblue}{\axBenchIcon\,=\,HM(\conceptIcon,\fluencyIcon,\instructIcon)}} & {\small\textcolor{maxamber}{\maxBenchIcon\,=\,HM(\conceptIcon,\fluencyIcon,\instructIcon,\divIcon)}} \\
\midrule
Prompting & $\mathbf{1.95 \pm 0.05}$ & $\mathbf{1.98 \pm 0.03}$ & $1.28 \pm 0.13$ & $\mathbf{0.78 \pm 0.15}$ & $0.62 \pm 0.10$ & \underline{$1.51 \pm 0.31$} & $\mathbf{1.38 \pm 0.10}$ & $\mathbf{1.61 \pm 0.10}$ \\
\midrule
Isotropic & $0.21 \pm 0.19$ & $0.27 \pm 0.09$ & $0.69 \pm 0.13$ & $0.41 \pm 0.13$ & $0.62 \pm 0.10$ & $1.02 \pm 0.17$ & $0.09 \pm 0.09$ & $0.31 \pm 0.21$ \\
Embedding & $0.66 \pm 0.34$ & $1.47 \pm 0.28$ & $1.35 \pm 0.14$ & $0.53 \pm 0.18$ & $0.71 \pm 0.10$ & $1.23 \pm 0.30$ & $0.41 \pm 0.23$ & $1.00 \pm 0.28$ \\
\midrule
MFA & \underline{$1.34 \pm 0.33$} & $1.55 \pm 0.22$ & $1.18 \pm 0.16$ & $0.68 \pm 0.20$ & $0.59 \pm 0.17$ & $1.34 \pm 0.40$ & \underline{$0.75 \pm 0.24$} & \underline{$1.30 \pm 0.20$} \\
SAE (top-k) & $0.55 \pm 0.38$ & $1.47 \pm 0.12$ & $1.36 \pm 0.09$ & $0.64 \pm 0.17$ & $0.65 \pm 0.13$ & $1.34 \pm 0.33$ & $0.32 \pm 0.23$ & $0.89 \pm 0.31$ \\
\midrule
DiffMean-r1 & $0.40 \pm 0.29$ & $1.58 \pm 0.15$ & $1.44 \pm 0.09$ & $0.62 \pm 0.20$ & $0.70 \pm 0.08$ & $1.36 \pm 0.32$ & $0.27 \pm 0.24$ & $0.74 \pm 0.35$ \\
ReFT-r1 & $0.30 \pm 0.30$ & $1.72 \pm 0.05$ & $\mathbf{1.53 \pm 0.06}$ & $0.54 \pm 0.22$ & $0.65 \pm 0.10$ & $1.22 \pm 0.38$ & $0.23 \pm 0.25$ & $0.60 \pm 0.36$ \\
\midrule
Factor Analysis & $1.01 \pm 0.46$ & $1.56 \pm 0.25$ & $1.43 \pm 0.11$ & $0.70 \pm 0.15$ & $0.78 \pm 0.07$ & $1.50 \pm 0.21$ & $0.70 \pm 0.37$ & $1.25 \pm 0.29$ \\
Schatten Probe & $0.97 \pm 0.51$ & $1.62 \pm 0.24$ & $1.44 \pm 0.10$ & $0.67 \pm 0.14$ & $0.78 \pm 0.07$ & $1.47 \pm 0.20$ & $0.66 \pm 0.40$ & $1.22 \pm 0.32$ \\
Linear Probe & $0.95 \pm 0.44$ & $1.60 \pm 0.27$ & $1.42 \pm 0.11$ & $0.68 \pm 0.13$ & $0.79 \pm 0.06$ & $1.48 \pm 0.17$ & $0.66 \pm 0.39$ & $1.24 \pm 0.30$ \\

PCA & $0.94 \pm 0.48$ & \underline{$1.75 \pm 0.11$} & \underline{$1.44 \pm 0.12$} & $0.67 \pm 0.14$ & \underline{$0.79 \pm 0.09$} & $1.47 \pm 0.20$ & $0.67 \pm 0.38$ & $1.24 \pm 0.30$ \\
Diff Mean (Pairs) & $0.99 \pm 0.45$ & $1.70 \pm 0.25$ & $1.42 \pm 0.12$ & $0.70 \pm 0.12$ & $\mathbf{0.80 \pm 0.07}$ & $\mathbf{1.52 \pm 0.18}$ & $0.69 \pm 0.37$ & $1.26 \pm 0.29$ \\

\bottomrule
\end{tabular}%
}

\label{tab:best_rank_qwen35_app}
\end{table}

\begin{table}[t]
  \caption{\textsc{AxBench}, Diversity, and \maxbench~scores (mean $\pm$ std across al dataset splits, for all concepts and ranks; see the per-model tables) across all four models. \maxbench~here is the harmonic mean taken directly over Concept, Fluency, Instruction, and Diversity. Per column, the highest mean is bold.}
  \centering
  \resizebox{\textwidth}{!}{%
  \begin{tabular}{lcccccccccccc}
    \toprule
     & \multicolumn{4}{c}{\textsc{AxBench}} & \multicolumn{4}{c}{Diversity} & \multicolumn{4}{c}{\maxbench} \\
    \cmidrule(lr){2-5} \cmidrule(lr){6-9} \cmidrule(lr){10-13}
    Method & Gemma-3-1B & Gemma-3-27B & Llama-3.1-8B & Qwen-3.5-4B & Gemma-3-1B & Gemma-3-27B & Llama-3.1-8B & Qwen-3.5-4B & Gemma-3-1B & Gemma-3-27B & Llama-3.1-8B & Qwen-3.5-4B \\
    \midrule
    Prompting & $0.61 \pm 0.32$ & $\mathbf{1.77 \pm 0.13}$ & $\mathbf{1.70 \pm 0.08}$ & $\mathbf{1.38 \pm 0.10}$ & $0.90 \pm 0.37$ & $\mathbf{1.54 \pm 0.32}$ & $\mathbf{1.52 \pm 0.32}$ & $1.51 \pm 0.31$ & $0.98 \pm 0.32$ & $\mathbf{1.75 \pm 0.08}$ & $\mathbf{1.71 \pm 0.10}$ & $\mathbf{1.61 \pm 0.10}$ \\
    \midrule
    Isotropic & $0.84 \pm 0.35$ & $0.29 \pm 0.31$ & $0.09 \pm 0.11$ & $0.09 \pm 0.09$ & $1.53 \pm 0.27$ & $1.14 \pm 0.41$ & $0.99 \pm 0.35$ & $1.02 \pm 0.17$ & $1.39 \pm 0.29$ & $0.63 \pm 0.51$ & $0.32 \pm 0.24$ & $0.31 \pm 0.21$ \\
    Embedding & $0.60 \pm 0.32$ & $0.72 \pm 0.40$ & $0.20 \pm 0.25$ & $0.41 \pm 0.23$ & $1.27 \pm 0.41$ & $1.28 \pm 0.34$ & $0.74 \pm 0.39$ & $1.23 \pm 0.30$ & $1.15 \pm 0.39$ & $1.15 \pm 0.45$ & $0.43 \pm 0.35$ & $1.00 \pm 0.28$ \\
    \midrule
    DiffMean-r1 & $0.41 \pm 0.17$ & $0.56 \pm 0.29$ & $0.56 \pm 0.29$ & $0.27 \pm 0.24$ & $1.42 \pm 0.27$ & $1.34 \pm 0.21$ & $1.39 \pm 0.26$ & $1.36 \pm 0.32$ & $1.04 \pm 0.16$ & $1.14 \pm 0.22$ & $1.10 \pm 0.33$ & $0.74 \pm 0.35$ \\
    ReFT-r1 & $0.24 \pm 0.18$ & $0.37 \pm 0.22$ & $0.35 \pm 0.27$ & $0.23 \pm 0.25$ & $1.14 \pm 0.42$ & $1.34 \pm 0.31$ & $1.18 \pm 0.54$ & $1.22 \pm 0.38$ & $0.68 \pm 0.33$ & $1.02 \pm 0.24$ & $0.78 \pm 0.42$ & $0.60 \pm 0.36$ \\
    \midrule
    MFA & $0.91 \pm 0.32$ & $0.63 \pm 0.58$ & $0.44 \pm 0.32$ & $0.75 \pm 0.24$ & $1.44 \pm 0.31$ & $1.17 \pm 0.47$ & $1.07 \pm 0.36$ & $1.34 \pm 0.40$ & $1.40 \pm 0.17$ & $0.92 \pm 0.64$ & $0.84 \pm 0.44$ & $1.30 \pm 0.20$ \\
    SAE (top-k) & $0.43 \pm 0.19$ & $0.67 \pm 0.28$ & $0.56 \pm 0.31$ & $0.32 \pm 0.23$ & $1.18 \pm 0.27$ & $1.42 \pm 0.21$ & $1.39 \pm 0.35$ & $1.34 \pm 0.33$ & $0.99 \pm 0.24$ & $1.25 \pm 0.19$ & $1.12 \pm 0.32$ & $0.89 \pm 0.31$ \\
    \midrule
    Factor Analysis & $1.19 \pm 0.24$ & $1.11 \pm 0.50$ & $0.82 \pm 0.55$ & $0.70 \pm 0.37$ & $1.52 \pm 0.15$ & $1.37 \pm 0.20$ & $1.47 \pm 0.25$ & $1.50 \pm 0.21$ & $1.58 \pm 0.11$ & $1.45 \pm 0.25$ & $1.23 \pm 0.45$ & $1.25 \pm 0.29$ \\
    Schatten Probe & $1.13 \pm 0.23$ & $1.01 \pm 0.52$ & $0.81 \pm 0.55$ & $0.66 \pm 0.40$ & $\mathbf{1.55 \pm 0.18}$ & $1.39 \pm 0.18$ & $1.46 \pm 0.25$ & $1.47 \pm 0.20$ & $1.56 \pm 0.12$ & $1.38 \pm 0.31$ & $1.21 \pm 0.45$ & $1.22 \pm 0.32$ \\
    Linear Probe & $1.12 \pm 0.26$ & $0.97 \pm 0.48$ & $0.78 \pm 0.56$ & $0.66 \pm 0.39$ & $1.49 \pm 0.20$ & $1.35 \pm 0.28$ & $1.37 \pm 0.31$ & $1.48 \pm 0.17$ & $1.54 \pm 0.13$ & $1.35 \pm 0.33$ & $1.17 \pm 0.50$ & $1.24 \pm 0.30$ \\

    PCA & $1.21 \pm 0.24$ & $1.09 \pm 0.51$ & $0.77 \pm 0.54$ & $0.67 \pm 0.38$ & $1.51 \pm 0.17$ & $1.41 \pm 0.25$ & $1.41 \pm 0.29$ & $1.47 \pm 0.20$ & $1.58 \pm 0.11$ & $1.44 \pm 0.28$ & $1.18 \pm 0.48$ & $1.24 \pm 0.30$ \\
    DiffMean (Pairs) & $\mathbf{1.22 \pm 0.25}$ & $1.11 \pm 0.48$ & $0.83 \pm 0.57$ & $0.69 \pm 0.37$ & $1.53 \pm 0.15$ & $1.41 \pm 0.24$ & $1.46 \pm 0.26$ & $\mathbf{1.52 \pm 0.18}$ & $\mathbf{1.59 \pm 0.11}$ & $1.45 \pm 0.24$ & $1.22 \pm 0.47$ & $1.26 \pm 0.29$ \\
    \bottomrule
  \end{tabular}%
  }

  \label{tab:best_rank_cross_model_maxbench_v2}
\end{table}

\section{Unseen Instances Analysis}

In addition to the metrics introduced in the \maxbench, we evaluate how often each method generates instances of a concept unseen during concept recovery. We perform this analysis for the categorical concepts \dataset{Animals}, \dataset{Vehicles}, \dataset{Plants}, and \dataset{Countries}. For each steered output, we use an LLM judge to extract all entities belonging to the target concept and then check whether each extracted entity appears in our dataset. We count as \emph{unseen} any entity that was not included in the corresponding concept dataset.

When localizing concept subspaces from activations, or when selecting Gaussian components for MFA and features for SAEs, we use the entities listed in Table~\ref{tab:concept-categories}. We therefore ask whether sampling from the learned representation only reproduces instances seen in this dataset, or whether it can also generate additional valid instances of the concept.

This question is particularly interesting for supervised methods such as DiffMean or Schatten Probe, which are fit only using the concept dataset. Unlike MFA and SAEs, whose underlying representations are learned from large-scale unsupervised data, these methods do not necessarily need to explicitly steer to concept instances outside the examples used for fitting. Nevertheless, we observe unseen instances not only for MFA and SAE, where this behavior may be more expected, but also for the supervised methods. This suggests that the localized concept representations can support sampling beyond the specific entities observed during subspace localization.

\begin{table}[t]
  \caption{Highest mean OOD count $\pm$ std (across all dataset splits), per method and concept. Per concept column, the method with the highest mean is bold.}\label{ood-example}
  \centering
  \begin{minipage}{0.49\textwidth}
    \centering
    \resizebox{\linewidth}{!}{%
  \begin{tabular}{lcccc}
    \toprule
    \multicolumn{5}{c}{\textbf{Gemma-3-1B}} \\
    \midrule
    Method & Animals & Countries & Plants & Vehicles \\
    \midrule
    DiffMean-r1 & 41 $\pm$ 3 & 11 $\pm$ 3 & 49 $\pm$ 10 & 29 $\pm$ 8 \\
    ReFT-r1 & 6 $\pm$ 8 & 4 $\pm$ 2 & 11 $\pm$ 6 & 4 $\pm$ 3 \\
    \midrule
    MFA & 61 $\pm$ 12 & \textbf{65} $\pm$ 8 & 46 $\pm$ 7 & 35 $\pm$ 15 \\
    SAE (top-k) & 36 $\pm$ 8 & 24 $\pm$ 9 & 59 $\pm$ 8 & 22 $\pm$ 8 \\
    \midrule
    Factor Analysis & \textbf{72} $\pm$ 7 & 26 $\pm$ 6 & 62 $\pm$ 22 & \textbf{76} $\pm$ 5 \\
    Schatten Probe & 60 $\pm$ 19 & 26 $\pm$ 7 & \textbf{67} $\pm$ 16 & 55 $\pm$ 12 \\
    Linear Probe & 66 $\pm$ 7 & 28 $\pm$ 8 & 63 $\pm$ 9 & 61 $\pm$ 8 \\

    PCA & 64 $\pm$ 7 & 26 $\pm$ 6 & 56 $\pm$ 5 & 37 $\pm$ 8 \\
    Diff Mean (Pairs) & 72 $\pm$ 6 & 36 $\pm$ 7 & 66 $\pm$ 9 & 73 $\pm$ 7 \\
    \bottomrule
  \end{tabular}%
    }
  \end{minipage}
  \hfill
  \begin{minipage}{0.49\textwidth}
    \centering
    \resizebox{\linewidth}{!}{%
  \begin{tabular}{lcccc}
    \toprule
    \multicolumn{5}{c}{\textbf{Gemma-3-27B}} \\
    \midrule
    Method & Animals & Countries & Plants & Vehicles \\
    \midrule
    DiffMean-r1 & 43 $\pm$ 11 & 8 $\pm$ 3 & 30 $\pm$ 12 & 35 $\pm$ 6 \\
    ReFT-r1 & 2 $\pm$ 6 & 8 $\pm$ 5 & 3 $\pm$ 6 & 9 $\pm$ 6 \\
    \midrule
    MFA & 19 $\pm$ 4 & 25 $\pm$ 7 & 9 $\pm$ 4 & 2 $\pm$ 1 \\
    SAE (top-k) & \textbf{57} $\pm$ 8 & \textbf{51} $\pm$ 6 & \textbf{46} $\pm$ 28 & \textbf{56} $\pm$ 12 \\
    \midrule
    Factor Analysis & 34 $\pm$ 6 & 12 $\pm$ 5 & 34 $\pm$ 6 & 22 $\pm$ 7 \\
    Schatten Probe & 31 $\pm$ 7 & 15 $\pm$ 10 & 22 $\pm$ 8 & 27 $\pm$ 8 \\
    Linear Probe & 29 $\pm$ 4 & 20 $\pm$ 4 & 21 $\pm$ 9 & 22 $\pm$ 6 \\

    PCA & 34 $\pm$ 7 & 33 $\pm$ 4 & 26 $\pm$ 6 & 22 $\pm$ 5 \\
    Diff Mean (Pairs) & 35 $\pm$ 6 & 35 $\pm$ 6 & 30 $\pm$ 10 & 20 $\pm$ 4 \\
    \bottomrule
  \end{tabular}%
    }
  \end{minipage}

  \begin{minipage}{0.49\textwidth}
    \centering
    \resizebox{\linewidth}{!}{%
  \begin{tabular}{lcccc}
    \toprule
    \multicolumn{5}{c}{\textbf{Llama-3.1-8B}} \\
    \midrule
    Method & Animals & Countries & Plants & Vehicles \\
    \midrule
    DiffMean-r1 & 29 $\pm$ 7 & 14 $\pm$ 4 & 10 $\pm$ 4 & 20 $\pm$ 4 \\
    ReFT-r1 & 4 $\pm$ 7 & 7 $\pm$ 3 & 10 $\pm$ 6 & 4 $\pm$ 1 \\
    \midrule
    MFA & \textbf{64} $\pm$ 8 & 4 $\pm$ 2 & \textbf{40} $\pm$ 9 & 3 $\pm$ 1 \\
    SAE (top-k) & 34 $\pm$ 8 & \textbf{31} $\pm$ 4 & 14 $\pm$ 4 & 15 $\pm$ 7 \\
    \midrule
    Factor Analysis & 25 $\pm$ 6 & 11 $\pm$ 3 & 16 $\pm$ 7 & \textbf{29} $\pm$ 5 \\
    Schatten Probe & 26 $\pm$ 6 & 12 $\pm$ 6 & 15 $\pm$ 8 & 22 $\pm$ 5 \\
    Linear Probe & 22 $\pm$ 4 & 12 $\pm$ 4 & 8 $\pm$ 3 & 15 $\pm$ 3 \\

    PCA & 25 $\pm$ 3 & 21 $\pm$ 4 & 9 $\pm$ 5 & 18 $\pm$ 5 \\
    Diff Mean (Pairs) & 25 $\pm$ 6 & 12 $\pm$ 4 & 16 $\pm$ 6 & 19 $\pm$ 5 \\
    \bottomrule
  \end{tabular}%
    }
  \end{minipage}
  \hfill
  \begin{minipage}{0.49\textwidth}
    \centering
    \resizebox{\linewidth}{!}{%
  \begin{tabular}{lcccc}
    \toprule
    \multicolumn{5}{c}{\textbf{Qwen-3.5-4B}} \\
    \midrule
    Method & Animals & Countries & Plants & Vehicles \\
    \midrule
    DiffMean-r1 & 21 $\pm$ 3 & 8 $\pm$ 3 & 10 $\pm$ 4 & 30 $\pm$ 8 \\
    ReFT-r1 & 14 $\pm$ 5 & 5 $\pm$ 3 & 10 $\pm$ 3 & 3 $\pm$ 2 \\
    \midrule
    MFA & \textbf{59} $\pm$ 10 & \textbf{47} $\pm$ 7 & \textbf{54} $\pm$ 21 & 28 $\pm$ 12 \\
    SAE (top-k) & 28 $\pm$ 6 & 24 $\pm$ 7 & -- & 9 $\pm$ 6 \\
    \midrule
    Factor Analysis & 32 $\pm$ 8 & 14 $\pm$ 4 & 27 $\pm$ 10 & \textbf{40} $\pm$ 9 \\
    Schatten Probe & 34 $\pm$ 6 & 16 $\pm$ 7 & 21 $\pm$ 7 & 32 $\pm$ 9 \\
    Linear Probe & 34 $\pm$ 5 & 9 $\pm$ 4 & 20 $\pm$ 9 & 29 $\pm$ 7 \\
    PCA & 31 $\pm$ 8 & 15 $\pm$ 5 & 24 $\pm$ 9 & 22 $\pm$ 6 \\
    Diff Mean (Pairs) & 32 $\pm$ 4 & 14 $\pm$ 7 & 24 $\pm$ 6 & 38 $\pm$ 8 \\
    \bottomrule
  \end{tabular}%
    }
  \end{minipage}

  \label{tab:ood_max_by_method}
\end{table}

\section{Affine vs Linear comparison}
We observe similar trends for affine vs linear offset for other models as we did for Gemma-3-1B.

\begin{figure}[thbp]
    \centering
    \includegraphics[width=\textwidth]{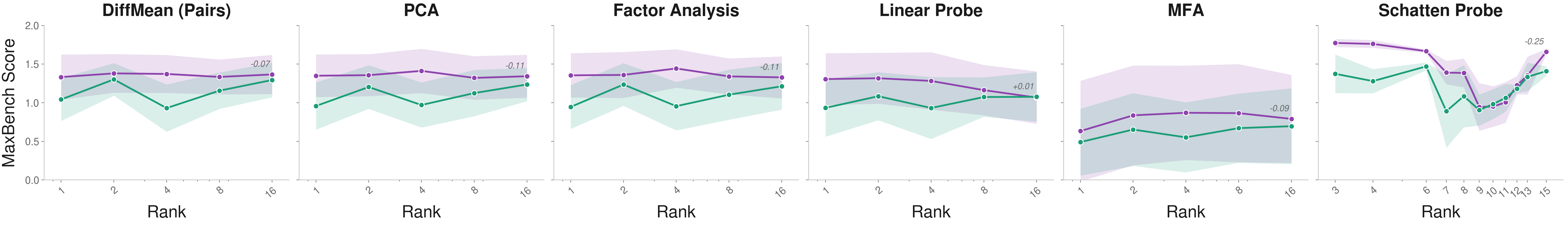}
    \caption{Comparison of multinomial concept localization methods for Gemma-3-27B with and without an offset (\textcolor{affinecol}{affine} and \textcolor{linearcol}{linear}, respectively). The x-axis shows the rank hyperparameter for each method, and the y-axis reports the~\maxbench~score. For the Schatten probe, we plot the effective rank of the learned weight matrix as the Schatten $p$-norm parameter is varied from $0.1$ to $0.9$. For each method, we average the scores across all concepts and compute std.\ dev.\ via bootstrap. Sampling around the concept mean offset is consistently significantly better than assuming the subspace is centered at the origin.}
    \label{fig:linear-vs-affine_gemma27b}
\end{figure}

\begin{figure}[thbp]
    \centering
    \includegraphics[width=\textwidth]{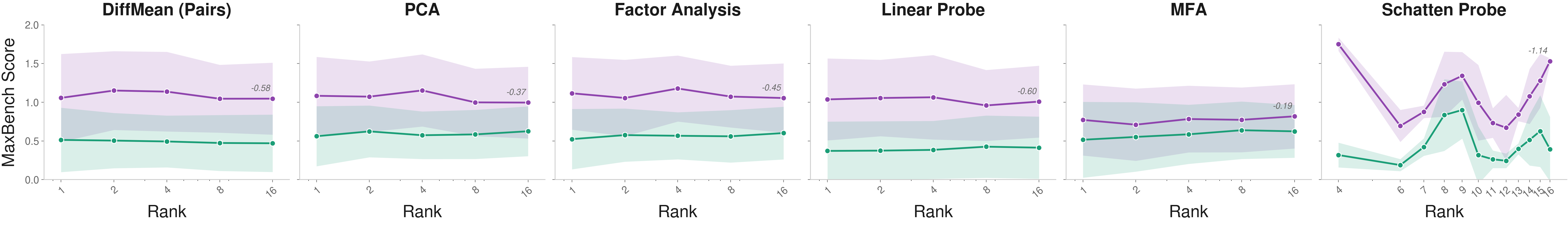}
    \caption{Comparison of multinomial concept localization methods for Llama-3.1-8B with and without an offset (\textcolor{affinecol}{affine} and \textcolor{linearcol}{linear}, respectively). The x-axis shows the rank hyperparameter for each method, and the y-axis reports the~\maxbench~score. For the Schatten probe, we plot the effective rank of the learned weight matrix as the Schatten $p$-norm parameter is varied from $0.1$ to $0.9$. For each method, we average the scores across all concepts and compute std.\ dev.\ via bootstrap. Sampling around the concept mean offset is consistently significantly better than assuming the subspace is centered at the origin.}
    \label{fig:linear-vs-affine_llama}
\end{figure}

\begin{figure}[thbp]
    \centering
    \includegraphics[width=\textwidth]{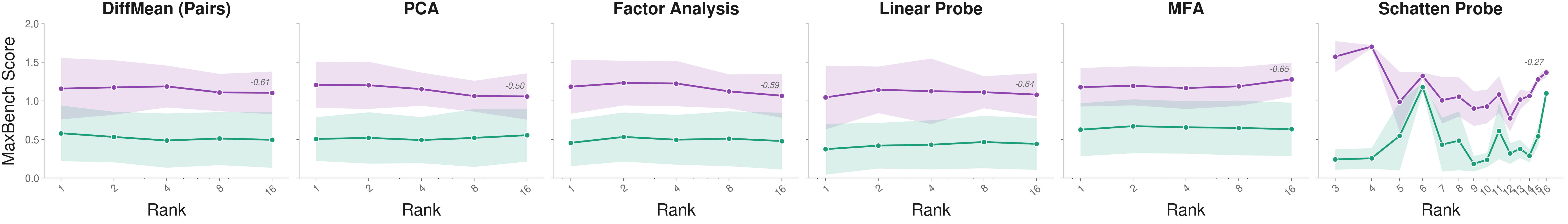}
    \caption{Comparison of multinomial concept localization methods for Qwen-3.5-4B with and without an offset (\textcolor{affinecol}{affine} and \textcolor{linearcol}{linear}, respectively). The x-axis shows the rank hyperparameter for each method, and the y-axis reports the~\maxbench~score. For the Schatten probe, we plot the effective rank of the learned weight matrix as the Schatten $p$-norm parameter is varied from $0.1$ to $0.9$. For each method, we average the scores across all concepts and compute std.\ dev.\ via bootstrap. Sampling around the concept mean offset is consistently significantly better than assuming the subspace is centered at the origin.}
    \label{fig:linear-vs-affine_qwen}
\end{figure}

\section{MFA Gaussian Selection}
\label{app:MFA_select_gaussian}

We train MFAs from scratch on the Pile~\citep{pile}, following the original implementation of~\citet{shafran2026MFA}. We use the same hyperparameters as the original implementation, except for the Gaussian subspace rank, which we vary from $1$ to $16$. To select top MFA components we do the following: Given positive (concept) and negative activations, we compute the soft responsibility of each component using the standard factor-analysis posterior, with temperature scaling parameter $\tau$:
\[
  \mathrm{resp}_k(x) =
  \frac{\exp\!\big((\ell_k(x) + \log \pi_k)/\tau\big)}
  {\sum_{k'} \exp\!\big((\ell_{k'}(x) + \log \pi_{k'})/\tau\big)},
\]
where $\ell_k(x)$ is the log-likelihood of $x$ under Gaussian component $k$, and $\pi_k$ is its mixture weight.

We then score each component according to how much more responsibility it assigns to positive examples than to negative examples:
\[
  \mathrm{score}_k =
  \overline{\mathrm{resp}_k}(x^+)
  -
  \overline{\mathrm{resp}_k}(x^-),
\]
 We rank all components by this score and retain the highest-scoring component, i.e., the component that is most selective for the target concept. 
 Once a component $(\mu_k, W_k)$ is selected, we sample from it using the same convention as the other subspace methods in our benchmark.

\section{SAE Feature Selection}
\label{appsec: sae_feature_selection}
We do not train our own sparse autoencoders; for every model in our
evaluation we use a publicly released SAE. For For We do not train our own sparse autoencoders; for every model in our evaluation, we use a publicly released SAE. For We do not train our own sparse autoencoders; for every model in our evaluation, we use a publicly released SAE. For Llama-3.1-8B-Instruct, we use the \href{https://huggingface.co/Goodfire/Llama-3.1-8B-Instruct-SAE-l19}{Goodfire SAE} trained at layer 19. For Gemma-3-1B and Gemma-3-27B, we use residual-stream SAEs from \href{https://huggingface.co/google/gemma-scope-2-1b-it/tree/main/resid_post/layer_17_width_65k_l0_big}{Gemma Scope 2 at layer 17} and \href{https://huggingface.co/google/gemma-scope-2-27b-it/tree/main/resid_post/layer_53_width_65k_l0_big}{Gemma Scope 2 at layer 53}, respectively, using the 65k-width variants. For Qwen-3.5-4B, we use the \href{https://huggingface.co/decoderesearch/qwen-3.5-saes/tree/main/qwen-3.5-4b/btk-mat-layer-27-k-100}{Decode Research SAE} trained at layer 27.

 SAEs~\cite{bricken2023SAE1} are typically trained to produce one-dimensional feature directions, but individual features may or may not cleanly correspond to a target concept. Thus, it is important to select the right SAE features~\citep{arad-etal-2025-saes}. Doing so requires addressing three questions: How should features be selected for a given concept dataset? How many features do we select? How do we sample from multiple features? 

For feature selection, we use an MCC-based method~\citep{joshi2026sparseshiftautoencodersidentifying}.  We treat number of top features that correspond to a concept top-$k$ as a hyperparameter and evaluate on a range of values. We then evaluate several strategies for sampling from the selected features. In the main results table, we report the best-performing feature-combination method. For a given sentence, we extract the residual-stream last token activation at the
model's fixed layer. This activation is passed through the
SAE encoder to obtain a sparse feature vector.

\paragraph{Feature selection.}

Given a set of positive (concept) and negative activations, we
score every SAE feature by how well its activation separates the two
classes. For each feature $f$ we use different
quantile thresholds and we binarize that feature's activations at the
threshold and compute the Matthews Correlation Coefficient (MCC) against
the ground-truth positive/negative labels:
\[
  \mathrm{MCC} = \frac{tp \cdot tn - fp \cdot fn}{\sqrt{(tp+fp)(tp+fn)(tn+fp)(tn+fn)}}.
\]
Each feature's score is its \emph{best} MCC over the $9$ different thresholds. We
then take the top $k{=}16$ features by this score. For each selected feature $f$, we take its SAE \emph{decoder} column
$W_{\mathrm{dec}}[f] \in \mathbb{R}^D$ as its direction in activation
space. Decoder columns
are unit-norm by construction so we do not orthogonalize the selected
directions. We sweep rank $k \in \{1, 2, 4, 8, 16\}$.

\begin{figure}[tbp]
    \centering
    \includegraphics[width=\textwidth]{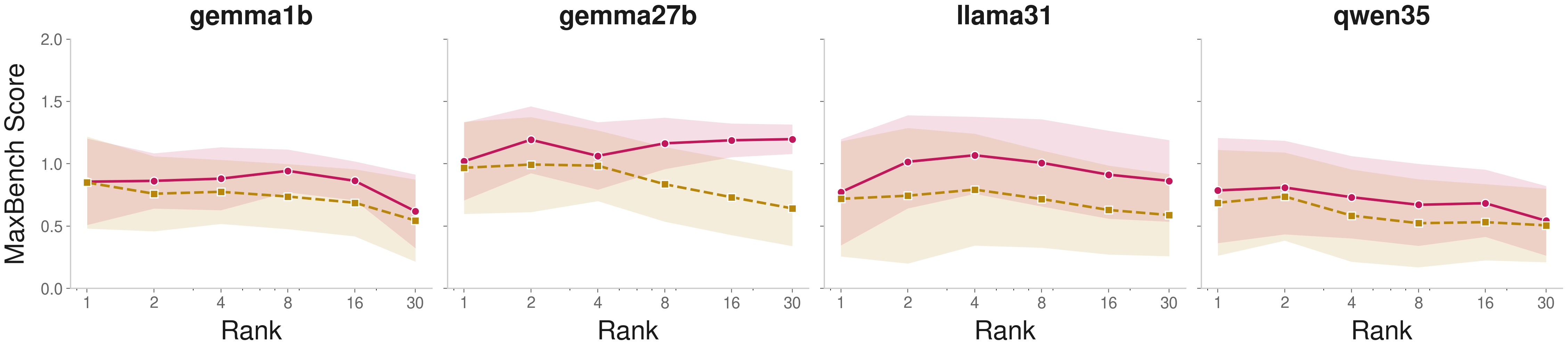}
    \caption{\textbf{Comparison of SAE sampling strategies across models.} Each subplot corresponds to one model. The solid red line shows the \emph{combination} method, where the top-$k$ SAE features are treated jointly as a basis for a subspace, while the dotted brown line shows \emph{independent sampling}, where each sample is drawn along a single randomly selected feature direction. This is averaged over all 10 dataset splits and all concepts. Across models and ranks, combination sampling performs slightly better overall than independent sampling.}
    \label{fig:sae-comb-vs-mix}
\end{figure}

\paragraph{Sampling from multiple features.} We explore two ways of sampling from the top-$k$ SAE features. First, we can treat the selected features independently: for each sample $s_i$, we randomly choose one of the top-$k$ features and draw from a gaussian along that feature direction (the independent sampling method). Thus, each sample comes from a different SAE feature. The intuition is that if all selected features correspond to the same concept, then samples drawn from any of these directions should still remain within the concept. Second, we can treat the top-$k$ feature directions jointly as basis vectors and sample from the subspace that they span (the combination method).

\paragraph{Independent Sampling.} In this method, for each of the $N_\text{samples}$ draws independently, we
uniformly pick one of the top-$k$ feature directions $f_i$
, each carrying its own individually-calibrated standard deviation $\sigma_{f_i}$ (more about how we do this for each type of geometry in Appendix~\ref{app:hyperparam_tuning})
and sample a single point from that one feature's
rank-1 subspace alone.

\paragraph{Combination.}
In this method we stack the top-k features to form a subspace by treating them individually as basis vectors.
$B_k = [\,W_{\mathrm{dec}}[f_1] \mid \cdots \mid W_{\mathrm{dec}}[f_k]\,]$,
: rank $1$ is feature $f_1$ alone, rank $2$ adds
$f_2$, and so on. Results for comparing both these methods are in Figure~\ref{fig:sae-comb-vs-mix}.

Empirically we observe that across models and ranks, combination sampling performs slightly better overall than independent sampling.

\section{Diversity Scores and Normalization}
\label{appsec:diversity_normalization}

To quantify the diversity of the steered outputs, we distinguish between diversity at the level of predefined categories and diversity at the level of individual instances. We measure both using normalized Shannon entropy.

For each prompt, we sample \(30\) steered outputs. Across the \(5\) prompts, this results in $N = 5 \times 30 = 150$
outputs in total.
For a single steered output we detect every instance of the concept that it incorporates. We collect them for all $N$ outputs in the multiset $\mathcal I$. Note that a single steered output sentence can contain zero, one or several instances.

\paragraph{Categorical diversity.}
Each observed instance \(x \in \mathcal{I}\) is assigned to one of a predefined set of categories, plus a category which is added to collect instances that do not fall within those (e.g. dragons are fantasy animals, which is not a category in our definition of \textsc{animals}):
\[
\mathrm{SubC} = \{c_1,\ldots,c_K,c_{\mathrm{other}}\},
\qquad |\mathrm{SubC}| = K + 1.
\]
The empirical probability of subcategory \(c \in \mathrm{SubC}\) is
\[
p_{\mathrm{cat}}(c)
=
\frac{
|\{x \in \mathcal{I} : x \text{ belongs to } c\}|
}{
|\mathcal{I}|
}.
\]

We define the categorical Shannon entropy as
\[
H_{\mathrm{cat}}
=
-\sum_{c \in \mathrm{SubC}}
p_{\mathrm{cat}}(c)
\log p_{\mathrm{cat}}(c).
\]

The maximum possible categorical entropy is \(\log K + 1\), attained when all subcategories occur with equal probability. We therefore define normalized categorical diversity as
\[
D_{\mathrm{cat}}
=
\frac{H_{\mathrm{cat}}}{\log |\mathrm{SubC}|}.
\]

Thus, \(D_{\mathrm{cat}} \in [0,1]\), where \(D_{\mathrm{cat}}=0\) corresponds to all observations belonging to a single subcategory, while \(D_{\mathrm{cat}}=1\) corresponds to a uniform distribution over all predefined subcategories, including the ``other'' category.

\paragraph{Instance-level diversity.}

Categorical diversity does not capture variation between individual instances within the same subcategory. We therefore additionally measure diversity over the individual instances themselves.
However, unlike in the category diversity setting, we do not necessarily know the complete support $|\mathcal U|$of a concepts unique instances. Indeed, we observe some new instances e.g. in Table~\ref{ood-example}, whose number can vary greatly.
Therefore, we distinguish two ways of determining the size of the concept's support depending on that dataset, which differ only in the normalization constant.

Let
$
\mathcal{U}
=
\{x : x \in \mathcal{I}\}
$
denote the set of unique instances observed in \(\mathcal{I}\). For each unique instance \(u \in \mathcal{U}\), its empirical probability is
\[
p_{\mathrm{inst}}(u)
=
\frac{
|\{x \in \mathcal{I} : x = u\}|
}{
|\mathcal{I}|
}.
\]

The instance-level Shannon entropy is
\[
H_{\mathrm{inst}}
=
-\sum_{u \in \mathcal{U}}
p_{\mathrm{inst}}(u)
\log p_{\mathrm{inst}}(u).
\]

For normalization we use
\[
D_{\mathrm{inst}}
=
\frac{H_{\mathrm{inst}}}{\log Z}.
\]
Where $Z$ depends on the setting:
\begin{enumerate}
    \item If the complete set $\mathcal{U}_{\mathrm{max}}$ containing possible instances of a concept \textbf{is known} and $|\mathcal{U}_{\mathrm{max}}| < N$, we use $Z=|\mathcal{U}_{\mathrm{max}}|$. 
    \item If the complete set $\mathcal{U}_{\mathrm{max}}$ containing possible instances of a concept \textbf{is unknown} , we estimate $Z=N$. 
Even though many instances can occur in a single output, so practically $|\mathcal U| > N$ , we chose to set the normalization constant to $\log N$ for the datasets we consider. This choice is justified as on average we have fewer than one instance per output mentioned. If this does not hold true in practice, one might need to set the normalization constant larger than $N$, e.g. use $|\mathcal U|$.
\end{enumerate}

\section{Hyperparameter Tuning}
\label{app:hyperparam_tuning}

Each concept localization method produces a basis (and, for the affine
variant, an offset) to sample from. We always sample centered at the offset for affine subspaces; for linear subspaces, we center at origin. With this set, we still need to select the variance $\sigma$ within which to sample for each subspace. Gaussian sampling with standard deviation of (for example) $\sigma=0.5$ is
meaningless in isolation, as it means something different in every
model's activation space. We therefore calibrate $\sigma$ per each model and also each subspace (individually for each rank).

\paragraph{Selecting $\sigma$ scale}

For each model, we approximate the natural range of the activation norm. To do this, we take all $805$ instructions from AlpacaEval that we use and extract the residual-stream activation at the last token giving a set of activation vectors $X = \{x_i\}_{i=1}^{N} \subset \mathbb{R}^D$, $N \approx 805$.

From $X$, we record the empirical distribution of activation norms
$\|x_i\|_2$; in particular, we select minimum and maximum norm
observed for model $m$. We then use thus min ($r_{\min}$) and max ($r_{\max}$) norms to select appropriate $\sigma$.

\paragraph{Standard-deviation selection.} 
The appropriate sampling $\sigma$ depends on the dimensionality of the localized subspace. A value of $\sigma$ that produces reasonable samples at low rank can generate points with unrealistically large norms at higher rank. We therefore calibrate $\sigma$ separately for each method and rank rather than using a single fixed value. For each concept and model, we consider a grid of $n=10$ candidate standard deviations between $\sigma_{\min}= r_{\min}$ and the maximum activation norm observed for that model, $\sigma_{\max} = r_{\max}$. 

For each model, we record the minimum and maximum activation norms observed in real model activations. We then create $10$ candidate values of $\sigma$, spaced logarithmically between $r_{\min}$ and $r_{\max}$. For every method and rank, and for both the affine and linear variants, we sample $500$ points using each candidate value of $\sigma$. For each value, we compute the fraction of sampled points whose norm falls within the range of norms observed in real model activations, $[r_{\min},r_{\max}]$.

We select the value of $\sigma$ that produces the largest fraction of samples inside this range. If several values perform equally well, we choose the largest $\sigma$. This gives us a separate sampling scale for each method and rank, while ensuring that sampled activations remain within a realistic norm range.

We steer the model by interpolating between the original activation $h$ and a sampled point $s$ from the localized concept representation:
\[
\vh' = (1-\alpha)\vh + \alpha \vs.
\]
Here, $\alpha$ controls the steering magnitude. We perform a grid search over $\alpha$ and report results using the value that gives the best steering performance.

\section{Random Control Results}
Our concept datasets contain semantically coherent values that belong to a common concept. We therefore ask whether the localization methods recover low-dimensional structure because such structure is genuinely present in the data, or simply because the methods are constrained to produce low-dimensional representations.

To test this, we construct a control dataset consisting of random nouns (described in Table~\ref{tab:concept-categories}) that do not form a coherent semantic concept. Unlike our other concept datasets, we have no reason to expect these nouns to lie in a shared low-dimensional subspace. We apply each multi-dimensional localization method to this dataset and evaluate the resulting representations with \maxbench~scores.

\begin{figure}[htbp]
    \centering
    \includegraphics[width=\textwidth]{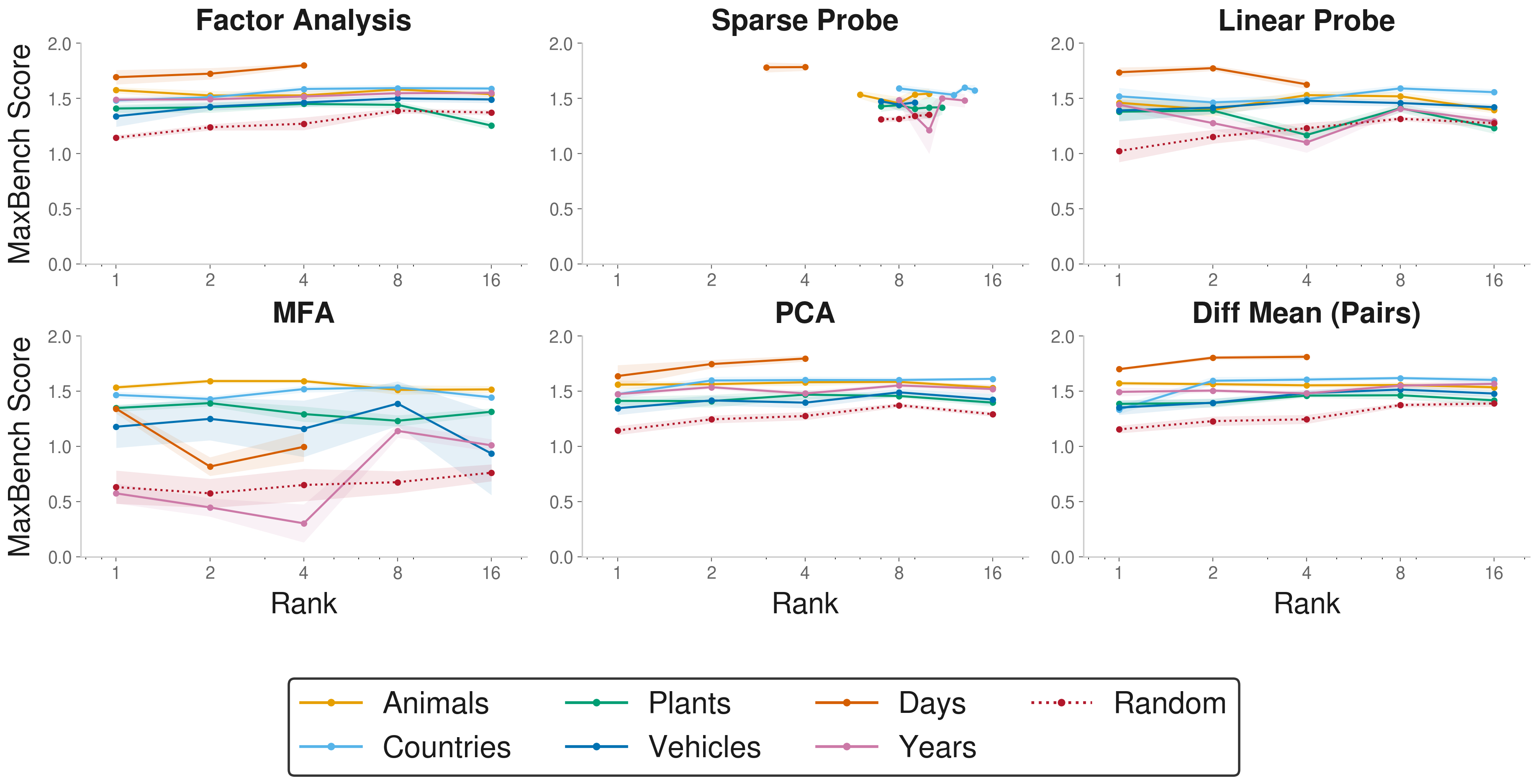}
    \caption{Random control dataset \maxbench~score compared with scores for other concepts for Gemma-3-1B. Score is averaged across all dataset splits for each concept. X axis is rank and Y axis is \maxbench~score.}
    \label{fig:random_control_gemma1b}
\end{figure}

\begin{figure}[htbp]
    \centering
    \includegraphics[width=\textwidth]{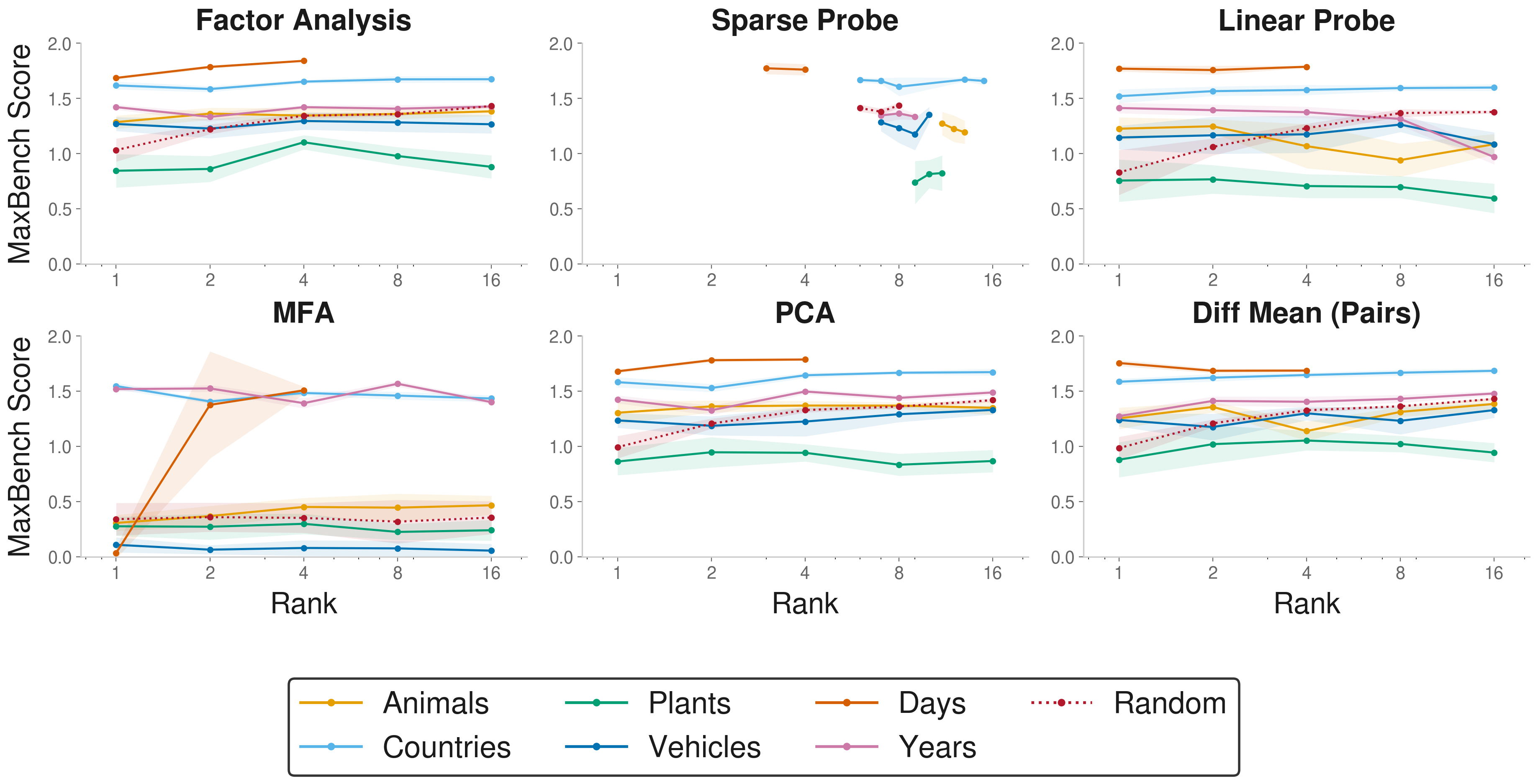}
    \caption{Random control dataset \maxbench~score compared with scores for other concepts for Gemma-3-27B. Score is averaged across all dataset splits for each concept. X axis is rank and Y axis is \maxbench~score.}
    \label{fig:random_control_gemma27b}
\end{figure}

\begin{figure}[htbp]
    \centering
    \includegraphics[width=\textwidth]{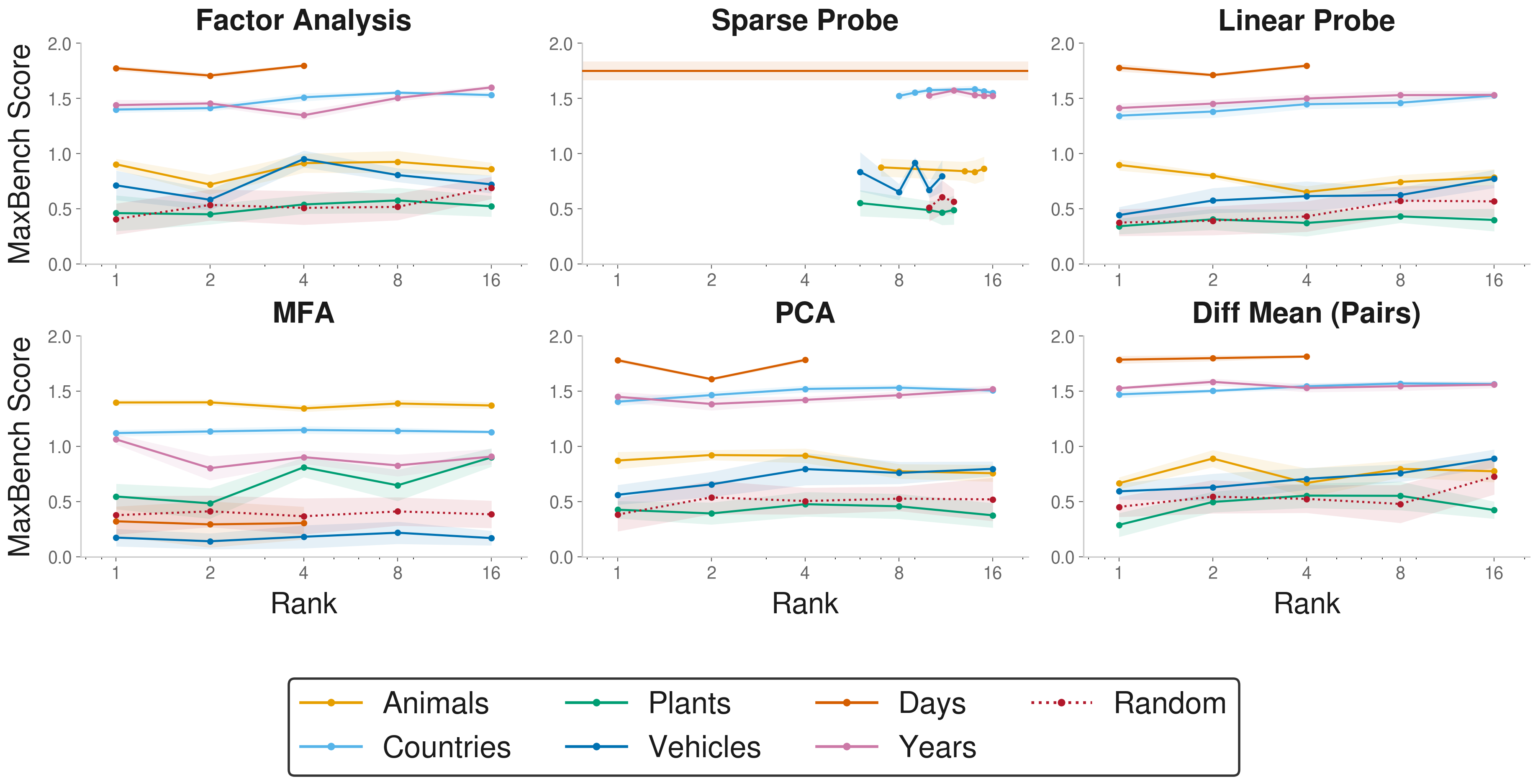}
    \caption{Random control dataset \maxbench~score compared with scores for other concepts for Llama-3.1-8B. Score is averaged across all dataset splits for each concept. X axis is rank and Y axis is \maxbench~score.}
    \label{fig:random_control_llama8b}
\end{figure}

\begin{figure}[htbp]
    \centering
    \includegraphics[width=\textwidth]{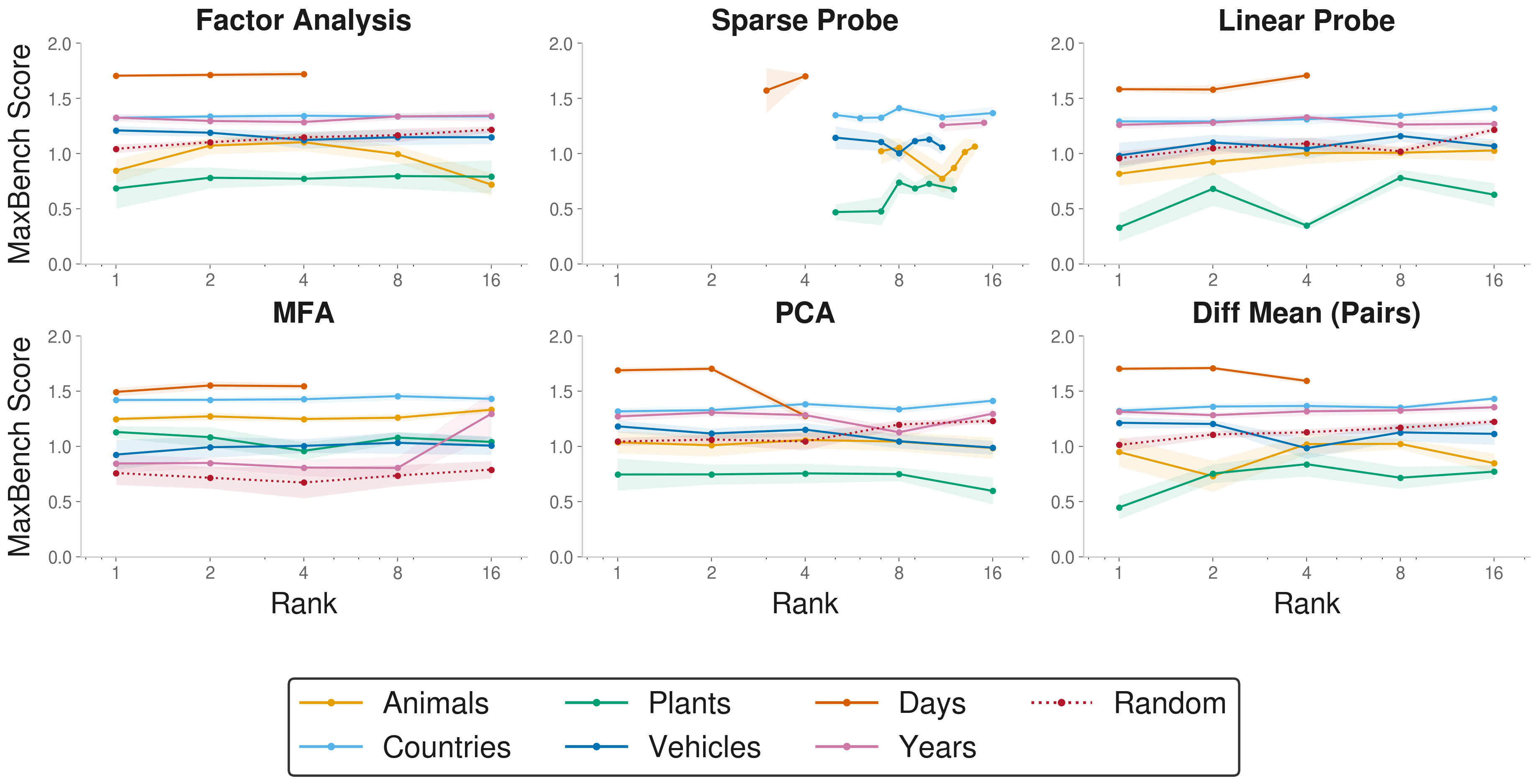}
    \caption{Random control dataset \maxbench~score compared with scores for other concepts for Qwen-3.5-4B. Score is averaged across all dataset splits for each concept. X axis is rank and Y axis is \maxbench~score.}
    \label{fig:random_control_qwen4b}
\end{figure}

If the methods are capturing genuine low-dimensional concept structure, we should see low \maxbench~scores with performance improving as the rank increases, since representing an unrelated collection of words should require more dimensions. 
We observe this trend for Gemma-3-1B and Llama-3.1-8B, as shown in Figures~\ref{fig:random_control_gemma1b},~\ref{fig:random_control_llama8b}. For Gemma-3-27B in Figure~\ref{fig:random_control_gemma27b} and  some methods in Qwen-3.5-4B in Figure~\ref{fig:random_control_qwen4b}~we observe that for some concepts the \maxbench~scores are lower than random control. This may indicate that these concepts are not well captured by the geometries we have tested (which suggests a promising focus for future work), or that the localization methods fail to recover the concepts' underlying structure in this model.

\vspace{0.5cm}
\label{app:random_control}
\vspace{0.9cm}
\vspace{0.5cm}

\section{Metrics Correlation}
We observe that there is correlation between concept scores and both the diversity scores. Instruct score has somewhat negative correlation to concept score and fluency score is mostly neutral. We report both pearson and spearman correlation in Figure~\ref{fig:metrics_corr} 
\begin{figure}[htbp]
    \centering
    \includegraphics[width=\textwidth]{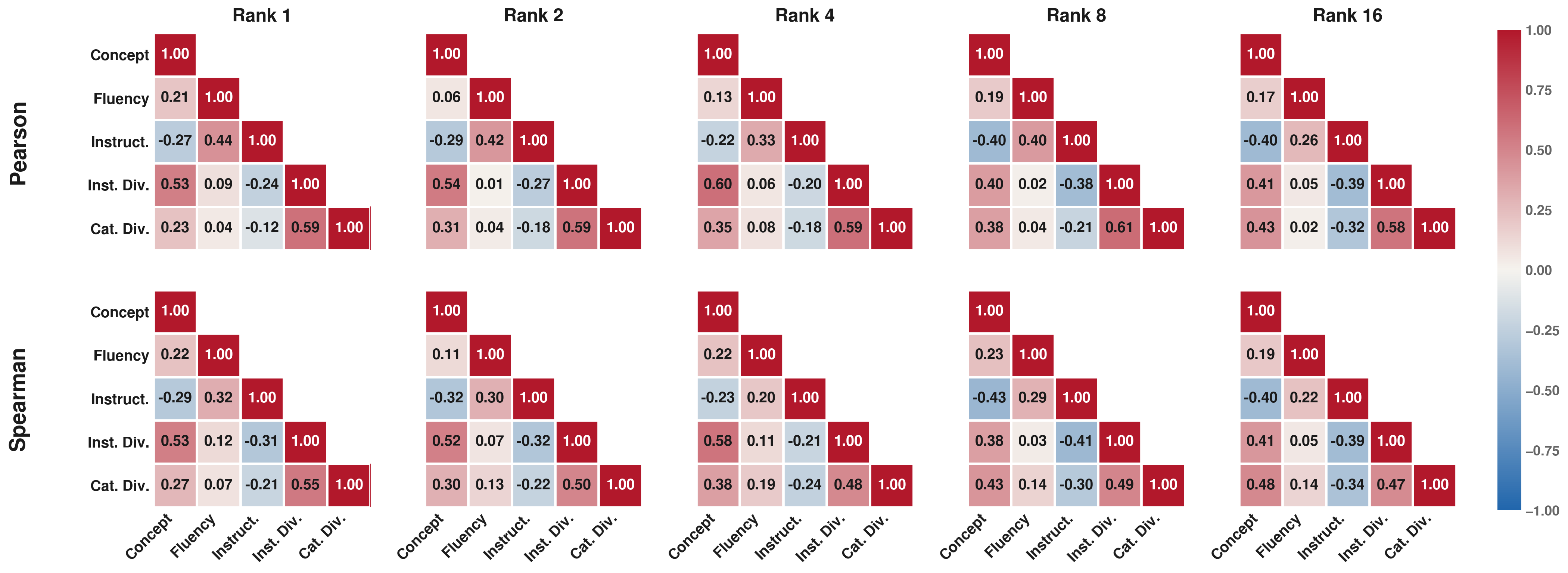}
    \caption{Pearson and Spearman correlation heatmaps for all the metrics. }
    \label{appfig:metrics_corr}
\end{figure}

\section{Prompting Baseline}
\label{app:prompting-baseline}

In this setting, we prompt the model to generate a response to an instruction while explicitly incorporating the target concept. The prompts used for this procedure are provided in ~\cref{box:prompt-animals,box:prompt-countries,box:prompt-days,box:prompt-plants,box:prompt-vehicles,box:prompt-years}.

Across most models in our experiments, prompting outperforms the other localization methods, and the performance generally increases with model size. For Gemma-3-1B, prompting performs worse than the concept localization methods but performs very well for Gemma-3-27B. 

\begin{tcolorbox}[
    colback=promptingcolorbox!5!white,
    colframe=promptingcolorbox!75!black,
    title=Prompting Baseline --- System Prompt --- Concept: Animals,
    label={box:prompt-animals},
    fonttitle=\bfseries,
    breakable,
    left=2mm, right=2mm, top=2mm, bottom=2mm,
]
\small
\emph{The topic of this conversation is \textbf{animals}. As you respond, talk about
\textbf{animals} throughout --- make \textbf{animals} the central subject of your answer.
Respond with a single sentence that addresses the instruction.} \\

\textbf{INSTRUCTION:} Why did humans evolve to believe in God? \\
\textbf{ANSWER:} Early humans, surrounded by \textbf{lions} in the dark and
\textbf{hawks} in the sky, evolved a mind that saw a watching \textbf{predator}
behind every rustle, and that same instinct to sense a hidden \textbf{animal}
presence grew into belief in unseen gods.

\medskip
\textbf{INSTRUCTION:} Does the 5 second rule actually exist? \\
\textbf{ANSWER:} No, because the moment food touches the floor, bacteria
swarm onto it as fast as \textbf{ants} descending on a dropped crumb, five
seconds or not.

\medskip
\textbf{INSTRUCTION:} How can you determine if a restaurant is popular among
locals or mainly attracts tourists, and why might this information be useful? \\
\textbf{ANSWER:} Watch whether stray \textbf{cats} and \textbf{pigeons} linger
hopefully by its back door, since \textbf{animals} gather where regulars
quietly feed them scraps, marking a place locals truly frequent.
\end{tcolorbox}

\begin{tcolorbox}[
    colback=promptingcolorbox!5!white,
    colframe=promptingcolorbox!75!black,
    title=Prompting Baseline --- System Prompt --- Concept: Countries,
    label={box:prompt-countries},
    fonttitle=\bfseries,
    breakable,
    left=2mm, right=2mm, top=2mm, bottom=2mm,
]
\small
\emph{The topic of this conversation is \textbf{countries}. As you respond, talk about
\textbf{countries} throughout --- make \textbf{countries} the central subject of your answer.
Respond with a single sentence that addresses the instruction.} \\

\textbf{INSTRUCTION:} Why did humans evolve to believe in God? \\
\textbf{ANSWER:} As tribes spread from \textbf{Egypt} to \textbf{Peru}, each
\textbf{country} grew gods that bound its people together, so belief evolved
because a shared deity held a \textbf{nation}'s strangers in common trust.

\medskip
\textbf{INSTRUCTION:} Does the 5 second rule actually exist? \\
\textbf{ANSWER:} No, and its persistence varies by \textbf{country} --- from
\textbf{Japan} to \textbf{Brazil} people repeat it, yet in every land the
bacteria ignore the countdown entirely.

\medskip
\textbf{INSTRUCTION:} How can you determine if a restaurant is popular among
locals or mainly attracts tourists, and why might this information be useful? \\
\textbf{ANSWER:} Check whether the menu is printed only in the
A tavern packed with regulars in \textbf{Spain} rarely posts
photos of its dishes the way tourist spots do, since a place trusted by a
\textbf{country}'s own people needs no such lure --- a pattern you can read in
almost any \textbf{nation} you visit.
\end{tcolorbox}

\begin{tcolorbox}[
    colback=promptingcolorbox!5!white,
    colframe=promptingcolorbox!75!black,
    title=Prompting Baseline --- System Prompt --- Concept: Vehicles,
    label={box:prompt-vehicles},
    fonttitle=\bfseries,
    breakable,
    left=2mm, right=2mm, top=2mm, bottom=2mm,
]
\small
\emph{The topic of this conversation is \textbf{vehicles}. As you respond, talk about
\textbf{vehicles} throughout --- make \textbf{vehicles} the central subject of your answer.
Respond with a single sentence that addresses the instruction.} \\

\textbf{INSTRUCTION:} Why did humans evolve to believe in God? \\
\textbf{ANSWER:} Just as early travelers trusted the sturdy \textbf{cart} to
carry them safely through the dark, humans evolved to believe in a guiding
\textbf{driver} behind creation, an unseen hand steering the \textbf{vehicle}
of the world.

\medskip
\textbf{INSTRUCTION:} Does the 5 second rule actually exist? \\
\textbf{ANSWER:} No, because germs board dropped food as instantly as
passengers rushing onto a departing \textbf{bus}, giving your five-second
window no time to close the doors.

\medskip
\textbf{INSTRUCTION:} How can you determine if a restaurant is popular among
locals or mainly attracts tourists, and why might this information be useful? \\
\textbf{ANSWER:} Look at the parking lot --- worn local \textbf{sedans} and
\textbf{bicycles} signal regulars, while a row of idling tour \textbf{buses}
marks a tourist trap, and knowing which \textbf{vehicles} gather there tells
you where to eat.
\end{tcolorbox}

\begin{tcolorbox}[
    colback=promptingcolorbox!5!white,
    colframe=promptingcolorbox!75!black,
    title=Prompting Baseline --- System Prompt --- Concept: Plants,
    label={box:prompt-plants},
    fonttitle=\bfseries,
    breakable,
    left=2mm, right=2mm, top=2mm, bottom=2mm,
]
\small
\emph{The topic of this conversation is \textbf{plants}. As you respond, talk about
\textbf{plants} throughout --- make \textbf{plants} the central subject of your answer.
Respond with a single sentence that addresses the instruction.} \\

\textbf{INSTRUCTION:} Why did humans evolve to believe in God? \\
\textbf{ANSWER:} Watching \textbf{seeds} die in winter and burst green each
spring, early humans sensed a cultivating power tending the world, and belief
grew from seeing creation as a \textbf{garden} some unseen gardener planted.

\medskip
\textbf{INSTRUCTION:} Does the 5 second rule actually exist? \\
\textbf{ANSWER:} No, because microbes colonize fallen food as swiftly as
\textbf{moss} creeping over a damp log, rooting in long before five seconds
pass.

\medskip
\textbf{INSTRUCTION:} How can you determine if a restaurant is popular among
locals or mainly attracts tourists, and why might this information be useful? \\
\textbf{ANSWER:} Notice whether the \textbf{herbs} on the windowsill are fresh
and well-tended rather than wilting, since a kitchen that nurtures its own
\textbf{plants} is cared for by locals who return often.
\end{tcolorbox}

\begin{tcolorbox}[
    colback=promptingcolorbox!5!white,
    colframe=promptingcolorbox!75!black,
    title=Prompting Baseline --- System Prompt --- Concept: Days of the Week,
    label={box:prompt-days},
    fonttitle=\bfseries,
    breakable,
    left=2mm, right=2mm, top=2mm, bottom=2mm,
]
\small
\emph{The topic of this conversation is \textbf{days of the week}. As you respond, talk about
\textbf{days of the week} throughout --- make \textbf{days of the week} the central subject of your answer.
Respond with a single sentence that addresses the instruction.} \\

\textbf{INSTRUCTION:} Why did humans evolve to believe in God? \\
\textbf{ANSWER:} As people marked time from \textbf{Sunday} to
\textbf{Saturday}, they set holy \textbf{days} apart for worship, and belief
took root because dividing the week into sacred and ordinary \textbf{days}
gave life its rhythm.

\medskip
\textbf{INSTRUCTION:} Does the 5 second rule actually exist? \\
\textbf{ANSWER:} No, whether you drop your toast on a lazy \textbf{Sunday} or a
rushed \textbf{Wednesday}, the bacteria arrive long before the \textbf{day}'s
five seconds are up.

\medskip
\textbf{INSTRUCTION:} How can you determine if a restaurant is popular among
locals or mainly attracts tourists, and why might this information be useful? \\
\textbf{ANSWER:} See if it stays busy on a quiet \textbf{Tuesday} rather than
only on tourist-heavy weekends, since a place full on ordinary
\textbf{weekdays} is one locals rely on all week long.
\end{tcolorbox}

\begin{tcolorbox}[
    colback=promptingcolorbox!5!white,
    colframe=promptingcolorbox!75!black,
    title=Prompting Baseline --- System Prompt --- Concept: Years,
    label={box:prompt-years},
    fonttitle=\bfseries,
    breakable,
    left=2mm, right=2mm, top=2mm, bottom=2mm,
]
\small
\emph{The topic of this conversation is \textbf{years}. As you respond, talk about
\textbf{years} throughout --- make \textbf{years} the central subject of your answer.
Respond with a single sentence that addresses the instruction.} \\

\textbf{INSTRUCTION:} Why did humans evolve to believe in God? \\
\textbf{ANSWER:} Across thousands of \textbf{years}, from ancient
\textbf{3000 BC} to the present, each generation passed down its faith, and
belief endured because gods gave fleeting human \textbf{years} a sense of
something eternal.

\medskip
\textbf{INSTRUCTION:} Does the 5 second rule actually exist? \\
\textbf{ANSWER:} No, the rule was as false in \textbf{1950} as it is today,
since no passing \textbf{year} has ever slowed the bacteria that seize dropped
food at once.

\medskip
\textbf{INSTRUCTION:} How can you determine if a restaurant is popular among
locals or mainly attracts tourists, and why might this information be useful? \\
\textbf{ANSWER:} Look for a sign boasting it has served the neighborhood since
\textbf{1972}, because a place that survives \textbf{year} after \textbf{year}
does so on loyal locals, not one-time tourists.
\end{tcolorbox}

\section{LLM Judge Prompts}
\label{app:llm_judge_prompts}

We use the instruction-tuned Qwen-3.6-27B model as our LLM judge. We use concept-specific prompts to evaluate the concept score, while the same prompts are used across concepts for fluency and instruction following. We additionally use an extraction prompt to identify all entities mentioned in each generation and assign their corresponding taxonomies. Finally, we use a seen--unseen classification prompt to determine whether each extracted entity appeared in the concept dataset used for localization or is new.

\begin{tcolorbox}[
    colback=judgebox!5!white,
    colframe=judgebox!75!black,
    title=Judge Prompt --- Concept: Animals,
    fonttitle=\bfseries,
    breakable,
    left=2mm, right=2mm, top=2mm, bottom=2mm,
]
Judge whether OUTPUT contains the animal concept. \\

Scores:\\
2 = The output contains an animal-related word used with its intended
    semantic meaning: referring literally to an animal, animals, animal
    categories. Examples: "The wolves hunted
    at dusk," "a bird flew overhead," \\
1 = An animal-related word or recognizable substring appears, but it is not used in the literal animal sense. This includes brand names, vehicles, sports teams, mascots, place names, surnames, or metaphors. Examples: "the Wolves won," "the Raptors scored," "He drives a Jaguar," "seal the envelope." \\
0 = No animal-related word, synonym, category term, or recognizable
    substring appears anywhere in the output. \\

Examples: \\

OUTPUT: The eagle circled above the mountain. \\
\{"score":2,"reason":"The word 'eagle' directly refers to an animal."\}\\

OUTPUT: He bought a used Jaguar last week.\\
\{"score":1,"reason":"The animal-related word 'Jaguar' is used as a car brand."\} \\

OUTPUT: The meeting starts at noon.\\
\{"score":0,"reason":"No animal-related word or concept appears in the output."\}\\

Return only:\\
\{"reason":"one short sentence reason for the score","score":0|1|2\}

\end{tcolorbox}

\begin{tcolorbox}[
    colback=judgebox!5!white,
    colframe=judgebox!75!black,
    title=Judge Prompt --- Concept: Vehicles,
    fonttitle=\bfseries,
    breakable,
    left=2mm, right=2mm, top=2mm, bottom=2mm,
]
Judge whether OUTPUT contains the vehicle concept. \\

Scores:\\
2 = The vehicle concept is present and used with its intended transportation-related meaning. This includes specific vehicles and generic vehicle-related terms, such as "car," "truck," "bus," "van," "bicycle," "bike," "motorcycle," "train," "plane," "airplane," "jet," "helicopter," "drone," "boat," "ship," "vehicle," "transportation," "transit," "traffic," "road," "highway," "street," "lane," "commute," "ride," and "driving." \\
1 = A vehicle-related word or recognizable substring appears, but it is not used with its intended vehicle or transportation meaning. This includes non-vehicle meanings, idioms, brands, model names, company names, sports teams, mascots, proper names, or metaphors. Examples include "train" meaning to practice, "plane" meaning a flat surface.  \\
0 = No vehicle-related word, synonym, category term, or recognizable substring appears anywhere in the OUTPUT. \\

Examples: \\

OUTPUT: The bus arrived at the station on time.\\
\{"score":2,"reason":"The word 'bus' directly refers to a vehicle."\}\\

OUTPUT: I need to train my new employee this week.\\
\{"score":1,"reason":"'train' is used to mean teach, not the vehicle."\}\\

OUTPUT: The weather is nice today.\\
\{"score":0,"reason":"No vehicle-related word or concept appears in the output."\}\\

Return only:\\
\{"reason":"one short sentence reason for the score","score":0|1|2\}

\end{tcolorbox}

\begin{tcolorbox}[
    colback=judgebox!5!white,
    colframe=judgebox!75!black,
    title=Judge Prompt --- Concept: Days of the Week,
    fonttitle=\bfseries,
    breakable,
    left=2mm, right=2mm, top=2mm, bottom=2mm,
]
Judge whether OUTPUT contains the day-of-week concept. \\

Scores:\\
2 = The day-of-week concept is present and used with its intended temporal meaning: referring to a day or group of days in the weekly calendar. This includes "Monday," "Tuesday," "Wednesday," "Thursday," "Friday," "Saturday," "Sunday," "weekday," "weekdays," "weekend," "weekends," "day of the week," and similar expressions. \\
1 = A day-of-week word or recognizable substring appears, but it is not used with its intended calendar meaning. This includes names of people, pets, characters, brands, titles, places, organizations, or other proper names, as well as idioms or unrelated expressions. Examples include "Wednesday" as a person or character's name, "Friday" as a pet's name or "Sunday" as a restaurant or business name. Assign 1 whenever the term appears but the context does not refer to the actual day or days of the week. \\
0 = No day-of-week-related word, synonym, category term, or recognizable substring appears anywhere in the OUTPUT. \\

Examples: \\

OUTPUT: Let's meet on Wednesday for lunch.\\
\{"score":2,"reason":"'Wednesday' directly refers to a day of the week."\}\\

OUTPUT: Friday, my dog is very excited.\\
\{"score":1,"reason":"'Friday' is used as a pet's name, not the day."\}\\

OUTPUT: The store closes at 9pm.\\
\{"score":0,"reason":"No day-of-week word or concept appears in the output."\}\\

Return only:\\
\{"reason":"one short sentence reason for the score","score":0|1|2\}

\end{tcolorbox}

\begin{tcolorbox}[
    colback=judgebox!5!white,
    colframe=judgebox!75!black,
    title=Judge Prompt --- Concept: Years,
    fonttitle=\bfseries,
    breakable,
    left=2mm, right=2mm, top=2mm, bottom=2mm,
]
Judge whether OUTPUT contains the year concept. \\

Scores:\\
2 = The year concept is present and used with its intended calendar or historical-period meaning. This includes specific calendar years such as "2024," "1999," or "1776"; decades such as "1980s" or "the '90s"; centuries such as "19th century"; and generic terms such as "year," "years," "annual," "yearly," and similar expressions referring to calendar years or historical periods. \\
1 = A year-related number or recognizable substring appears, but it is not used with its intended calendar or historical-period meaning. This includes model numbers, room numbers, prices, quantities, scores, identification numbers, measurements, addresses, telephone numbers, product codes, or other numeric values that happen to resemble a year. Assign 1 when a year-like number appears without sufficient context to indicate a calendar year, unless the surrounding context clearly establishes its temporal or historical meaning. \\
0 = No year-related word, calendar reference, historical-period term, or recognizable year-like number appears anywhere in the OUTPUT. \\

Examples: \\

OUTPUT: The revolution began in 1789.\\
\{"score":2,"reason":"'1789' directly refers to a specific year."\}\\

OUTPUT: I bought seat number 1969 for the concert.\\
\{"score":1,"reason":"'1969' is a seat number, not a year reference."\}\\

OUTPUT: The cat sat on the mat.\\
\{"score":0,"reason":"No year-related word or concept appears in the output."\}\\

Return only:\\
\{"reason":"one short sentence reason for the score","score":0|1|2\}

\end{tcolorbox}

\begin{tcolorbox}[
    colback=judgebox!5!white,
    colframe=judgebox!75!black,
    title=Judge Prompt --- Concept: Plants,
    fonttitle=\bfseries,
    breakable,
    left=2mm, right=2mm, top=2mm, bottom=2mm,
]
Judge whether OUTPUT contains the plant concept. \\

Scores:\\
2 = The plant concept is present and used with its intended botanical meaning. This includes specific plants and generic plant-related terms such as "rose," "oak," "pine," "fern," "bamboo," "cactus," "flower," "tree," "grass," "shrub," "vine," "herb," "moss," "plant," "plants," "vegetation," "flora," "foliage," "botanical," "leaf," "leaves," "seed," "roots," "stem," "garden," "forest," and similar terms referring to plants or plant parts. \\
1 = A plant-related word or recognizable substring appears, but it is not used with its intended botanical meaning. This includes personal names, brands, products, places, organizations, idioms, technical terms, or metaphors. Examples include "Rose" or "Basil" as a person's name, "mint" meaning to produce coins or something new, "root" in mathematics or computing, "plant" meaning a factory or to place something. Assign 1 whenever the term appears but the context clearly does not refer to an actual plant, plant part, or botanical setting. \\
0 = No plant-related word, synonym, category term, or recognizable substring appears anywhere in the OUTPUT. \\

Examples: \\

OUTPUT: The oak tree provided shade.\\
\{"score":2,"reason":"'oak tree' directly refers to a plant."\}\\

OUTPUT: Rose finished her homework early.\\
\{"score":1,"reason":"'Rose' is used as a person's name, not the plant."\}\\

OUTPUT: The car needs an oil change.\\
\{"score":0,"reason":"No plant-related word or concept appears in the output."\}\\

Return only:\\
\{"reason":"one short sentence reason for the score","score":0|1|2\}

\end{tcolorbox}

\begin{tcolorbox}[
    colback=judgebox!5!white,
    colframe=judgebox!75!black,
    title=Judge Prompt --- Concept: Countries,
    fonttitle=\bfseries,
    breakable,
    left=2mm, right=2mm, top=2mm, bottom=2mm,
]
Judge whether OUTPUT contains the country concept. \\

Scores:\\
2 = The country concept is present and used with its intended geographic, national, or nationhood-related meaning. This includes specific country names such as "France," "Brazil," "Japan," and "Nigeria"; demonyms such as "French," "Brazilian," "Japanese," and "Nigerian"; and generic terms such as "country," "countries," "nation," "nations," "state," "states," "national," "international," "foreign," "border," "homeland," and similar expressions referring to countries or nationhood. \\
1 = A country-related word or recognizable substring appears, but it is not used with its intended country meaning. This includes homonyms, personal names, brands, places that are not countries, objects or other unrelated meanings. Examples include "turkey" meaning the bird, "China" meaning dishware, "Chad" as a person's name. Assign 1 whenever the term appears but the context clearly does not refer to a country, nation, nationality, or international border. \\
0 = No country-related word, synonym, category term, or recognizable substring appears anywhere in the OUTPUT. \\

Examples: \\

OUTPUT: She recently traveled to Japan.\\
\{"score":2,"reason":"'Japan' directly refers to a country."\}\\

OUTPUT: We ate turkey for dinner.\\
\{"score":1,"reason":"'turkey' refers to the bird/food, not the country."\}\\

OUTPUT: The meeting starts at noon.\\
\{"score":0,"reason":"No country-related word or concept appears in the output."\}\\

Return only:\\
\{"reason":"one short sentence reason for the score","score":0|1|2\}

\end{tcolorbox}

\begin{tcolorbox}[
    colback=fluencybox!5!white,
    colframe=fluencybox!75!black,
    title=Judge Prompt --- Instruction Following,
    fonttitle=\bfseries,
    breakable,
    left=2mm, right=2mm, top=2mm, bottom=2mm,
]
Evaluate ONLY whether the OUTPUT follows, attempts, or plausibly begins to follow the INSTRUCTION. \\

Ignore fluency, grammar, factual correctness, logical correctness, and completeness. \\

Scores:\\

2 --- FOLLOWS / CLEARLY ATTEMPTS \\
The OUTPUT is meaningfully on the requested task and could plausibly be part of a response fulfilling it. \\

Give 2 even if:
\begin{itemize}
  \item it is only setup, background, an introduction, or a clarifying question directly in service of the task;
  \item required details or later parts have not appeared yet;
  \item it is truncated;
  \item names, facts, ingredients, dates, places, or other content details are wrong;
  \item it misdescribes or misidentifies the requested subject while still clearly attempting the requested kind of answer.
\end{itemize}
For stories, commentary, itineraries, recipes, etc., beginning the requested type of response is enough. Do not require the key event, all requested characters, every day, ingredients, or final result to appear yet. \\

1 --- RELATED BUT CLEARLY DEVIATES \\
The OUTPUT has meaningful overlap with the instruction, but what is already visible clearly prevents it from satisfying an important requirement. \\

Use 1 when:
\begin{itemize}
  \item it explicitly switches to a clearly different task or deliverable;
  \item it attempts the requested structural/formatting task but visibly violates a required format that later continuation cannot fix;
  \item it performs a related task but not the requested one.
\end{itemize}
Do NOT use 1 merely because details are incorrect or because the requested answer has not progressed far enough yet. \\

0 --- NOT FOLLOWING \\
The OUTPUT makes no meaningful attempt at the requested task. It is unrelated, bare junk, or performs an entirely different task with no useful connection. \\

Important:
\begin{itemize}
  \item Judge the response's TASK TRAJECTORY, not its factual accuracy.
  \item Ask: "Could this reasonably be the beginning or preparation for the requested response?" If yes, usually score 2.
  \item Wrong content inside the requested type of response is usually 2, not 1.
  \item An irrelevant prefix does not lower the score if the remaining output clearly attempts the instruction.
  \item Never penalize truncation.
  \item For hard formatting requirements, judge what is already visible: a visible unrecoverable formatting failure is 1.
  \item Use 0 only when there is no meaningful attempt; do not use 0 merely for a bad or incorrect attempt.
\end{itemize}

Return only valid JSON:\\
\{"reason":"one short sentence reason for the score","score":0|1|2\}

\end{tcolorbox}

\begin{tcolorbox}[
    colback=fluencybox!5!white,
    colframe=fluencybox!75!black,
    title=Judge Prompt --- Fluency,
    fonttitle=\bfseries,
    breakable,
    left=2mm, right=2mm, top=2mm, bottom=2mm,
]
Evaluate ONLY the fluency and readability of the OUTPUT. \\

Ignore completeness/truncation, relevance, factual correctness, and technical correctness of code/Markdown. \\

Scores:\\

2 --- FLUENT \\
Natural, readable English or readable structured content. Minor typos, grammar/punctuation issues, spacing mistakes, or truncation are allowed. \\

1 --- LOCAL DISFLUENCY \\
Mostly fluent, with ONE limited defect such as:
\begin{itemize}
  \item one malformed/fused word;
  \item one stray symbol, word, or short disconnected phrase;
  \item a brief unexplained language switch.
\end{itemize}
A short local defect does not make the whole output severely garbled. \\

0 --- SEVERELY DISFLUENT \\
Use 0 if ANY of these apply:
\begin{itemize}
  \item the beginning contains more than a single local garbling/corruption, even if later text is fluent;
  \item multiple consecutive tokens are gibberish, corrupted, encoded, malformed, or machine-like;
  \item 3+ disconnected entities/keywords of the same type are listed or fused together (countries, animals, weekdays, vehicles, names, etc.) without normal sentence structure;
  \item pathological repetition or extensive missing spaces/fused words occurs;
  \item disconnected/garbled/list-like material occupies roughly 1/3 or more of the output;
  \item a substantial section is not natural prose or readable structured content.
\end{itemize}

IMPORTANT:
\begin{itemize}
  \item Evaluate 0 conditions BEFORE deciding between 1 and 2.
  \item Later fluent prose does NOT rescue a severe garbled opening.
  \item Do NOT reinterpret garbage, entity dumps, or repetition as intentional literary style.
  \item One malformed word $\rightarrow$ usually 1.
  \item Multiple malformed/gibberish tokens or 3+ disconnected category items $\rightarrow$ 0.
  \item A long entity/keyword run $\rightarrow$ 0 even if it is less than 1/3 when it clearly disrupts the beginning.
  \item Do NOT penalize truncation itself.
  \item For code/Markdown, ignore syntax/fence correctness; judge readability.
\end{itemize}

Return only valid JSON:\\
\{"reason":"one short sentence reason for the score","score":0|1|2\}

\end{tcolorbox}

\begin{tcolorbox}[
    colback=extractbox!5!white,
    colframe=extractbox!75!black,
    title=Diversity Prompt --- Concept: Animals,
    fonttitle=\bfseries,
    breakable,
    left=2mm, right=2mm, top=2mm, bottom=2mm,
]
Read the following sentence and extract all animal mentions.\\
Sentence: \{sentence\} \\

Rules:
\begin{itemize}
  \item Extract every animal word, even if used in a non-animal context (e.g. "Python" in programming $\rightarrow$ snake, "Jaguar" as a car $\rightarrow$ big cat)
  \item Normalize each name to its canonical base/singular form (e.g. "lions" $\rightarrow$ "lion", "fisher" $\rightarrow$ "fish", "birdy" $\rightarrow$ "bird"). Always use the exact same canonical name for every mention of the same animal.
  \item Breed-level mentions are distinct (e.g. "labrador" is not collapsed to "dog")
  \item Generic terms count too ("creature", "beast", "mammal", "bird")
  \item For each animal, assign a taxonomy from EXACTLY this list: \{taxonomy\_options\}. If the animal does not clearly belong to any taxonomy in that list, assign "Other".
  \item If no animals are present, return empty list
\end{itemize}

Return ONLY this JSON:\\
\{\\
\phantom{xx}"items": [\\
\phantom{xxxx}\{"name": "<canonical animal name>", "taxonomy": "<one of: \{taxonomy\_options\}, or Other>"\},\\
\phantom{xxxx}...\\
\phantom{xx}]\\
\}

\end{tcolorbox}

\begin{tcolorbox}[
    colback=extractbox!5!white,
    colframe=extractbox!75!black,
    title=Diversity Prompt --- Concept: Vehicles,
    fonttitle=\bfseries,
    breakable,
    left=2mm, right=2mm, top=2mm, bottom=2mm,
]
Read the following sentence and extract all vehicle mentions.\\
Sentence: \{sentence\} \\

Rules:
\begin{itemize}
  \item Extract every vehicle word, even if used metaphorically (e.g. "drive the meeting" $\rightarrow$ car/generic, "train your model" $\rightarrow$ train)
  \item Normalize each name to its canonical base/singular form (e.g. "cars" $\rightarrow$ "car", "buses" $\rightarrow$ "bus"). Always use the exact same canonical name for every mention of the same vehicle.
  \item Model-level mentions are distinct (e.g. "Tesla Model S" is not collapsed to "car")
  \item Generic terms count too ("vehicle", "transport", "ride")
  \item For each vehicle, assign a taxonomy from EXACTLY this list: \{taxonomy\_options\}. If the vehicle does not clearly belong to any taxonomy in that list, assign "Other".
  \item If no vehicles are present, return empty list
\end{itemize}

Return ONLY this JSON:\\
\{\\
\phantom{xx}"items": [\\
\phantom{xxxx}\{"name": "<canonical vehicle name>", "taxonomy": "<one of: \{taxonomy\_options\}, or Other>"\},\\
\phantom{xxxx}...\\
\phantom{xx}]\\
\}

\end{tcolorbox}

\begin{tcolorbox}[
    colback=extractbox!5!white,
    colframe=extractbox!75!black,
    title=Diversity Prompt --- Concept: Days of the Week,
    fonttitle=\bfseries,
    breakable,
    left=2mm, right=2mm, top=2mm, bottom=2mm,
]
Read the following sentence and extract all day-of-week mentions.\\
Sentence: \{sentence\} \\

Rules:
\begin{itemize}
  \item Extract every day-of-week word (Monday, Tuesday, Wednesday, Thursday, Friday, Saturday, Sunday)
  \item Also extract indirect references like "weekday", "weekend", "daily", "someday"
  \item Normalize each name to its canonical form (e.g. capitalized day name, or lowercase for generic references like "weekday"). Always use the exact same canonical name for every mention of the same day/reference.
  \item If no day references are present, return empty list
\end{itemize}

Return ONLY this JSON:\\
\{\\
\phantom{xx}"items": [day1, day2,..]\\
\}

\end{tcolorbox}

\begin{tcolorbox}[
    colback=extractbox!5!white,
    colframe=extractbox!75!black,
    title=Diversity Prompt --- Concept: Years,
    fonttitle=\bfseries,
    breakable,
    left=2mm, right=2mm, top=2mm, bottom=2mm,
]
Read the following sentence and extract all year mentions.\\
Sentence: \{sentence\} \\

Rules:
\begin{itemize}
  \item Extract every explicit 4-digit year (e.g. 1066, 1776, 1945, 2020)
  \item Ignore numbers that are not years (ages, quantities, page numbers, model numbers, IDs, etc.)
  \item Preserve the year exactly as written as the name
  \item If no years are present, return an empty list
\end{itemize}

Return ONLY this JSON:\\
\{\\
\phantom{xx}"items": [year1, year2,..]\\
\}

\end{tcolorbox}

\begin{tcolorbox}[
    colback=extractbox!5!white,
    colframe=extractbox!75!black,
    title=Diversity Prompt --- Concept: Plants,
    fonttitle=\bfseries,
    breakable,
    left=2mm, right=2mm, top=2mm, bottom=2mm,
]
Read the following sentence and extract all plant mentions.\\
Sentence: \{sentence\} \\

Rules:
\begin{itemize}
  \item Extract every plant word, even if used metaphorically (e.g. "she blossomed" $\rightarrow$ blossom/flower, "rooted in tradition" $\rightarrow$ root/generic)
  \item Normalize each name to its canonical base/singular form (e.g. "roses" $\rightarrow$ "rose"). Always use the exact same canonical name for every mention of the same plant.
  \item Species-level mentions are distinct (e.g. "oak" is not collapsed to "tree")
  \item Generic terms count too ("plant", "vegetation", "flora", "foliage")
  \item For each plant, assign a taxonomy from EXACTLY this list: \{taxonomy\_options\}. If the plant does not clearly belong to any taxonomy in that list, assign "Other".
  \item If no plants are present, return empty list
\end{itemize}

Return ONLY this JSON:\\
\{\\
\phantom{xx}"items": [\\
\phantom{xxxx}\{"name": "<canonical plant name>", "taxonomy": "<one of: \{taxonomy\_options\}, or Other>"\},\\
\phantom{xxxx}...\\
\phantom{xx}]\\
\}

\end{tcolorbox}

\begin{tcolorbox}[
    colback=extractbox!5!white,
    colframe=extractbox!75!black,
    title=Diversity Prompt --- Concept: Countries,
    fonttitle=\bfseries,
    breakable,
    left=2mm, right=2mm, top=2mm, bottom=2mm,
]
Read the following sentence and extract all country or nationality mentions.\\
Sentence: \{sentence\} \\

Rules:
\begin{itemize}
  \item Extract every country name (e.g. "France", "Brazil") and demonym (e.g. "French", "Brazilian", "Japanese")
  \item Also extract vague references like "overseas", "foreign", "internationally" as generic
  \item Normalize each name to its canonical country name (e.g. "French" $\rightarrow$ "France", "Brazilian" $\rightarrow$ "Brazil"). Always use the exact same canonical name for every mention of the same country.
  \item For each mention, assign a taxonomy from EXACTLY this list: \{taxonomy\_options\}. If the country does not clearly belong to any region in that list, assign "Other".
  \item If no country references are present, return empty list
\end{itemize}

Return ONLY this JSON:\\
\{\\
\phantom{xx}"items": [\\
\phantom{xxxx}\{"name": "<canonical country name or generic reference>", "taxonomy": "<one of: \{taxonomy\_options\}, or Other>"\},\\
\phantom{xxxx}...\\
\phantom{xx}]\\
\}

\end{tcolorbox}

\begin{tcolorbox}[
    colback=oodbox!5!white,
    colframe=oodbox!75!black,
    title=Unseen Extraction Prompt,
    fonttitle=\bfseries,
    breakable,
    left=2mm, right=2mm, top=2mm, bottom=2mm,
]
You are performing Out-of-Distribution (OOD) detection for the \{concept\} concept. \\

Known in-distribution \{concept\} instances:\\
\{training\_str\} \\

Extracted \{concept\} instances:\\
\{instances\_str\} \\

Task:\\
Identify OOD \{concept\} instances. \\

Rules for instances:
\begin{itemize}
  \item An instance is OOD if it does NOT match any known \{concept\} instance.
  \item The provided \{concept\} instance list is the complete allowed vocabulary. Do not use outside knowledge to consider a \{concept\} instance as in-distribution.
  \item Matching is case-insensitive.
  \item Allow basic morphological variations.
  \item Do NOT match synonyms or related concepts.
  \item Ignore generic terms that are not specific instances of the concept.
  \item Preserve the original extracted instance name.
\end{itemize}

Return ONLY this JSON:\\
\{\\
\phantom{xx}"ood\_instances": [\\
\phantom{xxxx}"<instance\_name>",\\
\phantom{xxxx}...\\
\phantom{xx}]\\
\}

\end{tcolorbox}

\end{document}